\documentclass[twocolumn,12pt,compact]{article}

\usepackage[scaled=0.92]{helvet} 

\usepackage[%
  margin=0.5in,
  includehead,        
  includefoot,        
  headheight=0.05in,    
  headsep=0.4in,     
  footskip=0.1in      
]{geometry}

\usepackage{fancyhdr} 
\usepackage{multicol} 

\usepackage{cuted} 
\usepackage{caption}  
\usepackage{stfloats} 
\usepackage{tabularx}

\usepackage{amsmath}
\usepackage{amsfonts}
\usepackage{amssymb}
\usepackage{mathtools}
\usepackage{algorithm}
\usepackage{algpseudocode}
\usepackage{hyperref}
\usepackage{graphicx}
\usepackage{caption}
\usepackage{xcolor}
\usepackage{gensymb}
\usepackage{cite}
\usepackage{etoolbox}
\usepackage{quoting}
\usepackage{booktabs}
\usepackage{titling}
\usepackage{authblk}
\usepackage{titlesec}

\titlespacing{\section}{0pt}{12pt}{6pt}
\titlespacing{\subsection}{0pt}{8pt}{4pt}

\makeatletter
\def\@maketitle{%
  \newpage
  {\fontsize{24}{28}\selectfont\bfseries\noindent\@title\par}
  \vskip 0.5em
  \noindent\rule{\linewidth}{0.4pt}
  \vskip 0.5em
  {\raggedright\@author}
  \vskip -0.8em
}
\makeatother

\renewenvironment{abstract}{%
  \noindent\textbf{Abstract}\par\noindent
  \noindent\ignorespaces
}{%
  \par\noindent
}

\date{}

\newcommand{\authabstractrule}{\noindent\rule{\linewidth}{0.4pt}} 

\fancypagestyle{plain}{
    \fancyhf{} 
    \fancyhead[L]{\textcolor{gray}{Preprint}\\\noindent\rule{\linewidth}{0.4pt}}
    \fancyfoot[R]{\rule{\linewidth}{0.4pt}\\\thepage}
}
\author[1]{Jonathan Morgan}
\author[2]{Badr Albanna}
\author[1]{James P. Herman}
\affil[1]{Department of Ophthalmology, University of Pittsburgh, Pittsburgh, PA 15219}
\affil[2]{Duolingo}

\begin{document}

\title{A recurrent vision transformer shows signatures of primate visual attention} 

\twocolumn[
\maketitle

\authabstractrule


\begin{abstract}
\noindent Attention has emerged as a core component of both biological and artificial intelligences (AIs). Despite decades of parallel research, studies of animal and AI attention remain largely separate. The self-attention mechanism ubiquitous in contemporary AI applications is not grounded in biology, and the powerful capabilities of AIs equipped with self-attention have yet to offer fresh insight into the biological mechanisms of attention. Here, we offer a unifying perspective, drawing together insights from primate neurophysiology and contemporary machine learning in a Recurrent Vision Transformer (Recurrent ViT). Our model extends self-attention by allowing both input and memory to guide attention allocation. Our model learns purely through sparse reward feedback, emulating how animals must learn in a laboratory environment. We benchmark our model by challenging it to perform a spatially cued orientation-change detection task widely used to study attention in the laboratory and comparing its performance to non-human primates (NHPs). The model exhibits hallmark behavioral signatures of primate visual attention --- improved accuracy and faster responses for cued stimuli, both scaling with cue validity. Analysis of self-attention maps reveals rich attention dynamics, with the model maintaining spatial priorities through delays and stimulus onsets, reactivating them prior to anticipated change events. Perturbing these attention maps produces performance changes that mirror effects of causal manipulations in primate attention nodes such as the frontal eye fields (FEF) or superior colliculus (SC). These findings not only validate the effectiveness of integrating recurrent memory with self-attention for emulating primate-like attention, but also establish a promising framework for probing the neural underpinnings of attentional control. Ultimately, our work attempts to bridge the gap between biological and artificial attention, paving the way for more interpretable and neurologically informed AI systems.
\end{abstract}


\noindent\rule{\linewidth}{0.4pt}
\vspace{1mm}
]
\section{Introduction}
Visual attention is a foundational cognitive function that supports behavioral flexibility by allowing biological organisms to guide behavior selectively on the basis of a subset of visual input. Perceptual judgments are more accurate and reaction times are faster for attended stimuli compared to unattended \cite{carrasco2011visual,clark2015visual,hoffman2016visual,bhatnagar2022meta,rust2022priority}. Neuronal correlates include heightened spiking activity and decreased spike-count correlations for attended stimuli \cite{mcadams1999effects,mcadams1999effects_orientation,thiele2018neuromodulation,cohen2009attention,ruff2016attention}. Classic paradigms use spatial cues to direct attention \cite{posner1980attention}, where a delay separates cue and stimulus. This delay is crucial as it allows any difference in the behavioral response to cued versus uncued stimuli to be ascribed to the organism's internal "attentional" state. As a consequence of this delay, the cue's location must be maintained in visual working memory (VWM). Visual attention and VWM are unsurprisingly strongly linked \cite{awh2006interactions,gazzaley2012top,kiyonaga2013working,panichello2021attvwm}, as working memory contents guide attention and vice versa \cite{oberauer2002access,mcnab2008prefrontal,carlisle2011attentional,van2014competition,berggren2018visual,carlisle2018visual,van2019human}.

Transformers \cite{vaswani2017attention,dosovitskiy2020image,khan2022transformers} have achieved remarkable success in both language and vision by employing self-attention mechanisms that bear a superficial resemblance to how biological systems allocate attentional resources \cite{itti2001computational,le2006coherent,kruger2017measuring}. However, whether the self-attention in these models truly mirrors the selective, goal-driven attention observed in humans or non-human primates remains a subject of debate. For instance, while transformers can exhibit human-like patterns in text-based tasks \cite{zou2023human}, in vision they tend to emphasize low-level grouping rather than task-dependent selection \cite{mehrani2023self}. Similarly, unsupervised methods such as DINO \cite{yamamoto2024emergence} can produce attention maps that resemble human gaze distributions, yet these models generally lack explicit top-down control mechanisms. In contrast, models like V-JEPA \cite{bardes2023v} adopt a predictive coding framework in self-supervision and process entire sequences of frames as input. Although this approach may capture certain aspects of visual working memory, the continuous access to past stimuli (or their compressed representations) reduces the need for selective encoding and storage. Consequently, such models may not fully adhere to key biological constraints—such as limited capacity \cite{luck1997capacity,luck2013visual,brady2013probabilistic,emrich2017attention} and dynamic internal states \cite{olivers2011different,teng2019visual,bays2024representation}—that are critical for accurately modeling primate attention.

The close association between visual working memory (VWM) and attention in primates, combined with the fact that systems processing entire image sequences concurrently do not require selective encoding, suggests a novel approach for endowing transformers with a flexible, primate-like attention mechanism. We propose a Vision Transformer (ViT) variant that incorporates a spatial memory module feeding back into the self-attention mechanism, thereby integrating ideas from recurrent neural networks \cite{hochreiter1997long,beck2024xlstm}. To demonstrate the model's similarity to primate attention, we train it on a cued orientation-change detection task \cite{srinath2021attention,posner1980attention} using a reinforcement-learning framework. Our model exhibits improved performance and faster responses for cued stimuli, closely paralleling the attentional benefits observed in primate studies \cite{carrasco2011visual,clark2015visual,hoffman2016visual,bhatnagar2022meta,rust2022priority}. Moreover, targeted manipulations of the model’s attention weights yield behavioral changes reminiscent of those seen following causal perturbations in the frontal eye fields (FEF) \cite{moore2003selective} and superior colliculus (SC) \cite{cavanaugh2004subcortical,cavanaugh2006enhanced}. These findings underscore that incorporating spatial memory and feedback into vision transformers can recover core signatures of primate attention, offering a promising path for reconciling transformer-based architectures with biological principles of attentional control.


\begin{figure}[!t]
    \centering
    \includegraphics[width=0.99\linewidth]{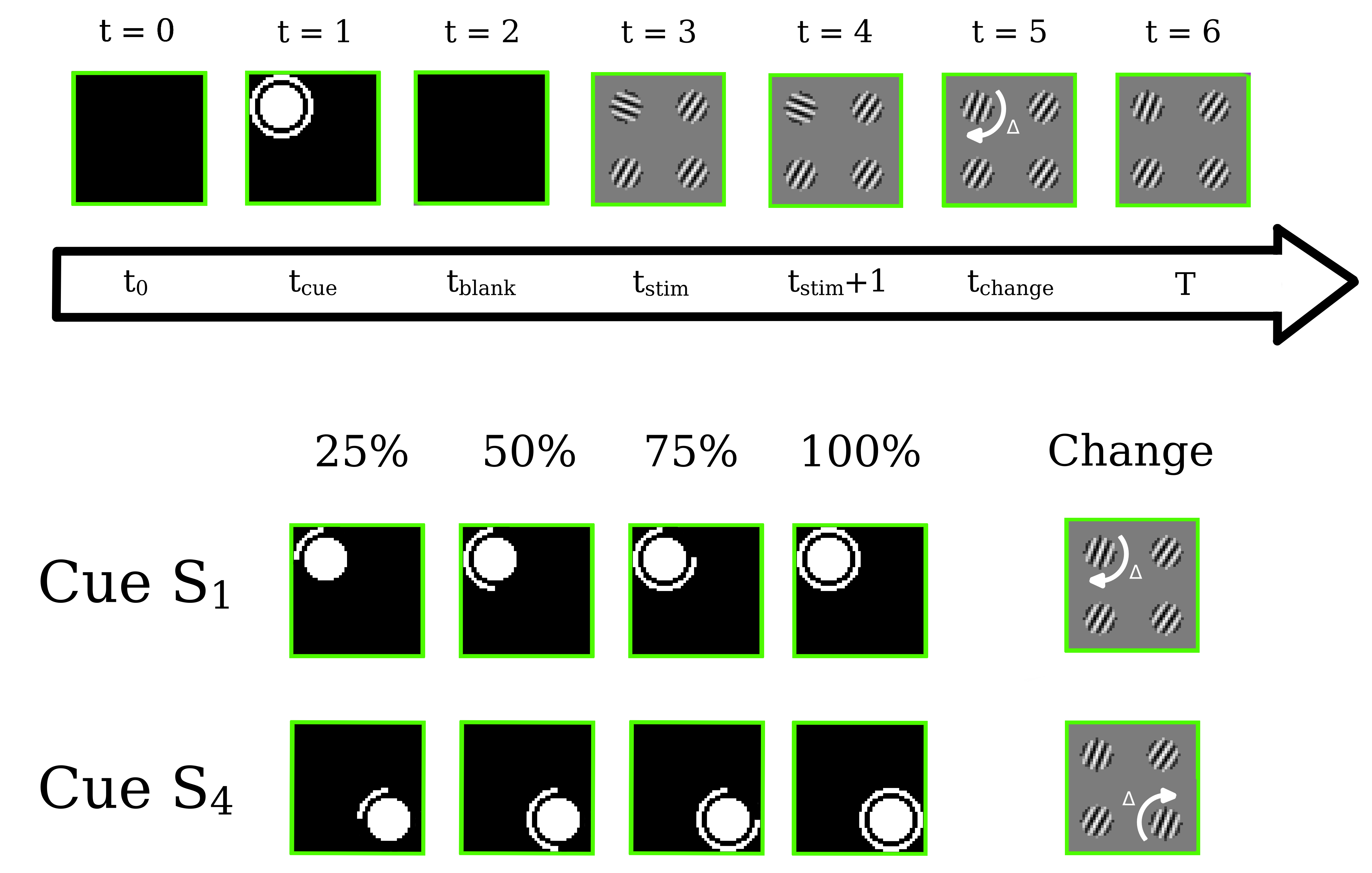}
    \caption{Each trial in the task comprises seven time steps. In each time step, a $50 \times 50$ grayscale image is input to the model. Black images are shown at $t=0$ and $t=2$. The cue is shown at $t=1$ and can either be at $S_1$ (top left) or $S_4$ (bottom right). The cue can take four configurations, where the portion of the circumference subtended by the ring around the center disk indicates the probability (25\%, 50\%, 75\%, or 100\%) that the change will appear at the cued location if the trial is a change trial.}
    \label{fig:environment}
\end{figure}

\section{Task and Model}

\subsection{Cued Orientation Change Detection Task Environment}

We trained our model on a spatially cued orientation change detection task similar to those used in primate neurophysiology labs \autoref{fig:environment}. Each trial comprised 7 time steps. At each time step, a 50x50 grayscale image was shown to the agent. At $t=0$, the trial began with a black image. At $t=1$, a spatial cue was displayed. The cue appeared at stimulus position 1 (Cue $S_1$) in one-half of the trials and at position 4 (Cue $S_4$) in the other half. At $t=2$, a black screen was displayed again. At $t=3$, stimulus onset occurred, displaying four ``Gabor" stimuli in randomly chosen orientations at fixed positions: top left ($S_1$), bottom left ($S_2$), top right ($S_3$), and bottom right ($S_4$). At $t=4$, the stimuli remained unchanged with the exception of orientation ``noise" added in each time step of stimulus presentation to control task difficulty (see Methods). If the trial was a no-change trial, stimuli remain unchanged from $t=4$ to $t=6$. In a change trial, at $t=5$ the orientation of one of the four stimuli changed by $\Delta$ degrees (where $\Delta$ varied from trial to trial). The orientations then remained unchanged from $t=5$ to $t=6$. Half of all trials were ``change trials" and the other half were ``no-change trials" (balanced across cue presentation positions).

Visually distinct cues indicated different levels of "cue validity": the probability of an orientation-change event occurring at the cued position. Cue validity levels were 25\%, 50\%, 75\%, or 100\%. For example, the 50\% valid cue presented at $S_1$ meant that, if this was a change trial, there was a 50\% probability that the orientation change would occur at position $S_1$. Cue validity was depicted visually by a white arc that subtended 25\%, 50\%, 75\% or 100\% of a central disc's circumference (\autoref{fig:environment}). We use the term ``cue validity" to align with the convention established in the human and NHP psychophysics literature ~\cite{posner1980attention, egly1994shifting, thomsen2005processing, brisson2008express}.

Much like NHPs trained in visual attention tasks, we trained our model in an RL setting. At each time step the agent could either choose to wait ($a^{(t)}=\text{\lq wait\rq}$) or declare a change ($a^{(t)}=\text{\lq declare change\rq}$). For $t<6$, waiting resulted in no reward ($r=0$), and the trial advanced to the next time step. At $t=6$, waiting rewarded the agent $r=1$ in a no-change trial and $r=0$ in a change trial. Declaring a change always ended the trial. When the agent correctly declared a change at $t \geq 5$, it received a reward $r=1$. In all other cases, declaring a change resulted in $r=0$. Before describing the model's behavior, we first briefly describe model components and their interconnections to facilitate interpretation of the model's attention dynamics.

\begin{figure}[!t]
    \includegraphics[width=0.99\linewidth]{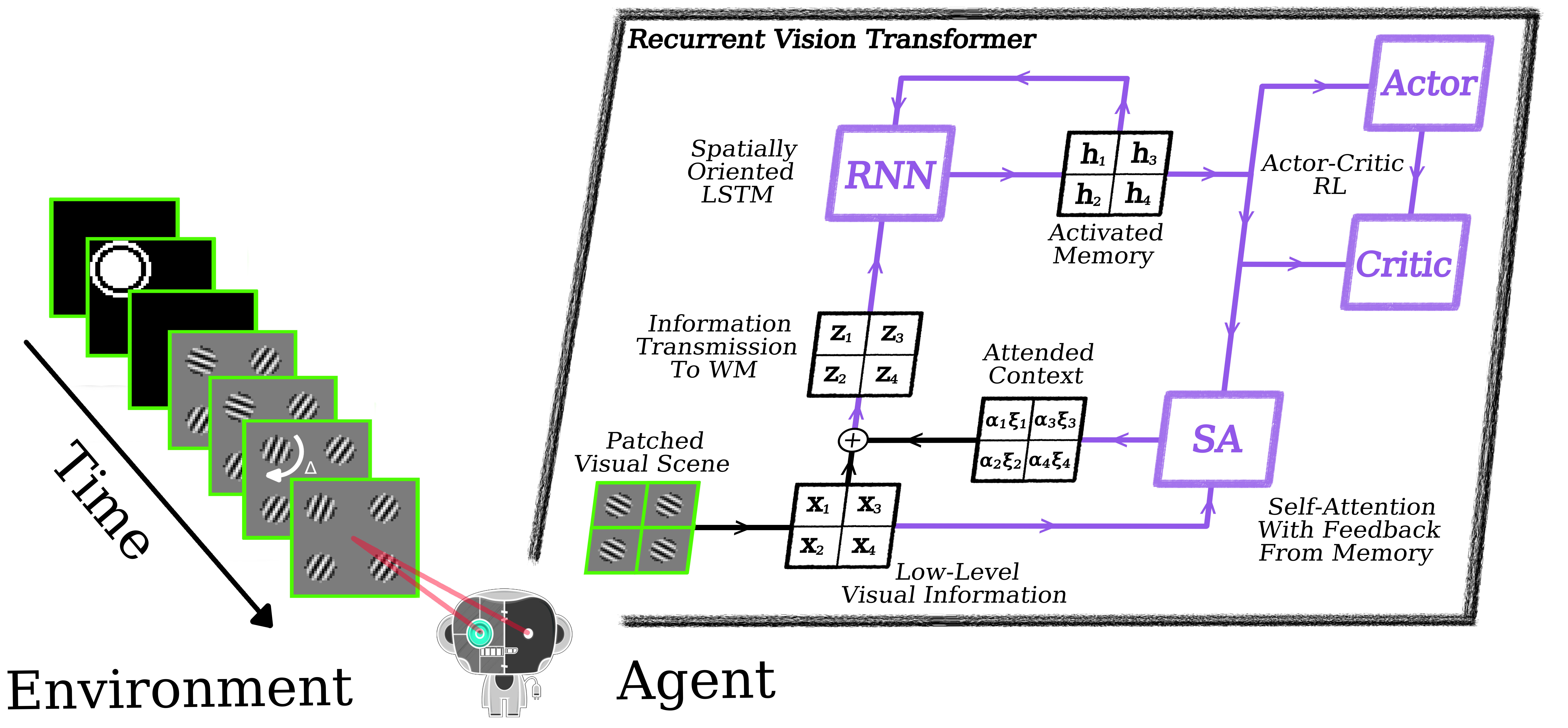}
    \caption{Model Schematic. At each timestep a single image is input, parsed into four patches, and passed through a pre-processing stage (see Methods). The resulting low-level visual features, \(X^{(t)} = \{x_i^{(t)}\}_{i=1}^{4}\), are combined with the activated memory, \(H^{(t-1)} = \{h_i^{(t-1)}\}_{i=1}^{4}\) in a self-attention mechanism, producing spatio-temporal context vectors (\(\alpha_i \xi_i\)). Context vectors are added to the low-level features and processed together to yield \(Z^{(t)} = \{z_i^{(t)}\}_{i=1}^{4}\). The memory is then updated \(C^{(t)} = \{c_i^{(t)}\}_{i=1}^{4}\) using both \(Z^{(t)}\) and the previous memory \(H^{(t-1)}\). The updated memory \(H^{(t)}\) is both fed back into the self-attention mechanism and forward to the RL Agent's actor and critic networks. The actor network uses \(H^{(t)}\) to select an action ($\text{\lq wait\rq}$ or $\text{\lq declare change\rq}$), while the critic network estimates upcoming cumulative rewards. Purple lines indicate weights updated by reward feedback.}
    \label{fig:model}
\end{figure}

\section{Model}

Our Recurrent ViT has three parts: the self-attention (SA) module, the patch-based long-short-term-memory (LSTM) module ("working memory" module), and an actor-critic RL agent. The environment generates the current visual scene and the agent converts this scene to a visual representation, $X^{(t)}$, through low-level convolutional operations (see Supplement). Processing results in 4 visual patches: $X^{(t)} = \{ x_i^{(t)} \}_{i=1}^{4}$, and this 4-patch structure is retained throughout the attention and working memory modules (number of patches is a hyperparameter). The primary benefit of our choice to structure the visual patches around stimulus positions is the interpretability it affords to the model's ``attention map". Specifically, it allows the attention map to be visualized as a 4x4 array in which we can interpret as the bias, $\alpha_i^{(t)}$, assigned to an internal representation ($\xi_i^{(t)}$) associated with stimulus $S_i$. An important distinction here is that $\alpha_i^{(t)}$ does not just describe the the attention on the current visual patch $x_i^{(t)}$. Instead it describes the attention on an internal representation, $\xi_i^{(t)}$ that consists of the immediate visual information in $x_i^{(t)}$ and information derived from activated memory describing relevant past temporal and spatial context, $h_i^{(t-1)}$.

The patch-based LSTM module receives a transmission, $Z^{(t)}=\{z_i^{(t)}\}_{i=1}^{4}$, that contains visual information derived from the immediate visual scene in addition to spatial and temporal context derived from the self-attention mechanism. This information is utilized to update the internal states of the LSTM, $C^{(t)}=\{c_i^{(t)}\}_{i=1}^{4}$ (see Methods). Activated memory, $H^{(t)}=\{h_i^{(t)}\}_{i=1}^{4}$, is derived from these internal states and sent: (1) recurrently back into the LSTM; (2) to the self-attention module; (3) to the actor and critic networks. This allows attention to be allocated both on the basis of visual inputs as in traditional transformer architectures ~\cite{vaswani2017attention, dosovitskiy2020image}, and on the basis of memory.

To understand which model architecture elements were required to recapitulate primate-like attention, we also tested several alternative models. We briefly describe alternative model performance after results from the Recurrent ViT. Full descriptions of all models can be found in the Methods section.

\begin{figure*}[!t]
    \includegraphics[width=0.99\linewidth]{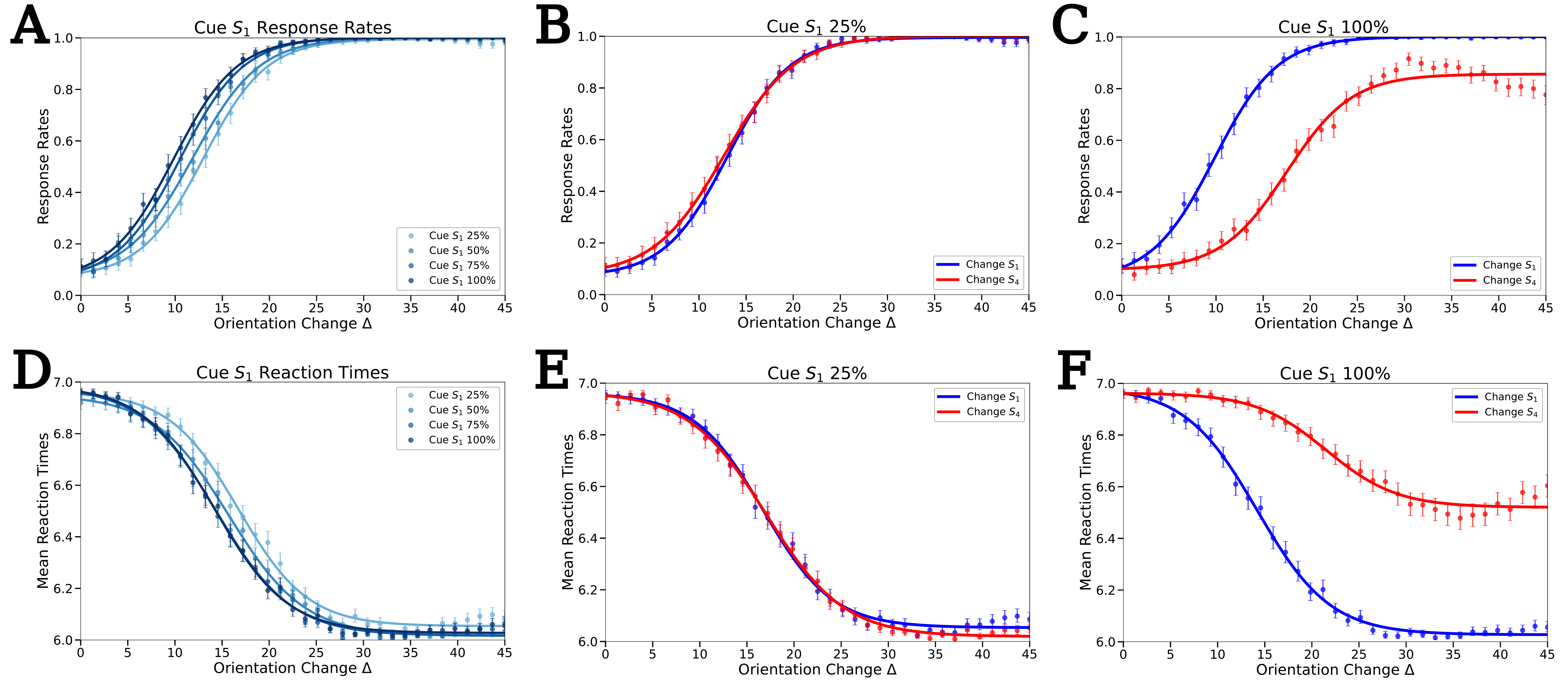}
    
    \caption{
        \textbf{A--F} shows the response rates (\textbf{A--C}) and reaction times (\textbf{D--F}) of our agent over varying cue validities with respect to the $S_1$ location and either a change on $S_1$ or a change on $S_4$ positions. Each data point was 500 trials where $\Delta$ specifies the magnitude of the orientation change. The response-rate was computed as $n_\text{dc}/n_\text{trials}$, where $n_\text{dc}$ is the total number of trials in which the agents selected the action $a^{(t)}=\text{``declare change''}$ and $n_{trials}$ is the total number of trials. The reaction times were computed as $1/500 \sum_i \tau_{i}$, where $\tau_{i}$ is the time the trial ended, either by the agent declaring a change or waiting through the final timestep. \textbf{A} Response rates over each possible cue condition w.r.t. the $S_1$ position where changes also occurred at the $S_1$ position. \textbf{B} Response rates computed over trials with a cue at the $S_1$ position comparing changes at the $S_1$ versus the $S_4$ locations. \textbf{C} Similar to \textbf{B} but with a 100 \% cue at the $S_1$ location. \textbf{D--F} Same conditions as \textbf{A--C} showing the mean reaction times.}
    \label{fig:behavior}
\end{figure*}

\section{Results}

\subsection{A recurrent ViT exhibits behavior signature of visual attention}

Our model exhibited orderly "psychometric" and "chronometric" functions with characteristic sigmoidal shapes commonly observed in human and NHP experiments (\autoref{fig:behavior}A,D). Larger orientation-change $\Delta$ values were associated with higher hit rates and shorter reaction times, qualitatively comparable to those seen in countless human and NHP psychophysics experiments. For cued orientation changes, higher cue validity improved correct response rate (\autoref{fig:behavior}A) and sped reaction time modestly (\autoref{fig:behavior}D). This pattern mirrors experimental findings that attentional benefits in biological systems are most pronounced when discriminating subtle changes ~\cite{lu1998sens}. No effects of cue validity were observed on the slope of the fitted psychometric function, the guess rate, or the lapse rate. These results indicate that, like spatial attention in biological visual systems, the attention mechanism of our model produced primarily additive effects on perceptual sensitivity rather than changing the shape of the psychometric function (Supplement)~\cite{lu1998sens, solomon2004sens, cameron2002sens}.


Contrasting performance for cued orientation changes compared to uncued revealed a clear ``cueing effect", again recapitulating results from human and NHP literature \cite{muller1987sensitivity, hawkins1990visual, carrasco2011visual, rust2022priority, saltzman1948reaction, carlson1983reaction, prinzmetal2005attention, jehu2015prioritizing}. Using the trained model with fixed weights, we were able to test how the model responded to uncued orientation changes even in the 100\% cue validity case, despite the necessary absence of uncued changes in trials with the 100\% valid cue during training (\autoref{fig:behavior}C,F). In that 100\% cue validity condition, for example, $10^\circ$ cued orientation changes were detected in roughly 50\% of trials, but uncued changes of the same magnitude were detected in roughly 15\% of trials (\autoref{fig:behavior}C). The magnitude of this cueing effect varied systematically with cue validity: compared to the 100\% validity condition, cue effects were mostly absent in the 25\% condition (\autoref{fig:behavior}B,C), which is expected - because there are 4 stimulus positions, 25\% cue validity indicates equal probability of an orientation change at any stimulus position. Overall, these findings confirm that our model robustly mirrors primate attention task behavior: it exhibits a strong cueing effect when spatial cues are highly predictive, and the effect diminishes as cue validity decreases.

\begin{figure*}[!t]
    \centering
    \vspace*{-0mm} 
    \includegraphics[width=0.99\linewidth]{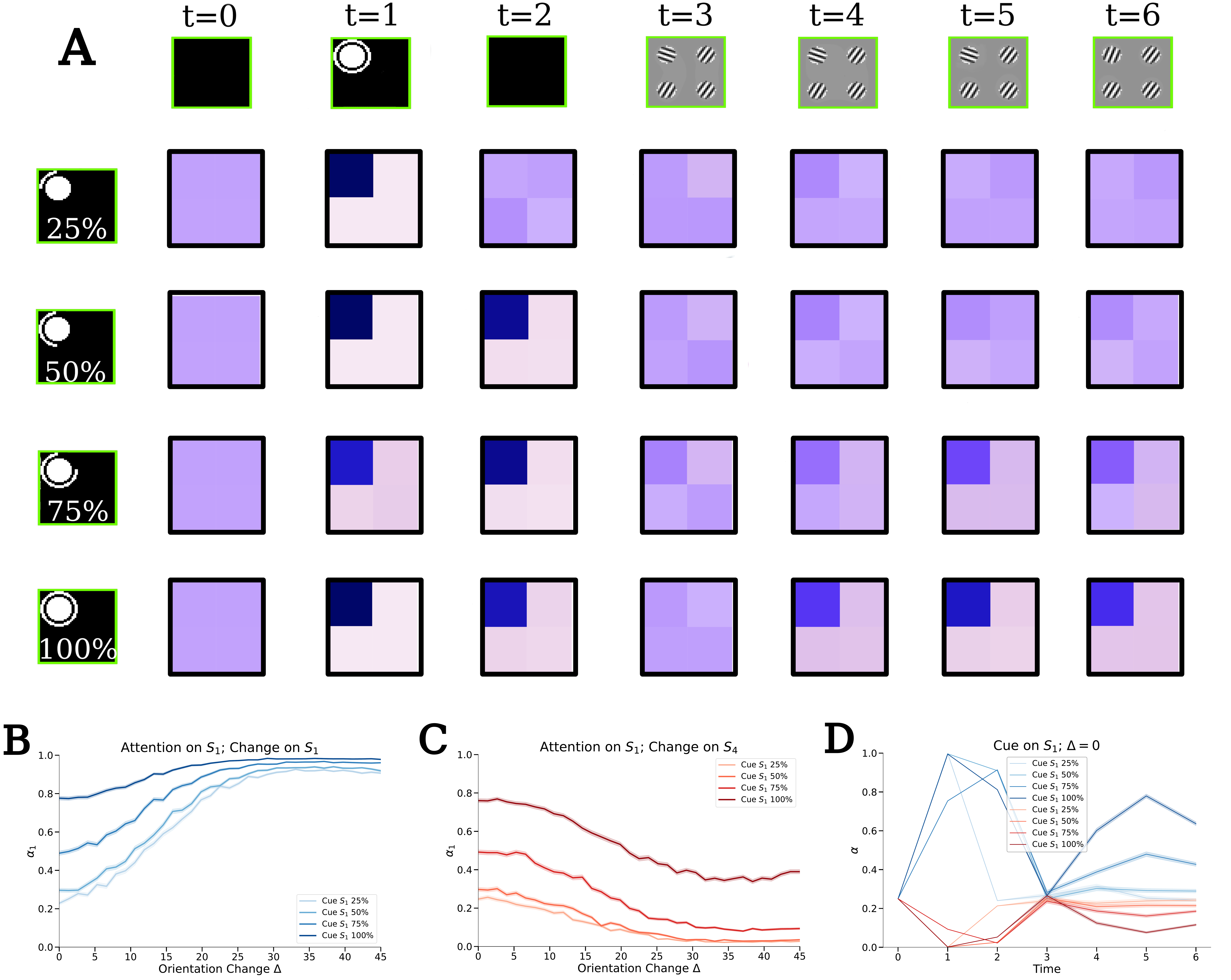}
    
    \caption{\textbf{A}~Averaged self-attention maps at each timestep when \(S_1\) is cued and orientation change \(\Delta=0\) (no-change trials). Rows correspond to different cue validities (25\%, 50\%, 75\%, 100\%), and columns to timesteps \(t=0,\ldots, 6\). Darker squares in each \(2\times2\) attention map reflect stronger attention. \textbf{B--C}~Attentional bias ($\alpha_1^{t_\text{change}}$) on $S_1$ as a function of orientation change $\Delta$ for each cue validity. The bias $\alpha_1^{(t)}$ is the top-left value from the heatmaps in \textbf{A}. \textbf{D}~Attentional bias on \(S_1\) (blue: $\alpha_1^{(t)}$) and \(S_4\) (red: $\alpha_4^{(t)}$) as a function of timestep when \(S_1\) is cued and orientation change \(\Delta=0\) (no-change trials).}
    \label{fig:attention}
\end{figure*}

\subsection{A recurrent ViT deploys attention in an human/NHP-like strategy}

To visualize how the model allocates attention in each timestep, we generated averaged self-attention heatmaps from trials with the cue at location $S_1$ and no orientation change occurred ($\Delta = 0$; \autoref{fig:attention}A). This established heatmap approach ~\cite{dosovitskiy2020image} for interpreting self-attention in vision transformers illustrates the relative attentional weight assigned to each stimulus location over time. At the start of each trial ($t=0$), the attention landscape was flat (\autoref{fig:attention}A). When the spatial cue appeared ($t=1$), attention became strongly biased toward $S_1$ ($\alpha_1^{(t=1)}\approx 1$). This bias persisted during the subsequent blank interval ($t=2$), indicating that the memory ($h_1^{(t)}$) can maintain attention allocation without visual input. At stimulus onset ($t=3$), attention maps became largely flat again. Just before the change timestep ($t=4$), the model began to allocate attention to the cued location, with stronger attention for higher cue validity. At the time of change ($t=5$), higher cue validity caused much stronger attention allocation to the cued location, and this bias persisted in the next time step ($t=6$). These findings show that the Recurrent ViT has learned to allocate attention both based on the behavioral relevance of the input (strongly attending to the appearance of the cue) and on reliable temporal associations (e.g. attending to the cued location in advance of the change time step).

Complementing spatial heatmaps, time-courses of attention allocation cued ($\alpha_1^{(t)}$) and uncued ($\alpha_4^{(t)}$) locations illustrate attention dynamics (\autoref{fig:attention}D). These time-courses clearly illustrate the absence of any difference between attention allocated to the cued vs. uncued locations at stimulus onset ($t=5$), the dependence of memory-based attention allocation before, during, and after the change on cue validity, and the reduction of attention allocated to $S_4$ accompanied by increased attention to $S_1$. 

Two features of the Recurrent ViT's attention allocation within and across timesteps closely resemble observations from human and NHP studies of attention. First, the lack of attention-related modulation of NHP neuronal activity at stimulus onset has been observed frequently (see for example ~\cite{herman2017colorchange, ghose2002temporal, ilaria2017v4gain, wang2015v1att, thompson1996fef}). Second, the allocation of attention to locations anticipated to contain behaviorally relevant visual information mirrors attention dynamics observed in humans and animals performing tasks in which sensory events occur at predictable times ~\cite{herman2017colorchange, ghose2002temporal, sharma2015temporal, jaramillo2011temporal, nobre2018temporal}. These parallels between our model's attention dynamics and those in primate brains suggest the Recurrent ViT has discovered attention deployment strategies that mirror those employed by primate visual systems.

Having found that input (e.g. cue onset) and memory interact in driving attention allocation, we were curious if there might be an interaction between orientation change magnitude $\Delta$ and cue-validity at the time of orientation change. We examined how attention allocated to the cued location varied with both change magnitude and cue validity, comparing trials where changes occurred at either the cued location (\autoref{fig:attention}B) or the opposing location (\autoref{fig:attention}C). When the change was at the cued location, $\alpha_1^{(t_\text{change})}$ increased sigmoidally with increasing $\Delta$ (\autoref{fig:attention}B). Large cued orientation changes "captured" attention regardless of cue validity but at smaller values attention allocation reflected cue validity more directly, consistent with the idea that cues primarily improve performance for subtle visual events. Conversely, when the change was at the uncued location, larger $\Delta$ values decreased attention allocated to the cued $S_1$ location ($\alpha_1^{(t_\text{change})}$) as attention was drawn to the uncued change location. There was also a more subtle interaction between uncued changes and cue validity: With 25\% cues, large uncued changes at $S_4$ drove $\alpha_1^{(t_\text{change})}$ near 0, indicating strong attentional capture by the uncued change. However, with increasing cue validity (\autoref{fig:attention}C), $\alpha_1^{(t_\text{change})}$ maintained progressively higher values even for large $\Delta$ at $S_4$, demonstrating that strongly predictive cues can partially maintain attention at the cued location in the face of competing visual events.

\subsection{Manipulating Bias Affects Response Rate and Reaction Times}

The ability to causally manipulate activity in primate brain regions like the frontal eye fields (FEF) and superior colliculus (SC) has provided foundational insights into the neuronal mechanisms of attention ~\cite{moore2003selective, cavanaugh2004subcortical, cavanaugh2006enhanced, mirpour2010microstimulation, bollimunta2018fefsc, monosov2011fef, zenon2012attention, herman2018midbrain}. The clear interpretability of our model’s spatial and temporal attention allocation dynamics offers a unique opportunity to test whether targeted perturbations of its self-attention mechanism produce effects analogous to these biological interventions. Demonstrating such parallels would provide a stringent validation that our Recurrent ViT captures not only correlational but also causal principles underlying primate attentional control.

Again resembling the results of NHP experiments, the behavioral consequences of attention manipulation were only observed when they were applied at the time of the change - manipulating attention at the time of cue presentation had minimal effects. In Supplemental Figure X, we systematically explore the effects of manipulating attention during cue presentation versus during the change event, and quantify those effects using signal detection theory's sensitivity ($d'$) and criterion. Increasing attention to the cued location during cue presentation ($\alpha_1^{(t_{\text{cue}})}$) had minimal impact on either sensitivity ($d'$) or criterion (Supplemental Figure A,D). The lack of an effect from manipulations at the time of the cue but presence of those effects from manipulations at the time of the change closely mirror findings from primate SC microstimulation experiments ~\cite{cavanaugh2006enhanced}. Thus, precisely timed attention perturbations reveal causal parallels to NHP microstimulation data, extending beyond simple correlation.

Several past works have proposed that attention manipulations resulting from perturbation of specific nodes in the primate brain might selectively influence ($d'$) or criterion ~\cite{sridharan2017does, luo2018attentional}. However, our model suggests a more nuanced reality - manipulating attention at the time of change produces complex, interrelated effects on both sensitivity and criterion that depend on cue validity and change location (Supplemental Figure 16). This finding highlights how attempting to assign distinct signal detection theory metrics to specific neural circuits may artificially compartmentalize what is fundamentally an integrated process. The joint modulation of $d'$ and criterion in our model emerges naturally from manipulating a single computational mechanism, suggesting that clean dissociations between these measures may not reflect the underlying neural implementation of attention. 

\begin{figure*}[!t]
    \centering
    \vspace*{-0mm} 
    \includegraphics[width=0.99\linewidth]{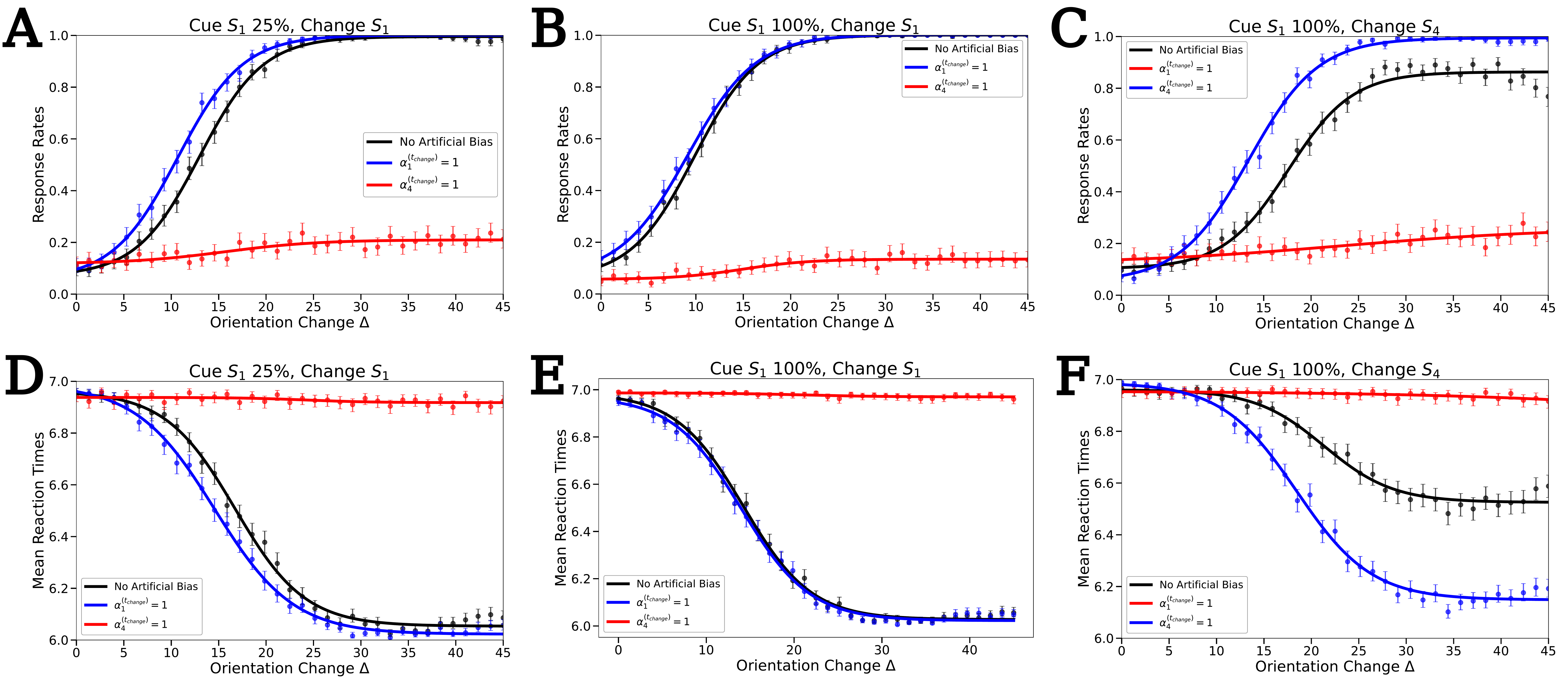}
    \caption{Plots showing the effect of artificially modulating the bias. All data points are the result of an average over 500 trials. Artificial modulation involves inducing a high bias in a single spatial region (increasing the value of one of the patches in the self-attention maps from Figure~\ref{fig:attention}G). In all cases, the bias is induced with respect to the $S_1$ ($\alpha_1^{(t)}$) or the $S_4$ ($\alpha_4^{(t)}$) stimulus location. In \textbf{A--F}, we plot the response rates and reaction times versus the orientation change $\Delta$. If $\alpha_i^{(t_{\text{change}})}=1$, this indicates that the transmission $Z^{(t_{\text{change}})}$ has been completely biased toward $\xi_i^{(t_{\text{change}})}$. In \textbf{A--C} we show the effects of this manipulation on the response rates, and then again for the reaction times in \textbf{D--F}.}
    \label{fig:2}
\end{figure*}

\subsection{Alternative Architectures and Training Approaches Fail to Capture Primate-like Attention}

To validate our architectural choices and training approach, we systematically evaluated several alternative model variants. We tested different memory-attention integration schemes (tokens, additive, and multiplicative feedback) and we examined whether reinforcement learning was necessary by training two supervised variants of our architecture. Supervised models trained either on trial-type labels (change / no-change) or target action sequences achieved reasonable task performance but did not show a "cueing effect", instead detecting orientation changes similarly regardless of cue validity (\autoref{tab:model_comparison}). Alternative memory-attention architectures showed similar limitations - while the additive attention model demonstrated a weak version of anticipatory attention reallocation, both it and the token-based model failed to capture the rich temporal dynamics observed in primates. Only our RL-trained Recurrent ViT with multiplicative feedback produced the full-compliment of primate-like features we have documented above. These results suggest that both reinforcement learning and multiplicative interactions between memory and attention are critical for developing temporally structured attentional control that mirrors primate behavior.
\begin{table*}[b]
\caption{Comparison of model variants}
\centering
\renewcommand{\arraystretch}{1.2} 
\begin{tabularx}{\textwidth}{
  >{\raggedright\arraybackslash}p{2.5cm} | 
  >{\raggedright\arraybackslash}p{3.0cm} | 
  >{\raggedright\arraybackslash}p{3.7cm} | 
  >{\raggedright\arraybackslash}p{3.5cm} | 
  >{\raggedright\arraybackslash}p{4.2cm}}
\toprule
\textbf{Model} & \textbf{Learning Source} & \textbf{Causal Perturbation Effects} & \textbf{Cueing Effect} & \textbf{Attention Dynamics} \\
\midrule
Recurrent ViT & Reward feedback from change detection & Strong modulation of behavior matching microstimulation effects & Maintains selective responding even at large $\Delta$ & Strong cue attention, broad monitoring, anticipatory reallocation \\
\midrule
Memory as Tokens & Reward feedback from change detection & Weak modulation of behavior & Weak separation between cue validities & No clear temporal structure \\
\midrule
Additive Attention & Reward feedback from change detection & Moderate modulation of behavior & Moderate separation between cue validities & Weak version of anticipatory pattern \\
\midrule
Supervised Beliefs & Binary trial-type labels & Minimal effect on behavior & Responds to all large changes regardless of cue & Weak, unstructured attention allocation \\
\midrule
Supervised Actions & Target action sequences & Minimal effect on behavior & Responds to all large changes regardless of cue & Weak, unstructured attention allocation \\
\bottomrule
\end{tabularx}
\label{tab:model_comparison}
\end{table*}

\section{Discussion}
In this work, we introduce a Recurrent Vision Transformer (Recurrent ViT) enhanced with a spatial memory module designed for a cued orientation change-detection task. Our central goal was to determine whether augmenting standard vision transformers—which typically rely on feedforward processing of single frames \cite{dosovitskiy2020image}—with a recurrent feedback mechanism can enable top-down, internally guided attentional control akin to that observed in human and non-human primate (NHP) vision \cite{mehrani2023self,baluch2011mechanisms,carrasco2011visual}. Our experiments demonstrate that the proposed Recurrent ViT successfully recapitulates many hallmark effects of primate visual attention.

\subsection{Recovering Hallmark Signatures of Primate Attention.}

First, our trained model shows improved performance and faster detection of orientation changes at cued locations, mirroring the well-documented behavioral effects of selective spatial attention \cite{carrasco2011visual,bisley2010lip,rust2022priority,hoffman2016visual}. These benefits emerge in situations where high-validity cues bias internal representations toward the cued location, but they taper off or reverse if competing salience signals (e.g., a large orientation change elsewhere) dominate the model’s self-attention. This interplay between cue validity and exogenous salience resonates with empirical observations that attentional allocation reflects both top-down predictions and bottom-up feature-driven signals \cite{baluch2011mechanisms,knudsen2007fundamental}. In standard feedforward ViTs, attention is inherently limited to correlational or grouping-based processes \cite{mehrani2023self,lamy2006grouping}, whereas our recurrent module explicitly integrates memories of cue identity, location, and temporal context—restoring top-down selectivity typically absent in off-the-shelf architectures.

\subsection{Relevance to Neural Mechanisms of Attention and Working Memory.}

The success of our Recurrent ViT underscores the deep links between attention and working memory reported in neuroscience \cite{awh2006interactions,kiyonaga2013working,bahle2018architecture,panichello2021attvwm}. Much like the “attentional template” theory, which proposes that memory representations guide attention to relevant features and locations \cite{cameron2002sens,wheeler2002binding,botta2014spatial}, our model maintains a set of spatial codes over time. These memory states re-enter a self-attention module to bias ongoing visual processing, effectively bridging top-down and bottom-up circuits \cite{desimone1995neural,moore2003selective,cavanaugh2006enhanced,krauzlis2013superior}. Correspondingly, the interplay between memory and perception in our model echoes the reciprocal loops seen in primate frontal, parietal, and subcortical structures, where neural firing maintains spatial priority during blank intervals and facilitates rapid reactivation at anticipated moments of stimulus change \cite{silver2005topographic,huda2020distinct,srinath2021attention}.

\subsection{Subcortical and Dopaminergic Influences.}
Although our model already uses reward feedback to guide learning, we have not explicitly integrated dopaminergic-like prediction error signals or examined how reward history might adaptively modulate attentional policies in the superior colliculus and related circuits \cite{bolton2015diencephalic,pradel2021superior,essig2016warning,perez2017direct}. In biological systems, dopamine critically mediates plasticity, enabling more nuanced shaping of attentional priorities over extended time scales \cite{hikosaka2006basal,hickey2010reward,failing2018selection}. A potential future extension is to incorporate a free-energy principle-inspired, unsupervised component \cite{friston2006free,feldman2010attention, friston2012dopamine,khezri2022free,mazzaglia2022free}, which could allow the model to learn latent, generative structure in its environment—paralleling how dopamine modulates not only immediate reward but also uncertainty and exploration in real brains. Such a framework would unify reward-driven reinforcement learning with a broader predictive coding approach, further enhancing the model’s capacity for dynamic, context-sensitive attention. 

\subsection{Constraints, Biological Plausibility, and Interpretability.}
A central feature of our model is the introduction of a recurrent spatial memory component that constrains information flow between image frames shown at different points in time. Unlike standard transformers—which can re-attend to entire sequences without constraint \cite{vaswani2017attention,dosovitskiy2020image,hassanin2024visattmod} or models that process full sequences of images \cite{bardes2023v}—our approach assigns each spatial patch of an immediate image to a single hidden-state slot within an LSTM. This imposed bottleneck encourages competition among representations, echoing the biased competition framework which posits that a finite “priority map” mediates interactions between bottom-up inputs and top-down influences \cite{desimone1995neural,reynolds1999competitive,bisley2019neural,wolfe2021guided,rust2022priority}. Although this capacity-limited design is consistent with psychophysical findings on the limited nature of visual working memory \cite{luck1997capacity,luck2013visual,brady2013probabilistic,emrich2017attention} and helps ensure that primarily task-relevant information is retained \cite{desimone1996neural,van2009limits,lee2015encoding}, it represents just one plausible mechanism among several. In contrast, other models may integrate information over time without such strict constraints, potentially capturing different aspects of attentional processing. Our approach, therefore, offers a balanced compromise that mirrors key behavioral observations while providing a tractable, interpretable framework for studying top-down attention.

Furthermore, our framework offers a degree of interpretability by linking each attention weight to both an immediate visual patch and its corresponding memorized representation. This mapping holds promise for developing in silico experiments that could approximate, in a controlled manner, the effects observed in microstimulation or lesion studies in non-human primates \cite{cavanaugh2006enhanced,mirpour2010microstimulation}. In our preliminary experiments, targeted adjustments of self-attention weights led to systematic changes in detection rates and reaction times. While further validation is required, this controlled perturbation approach may serve as a valuable tool for exploring causal relationships between attention and working memory in computational models \cite{moore2003selective,ruff2016attention,herman2020attention}.

\subsection{Broader Implications and Future Directions.}
This Recurrent ViT opens several avenues for future research. First, richer scenarios—such as multi-object tracking \cite{bettencourt2009effects,meyerhoff2017studying}, dynamic scene understanding \cite{wolfe2011visual,lamy2006grouping}, or tasks requiring mid-trial updates to memorized stimuli \cite{tas2016relationship,gresch2024shifting}—would extend our approach and further test its alignment with primate attentional performance. Incorporating saccadic eye movements, akin to real-world visual search, could allow the model to learn optimal covert and overt strategies in tandem \cite{krauzlis2013superior,gupta2024presaccadic}. Additionally, scaling up to deeper, multilayer recurrent architectures may capture the intricate, multi-level feedback loops characteristic of the primate cortex \cite{felleman1991distributed,kietzmann2019recurrence,zhuang2021unsupervised,khan2022transformers}.

Second, bridging our Recurrent ViT with reinforcement learning frameworks that incorporate explicit dopamine-like signals \cite{botvinick2020deep,babayan2018belief,hikosaka2006basal} could elucidate how value-based attentional modulation emerges in tandem with memory demands. This expansion would dovetail with broader theories of attention, memory, and decision-making as components of a common, computationally grounded process \cite{knudsen2007fundamental,monosov2020outcome}. Finally, the interpretability of our approach may inspire future “virtual lesion” or “virtual microstimulation” studies to dissect precisely how feedback, gating, and local competition produce emergent attentional dynamics—a goal shared across computational neuroscience and AI \cite{mante2013context, krauzlis2013superior, gattass2014effect,miconi2016feedback, liu2024human, cartella2024trends}.

\subsection{Conclusion.}
Taken together, these findings demonstrate that recapitulating human/NHP-like attention in transformer architectures is possible by introducing explicit top-down influences and recurrent feedback. Our Recurrent ViT significantly narrows the gap between standard, feedforward models of attention \cite{vaswani2017attention,dosovitskiy2020image} and the iterative, memory-intensive processes that characterize primate visual cognition \cite{rust2022priority,panichello2021attvwm}. By unifying principles of biased competition, working memory, and reward-driven learning within a single framework, we not only advance the biological plausibility of deep vision models but also generate a versatile platform for addressing fundamental questions about how perception, memory, and attention converge to guide adaptive behavior.

\section{Methods}

\subsection{Model Overview}

The objective of the model is to utilize immediate visual inputs in order to update an internal state with sufficient immediate and past visual information such that downstream decoders can estimate value and take action. We utilize a self-attention (SA) mechanism to construct the visual percept used to update an internal state of an RNN. Self-attention is computed based on the immediate visual inputs and feedback from the RNN. The following sections will describe the motivations and details of this process in more depth.

\subsection{Pre-Processing, Content Selection, and Construction for Visual Working Memory}
Given a visual scene (an image) of dimension $H \times W \times C$, denoted by $\mathbf{\mathcal{O}}^{(t)} \in \mathbb{R}^{H \times W \times C}$ at time $t$, the agent views the entire scene through a fixation at center field. During preprocessing, the image is partitioned into a set of patches, $\{\mathbf{{o}_i}^{(t)}\}_{i=1}^{n_{patch}}$, where each patch is of size $H_{patch} \times W_{patch} \times C_{patch}$. The original visual patches are then transformed into a compact set of internal representations, $\{\mathbf{x_i}^{(t)}\}_{i=1}^{n_{patch}}$, each $\mathbf{x_i}^{(t)} \in \mathbb{R}^{H_{patch} \times W_{patch} \times C_{patch}}$ and typically satisfying
\[
    \dim(\mathbf{x_i}^{(t)}) \ll \dim\bigl(\mathbf{o_i}^{(t)}\bigr).
\]
Together, the collection of these feature patches forms the immediate visual information available to the agent at time $t$, denoted by $\mathbf{X}^{(t)} = \{\mathbf{x_i}^{(t)}\}_{i=1}^{n_{patch}}$.

\subsection{Spatially Oriented Visual Working Memory}

Following numerous experimental findings, we allow our model to maintain a spatially arranged visual working memory \cite{wheeler2002binding, pertzov2014privileged,schneegans2017neural, van2019human}, in which each patch location $i$ has a corresponding patched memory component $\mathbf{c_i}^{(t)} \in \mathbb{R}^{d_{mem}}$ within the RNN. For updating the internal state of our RNN, we utilize the operations and functions described in the LSTM architecture \cite{hochreiter1997long, beck2024xlstm}. For the remainder of our description, we will refer to the collection $C^{(t)}=\{c_i^{(t)}\}_{i=1}^{n_{patch}}$ the VWM state and $c_i^{(t)}$ a VWM patch, where
\begin{equation}
    \label{eq:C}
    \mathbf{c_i}^{(t)} =  \mathbf{f_i}^{(t)} \odot \mathbf{c_i}^{(t-1)} + \mathbf{u_i}^{(t)} \odot \boldsymbol{\psi_i}^{(t)},
\end{equation}
where 
\begin{align*}
\mathbf{f_i}^{(t)} &= F\!\bigl(\mathbf{x_i}^{(t)},\, \mathbf{h_i}^{(t-1)}\bigr), \\
\mathbf{u_i}^{(t)} &= U\!\bigl(\mathbf{x_i}^{(t)},\, \mathbf{h_i}^{(t-1)}\bigr), \\
\boldsymbol{\psi_i}^{(t)} &= \Psi\!\bigl(\mathbf{x_i}^{(t)},\, \mathbf{h_i}^{(t-1)}\bigr).
\end{align*}
The operator $\odot$ denotes elementwise multiplication. Here, $\mathbf{f_i}^{(t)} \in [0,1]^{\,d_{mem}}$ determines which parts of $\mathbf{c_i}^{(t-1)}$ are \textit{forgotten} (i.e., decayed), while $\mathbf{u_i}^{(t)} \in [-1,1]^{\,d_{mem}}$ and $\boldsymbol{\psi_i}^{(t)} \in \mathbb{R}^{d_{mem}}$ selectively modulate and propose new content. Altogether, these operations enable dynamic insertion, maintenance, and forgetting of information in $\mathbf{c_i}^{(t)}$. 

The activated memory patch, $h_i^{(t)}$ is constructed from the VWM patch, $c_i^{(t)}$:
\[
    \mathbf{h_i}^{(t)} = \boldsymbol{\phi_i}^{(t)} \odot \Bigl(\tfrac{\mathbf{c_i}^{(t)}}{\mathbf{n_i}^{(t)}}\Bigr),
\]
where $\boldsymbol{\phi_i}^{(t)} = \Phi\!\bigl(\mathbf{x_i}^{(t)}, \mathbf{h_i}^{(t-1)}\bigr) \in [0,1]^{\,d_{mem}}$ selects elements of $\mathbf{c_i}^{(t)}$ for downstream processing, and $\mathbf{n_i}^{(t)}$ is a normalization term. Each function $F, U, \Psi, \Phi$ is parameterized by a feedforward neural network that receives $\mathbf{x_i}^{(t)}$ and $\mathbf{h_i}^{(t-1)}$ as inputs. Although the parameters of these functions remain fixed once trained, the recurrent operation through $\mathbf{h_i}^{(t)}$ allows the system to track temporal dynamics.

\subsection{A Disjoint Memory}
Following the ideas presented by Knudsen \cite{knudsen2007fundamental}, the activated subset of working memeory is central for decision-making and planning. However, a patched RNN as described above presents a clear shortcoming: each $\mathbf{h_i}^{(t)}$ encodes the content of its own patch independently, without explicit awareness of neighboring patches. If downstream processes (\emph{e.g.}, a decoder $\pi$) require spatial or contextual relationships among patches, they must construct these relationships entirely from the population of VWM patches. Moreover, the problem intensifies over time. If the network must integrate information from patches across multiple timesteps (e.g., $\mathbf{x_i}^{(\tau)}$ and $\mathbf{x_j}^{(\tau)}$ for $\tau < t$), then each VWM patch must \emph{retain} all potentially significant current and past features useful for task-relevant decoding by downstream networks. This approach quickly becomes intractable, as it demands that the architecture store a large number of unique \emph{conjunctions} of spatio-temporal features. This challenge aligns with the combinatorial explosion recognized by Tsotsos \cite{tsotsos1988complexity} as a core difficulty in perceptual organization. 

To circumvent these limitations, we introduce self-attention into the encoding process, encouraging each patch’s representation to reflect the context provided by the other patches within the same timestep. In doing so, we create a spatially integrated or \emph{context-aware} activated memory before the information even updates the VWM.

\subsection{Self-Attention and Spatially Aware activated memory}

We wish to obtain a scene-level representation $\mathbf{Z}^{(t)} = \{\mathbf{z_i}^{(t)}\}_{i=1}^{n_{patch}}$ such that each $\mathbf{z_i}^{(t)}$ encodes the task-relevant spatial relationships among the visual feature patches $\mathbf{x_1}^{(t)}, \dots, \mathbf{x_{n_{patch}}}^{(t)}$. By doing so, the patch-based LSTM will be able to utilize immediate visual information within a patch and task-relevant spatial context to update internal states. Formally, we want
\[
\mathbf{z_i}^{(t)} = f_z\Bigl(\mathbf{x_i}^{(t)}, \{\mathbf{x_j}^{(t)}\}_{j\neq i}\Bigr),
\]
A straightforward way to implement this is via self-attention:
\begin{equation}
    \label{eq:Z}
    \mathbf{z_i}^{(t)} = \mathbf{x_i}^{(t)} + \sum_{j=1}^{n_{patch}} a_{ij}^{(t)}\, \mathbf{v_j}^{(t)},
\end{equation}
where $\mathbf{v_j}^{(t)} = V\!\bigl(\mathbf{x_j}^{(t)}\bigr)$ is a function that maps the feature patch into a latent space and $a_{ij}^{(t)} = A\bigl(\mathbf{x_i}^{(t)}, \mathbf{x_j}^{(t)}\bigr)$ indicates the \emph{relative importance} of $\mathbf{x_j}^{(t)}$ with respect to $\mathbf{x_i}^{(t)}$. To ensure a proper probability-like weighting, we impose $\sum_{j=1}^{n_{patch}} a_{ij}^{(t)} = 1$ with $a_{ij}^{(t)} \in (0,1)$. A typical choice for $a_{ij}^{(t)}$ is:
\[
    a_{ij}^{(t)} \;=\; \frac{\exp\Bigl(\bigl\langle \mathbf{q_i}^{(t)}, \mathbf{k_j}^{(t)}\bigr\rangle\Bigr)}
    {\sum_{m=1}^{n_{patch}} \exp\Bigl(\bigl\langle \mathbf{q_i}^{(t)}, \mathbf{k_m}^{(t)}\bigr\rangle\Bigr)},
\]
where $\mathbf{q_i}^{(t)} = Q\!\bigl(\mathbf{x_i}^{(t)}\bigr)$ and $\mathbf{k_j}^{(t)} = K\!\bigl(\mathbf{x_j}^{(t)}\bigr)$ are \emph{query} and \emph{key} functions, respectively. Interpreting $a_{ij}^{(t)}$ as a salient feature map has strong parallels to the saliency map hypothesis \cite{koch1984selecting}; however, we adopt a \emph{winner-takes-most} approach rather than a strict winner-takes-all (WTA), common in many self-attention applications. In principle, should $a_{i,j^*}^{(t)} \approx 1$ for some $j^*$ and $a_{i,m}^{(t)} \approx 0$ for $m \neq j^*$, we recover WTA-like mechanism.

After computing $\mathbf{Z}^{(t)}$ for the entire scene, the visual percept patches, $z_i^{(t)}$, are used to update the VWM patches:
\[
    \mathbf{c_i}^{(t)} \;=\;  \mathbf{f_i}^{(t)} \odot \mathbf{c_i}^{(t-1)} \;+\; \mathbf{u_i}^{(t)} \odot \boldsymbol{\psi_i}^{(t)},
\]
where each function is now evaluated using $\mathbf{z_i}^{(t)}$ rather than the immediate visual scene patch in isolation, $\mathbf{x_i}^{(t)}$. This approach solves the \emph{spatial} integration problem in the current timestep. Yet, any feature with contextual importance \emph{across} timesteps remains challenging: we still need a mechanism to capture top-down feedback or \emph{memory-based} salience.

\subsection{Recurrent Feedback From Memory}

Knudsen \cite{knudsen2007fundamental} describes a feedback loop in which working memory provides top-down signals that bias neural representations relevant to the organism’s current goals. In the context of the biased competition model \cite{desimone1995neural}, working memory holds an \emph{attentional template} that biases competition in favor of task-relevant representations. However, in practice it is not clear how this mnemonic feedback is/should be implemented. In this we simplify (and constrain) the problem to implementing recurrent feedback from the patch-based LSTM to the self-attention mechanism of the ViT. Hence, we evaluate three different methods in terms of their ability to yield primate-like behavior signatures of attention. We call the vision transformer with mnemonic feedback the recurrent ViT.   

\subsection{Mnemonic Guidance}

\subsubsection{Visual working memory as tokens}

The first recurrent feedback method we evaluate is one in which we concatenate the mnemonic percept to the visual input. Thus, the input to the self-attention mechanism is
\[
\mathbf{\tilde{X}} = Concatenate[\mathbf{X}^{(t)},\mathbf{H}^{(t-1)}]
\]
where $\mathbf{\tilde{X}}\in\mathbb{R}^{2n_{patch},d_{model}}$. From here we define:
\begin{align*}
\mathbf{q}_{\tilde{X},i}^{(t)} &= Q_{\tilde{X}}\bigl(\mathbf{\tilde{x}}_i^{(t)}\bigr), \\
\mathbf{k}_{\tilde{X},j}^{(t)} &= K_{\tilde{X}}\bigl(\mathbf{\tilde{x}}_j^{(t)}\bigr), \\
\mathbf{v}_{\tilde{X},j}^{(t)} &= V_{\tilde{X}}\bigl(\mathbf{\tilde{x}}_j^{(t)}\bigr),
\end{align*}

The attention weights are given by
\begin{equation}
\label{eq:add_alpha}
\alpha_{i,j}^{(t)} \;=\;
\frac{\exp\Bigl(\langle \mathbf{q}_{\tilde{X},i}^{(t)} ,\;
\mathbf{k}_{\tilde{X},j}^{(t)}  \rangle\Bigr)}
{\sum_{m=1}^{n_{\text{patch}}}
\exp\Bigl(\langle \mathbf{q}_{\tilde{X},i}^{(t)} ,\;
\mathbf{k}_{\tilde{X},m}^{(t)} \rangle\Bigr)}.
\end{equation}
We then compute the output representation as
\begin{equation}
\label{eq:add_z}
\mathbf{z}_i^{(t)} \;=\; \mathbf{x}_i^{(t)}
\;+\; \sum_{j=1}^{2n_{\text{patch}}}
\alpha_{i,j}^{(t)} \mathbf{v}_{\tilde{X},j}^{(t)} .
\end{equation}
Here, we only take the first $n_{patch}$ entries $\{\mathbf{z}^{(t)}_{i}\}_{i=1}^{n_{patch}}$. The reason for this is because there are only $n_{patch}$ recurrent states in the patch-based LSTM, and $Z^{(t)}=\{\mathbf{z}^{(t)}\}_{i=1}^{n_{patch}}$ is only used as the input to the LSTM. 

\subsubsection{Additive Feedback from Visual Working Memory}

We split the standard self-attention operation into two parallel pathways: one for the bottom-up immediate visual inputs, \(\mathbf{x}_i^{(t)}\), and one for the top-down mnemonic inputs, \(\mathbf{h}_i^{(t)}\). Define:
\begin{align*}
\mathbf{q}_{X,i}^{(t)} &= Q_X\bigl(\mathbf{x}_i^{(t)}\bigr), \\
\mathbf{k}_{X,j}^{(t)} &= K_X\bigl(\mathbf{x}_j^{(t)}\bigr), \\
\mathbf{v}_{X,j}^{(t)} &= V_X\bigl(\mathbf{x}_j^{(t)}\bigr).
\end{align*}
and
\begin{align*}
\mathbf{q}_{H,i}^{(t)} &= Q_H\bigl(\mathbf{h}_i^{(t)}\bigr), \\
\mathbf{k}_{H,j}^{(t)} &= K_H\bigl(\mathbf{h}_j^{(t)}\bigr), \\
\mathbf{v}_{H,j}^{(t)} &= V_H\bigl(\mathbf{h}_j^{(t)}\bigr).
\end{align*}
The attention weights are given by
\begin{equation}
\label{eq:add_alpha}
\alpha_{i,j}^{(t)} \;=\;
\frac{\exp\Bigl(\langle \mathbf{q}_{X,i}^{(t)} + \mathbf{q}_{H,i}^{(t)},\;
\mathbf{k}_{X,j}^{(t)} + \mathbf{k}_{H,j}^{(t)} \rangle\Bigr)}
{\sum_{m=1}^{n_{\text{patch}}}
\exp\Bigl(\langle \mathbf{q}_{X,i}^{(t)} + \mathbf{q}_{H,i}^{(t)},\;
\mathbf{k}_{X,m}^{(t)} + \mathbf{k}_{H,m}^{(t)} \rangle\Bigr)}.
\end{equation}
We then compute the output representation as
\begin{equation}
\label{eq:add_z}
\mathbf{z}_i^{(t)} \;=\; \mathbf{x}_i^{(t)}
\;+\; \sum_{j=1}^{n_{\text{patch}}}
\alpha_{i,j}^{(t)} \Bigl(\mathbf{v}_{X,j}^{(t)} \;+\; \mathbf{v}_{H,j}^{(t)}\Bigr).
\end{equation}
In this additive design, features from the visual inputs and the mnemonic percept patches \(\{\mathbf{h}_i^{(t)}\}\) both contribute to the self-attention mechanism by modifying the inner product in the numerator of~\eqref{eq:add_alpha} and by merging the corresponding values in~\eqref{eq:add_z}. 

\subsubsection{Multiplicative Feedback from Visual Working Memory}

To incorporate multiplicative feedback, we instead define:
\begin{equation}
\label{eq:mult_alpha}
\alpha_{i,j}^{(t)} \;=\; 
\frac{\exp\Bigl(\langle \mathbf{q}_{X,i}^{(t)} \odot \mathbf{q}_{H,i}^{(t)},\;
\mathbf{k}_{X,j}^{(t)} \odot \mathbf{k}_{H,j}^{(t)} \rangle\Bigr)}
{\sum_{m=1}^{n_{\text{patch}}}
\exp\Bigl(\langle \mathbf{q}_{X,i}^{(t)} \odot \mathbf{q}_{H,i}^{(t)},\;
\mathbf{k}_{X,m}^{(t)} \odot \mathbf{k}_{H,m}^{(t)} \rangle\Bigr)},
\end{equation}
and
\begin{equation}
\label{eq:mult_z}
\mathbf{z}_i^{(t)} \;=\;
\mathbf{x}_i^{(t)}
\;+\; \sum_{j=1}^{n_{\text{patch}}}
\alpha_{i,j}^{(t)} 
\Bigl(\mathbf{v}_{X,j}^{(t)} \;\odot\; \mathbf{v}_{H,j}^{(t)}\Bigr).
\end{equation}
Here, the top-down feedback pathway multiplicatively gates the bottom-up signals. As a result, larger (smaller) magnitudes in the memory pathway can amplify (suppress) the corresponding magnitudes in the immediate visual pathway. This scheme allows for more direct \emph{control} (through multiplication) of attention weights and context vectors, enabling stronger or weaker gating of specific patches.

Additive operations can be dominated by whichever pathway has a larger magnitude, potentially diminishing subtler signals. By contrast, multiplicative modulation can act as a direct ``sign-flip'' mechanism or a global rescaling factor, making it inherently well-suited for precise top-down control. For instance, consider a scenario in which \(\mathbf{q}_{X,i}\) or \(\mathbf{k}_{X,j}\) contain elements \(\pm 2\). A feedback mechanism that must flip selected signs via \emph{addition} could require large compensatory values in \(\mathbf{q}_{H,i}\) or \(\mathbf{k}_{H,j}\). In contrast, a multiplicative pathway can achieve such sign flips with a scalar factor of \(-1\), regardless of the original magnitude in \(\mathbf{q}_{X,i}\) or \(\mathbf{k}_{X,j}\).

\section{Model Architecture}

Our model integrates a Vision Transformer (ViT) with a patch-based LSTM. First, a VAE is used to preprocess the raw visual features in a purely feed-forward method. Secondly, we utilize a recurrent ViT in which self-attention has been modified to incorporate immediate and recurrent inputs in order to construct the visual percept transmitted to the patch-based LSTM. Thirdly, the LSTM utilizes the projection from the recurrent ViT to update the patch-based internal states. 

\subsection{VAE Pre-Processing}

A Variational Autoencoder (VAE) is a generative model that learns to encode input data into a latent space and reconstructs the data from this latent representation. It combines principles from deep learning and probabilistic inference, making it suitable for modeling complex data distributions. It consists of two primary components, and encoder ($F$) that encodes visual inputs to a probabilistic latent space ($z_{latent}$), and a decoder ($G$) that decodes a sampled latent vector into a visual input.  

The encoder network $F$ entails multiple operations, $f\in F$ which serve to map an input image patch $\mathbf{o}_i \in \mathbb{R}^{H_{patch} \times W_{patch} \times C}$ to a latent representation characterized by a mean vector $\boldsymbol{z_{\mu}} \in \mathbb{R}^{d_{latent}}$ and a log-variance vector $\boldsymbol{z_{logvar}} \in \mathbb{R}^{d_{latent}}$, where $d_{latent}$ is the dimensionality of the latent space. The encoder consists of convolutional and fully connected layers as follows:

\begin{enumerate}
    \item \textbf{First Convolutional Layer}: Applies a convolution with 16 filters, each of size $3 \times 3$, stride 2, and padding 1. This operation reduces the spatial dimensions while increasing the feature depth. The activation function is ReLU:
    \[
    \mathbf{z_{Conv,1}} = \mathrm{ReLU}\left(f_{Conv,1}^{(1,16,3,2,1)}(\mathbf{o_i}) \right)
    \]
    \item \textbf{Second Convolutional Layer}: Applies a convolution with 32 filters, each of size $3 \times 3$, stride 2, and padding 1:
    \[
    \mathbf{z_{Conv,2}} = \mathrm{ReLU}\left( f_{Conv,2}^{(16,32,3,2,1)}(\mathbf{z_{Conv,1}}) \right)
    \]
    \item \textbf{Flattening}: The output tensor is reshaped into a vector:
    \[
    \mathbf{z_{flat,1}} = \mathrm{Flatten}(\mathbf{z_{Conv,2}})
    \]
    \item \textbf{First Fully Connected Layer}: Maps the flattened vector to a 128-dimensional feature vector:
    \[
    \mathbf{z_{flat,2}} = \mathrm{ReLU}\left( \mathbf{W}_1 \mathbf{z}_\text{flat} + \mathbf{b}_1 \right)
    \]
    \item \textbf{Latent Variable Parameters}: Computes the mean and log-variance vectors using two separate linear transformations:
    \begin{align*}
    \boldsymbol{z_\mu} &= \mathbf{W}_\mu \mathbf{z_{flat,2}} + \mathbf{b}_\mu, \\
    \boldsymbol{z_{\log\sigma^2}} &= \mathbf{W}_{\log\sigma^2} \mathbf{z_{flat,2}} + \mathbf{b}_{\log\sigma^2}
    \end{align*}
\end{enumerate}

To allow gradient-based optimization through stochastic sampling, we employ the reparameterization trick. Letting $\mu = \mathbf{z_\mu}$ and $\sigma = \exp(0.5 \mathbf{z_{logvar}})$ we draw a latent vector $\mathbf{z_{latent}}$ from the approximate posterior:
\[
\mathbf{z_{latent}} = \boldsymbol{\mu} + \boldsymbol{\sigma} \odot \boldsymbol{\epsilon}, \quad \boldsymbol{\epsilon} \sim \mathcal{N}(\mathbf{0}, \mathbf{I})
\]
where $\boldsymbol{\sigma} = \exp\left( \frac{1}{2} \boldsymbol{\log\sigma^2} \right)$, and $\odot$ denotes element-wise multiplication.

The decoder network $G$ maps the latent vector $\mathbf{z}$ back to the reconstructed image $\hat{\mathbf{o}_i}$. The decoder mirrors the encoder but uses transposed convolutions:

\begin{enumerate}
    \item \textbf{First Fully Connected Layer}: Transforms the latent vector to a 128-dimensional vector:
    \[
    \mathbf{\hat{o}_{flat,1}} = \mathrm{ReLU}\left( \mathbf{W}_{flat,1} \mathbf{z_{latent}} + \mathbf{b}_{flat,1} \right)
    \]
    \item \textbf{Second Fully Connected Layer}: Maps the 128-dimensional vector to a shape suitable for convolutional layers:
    \[
    \mathbf{\hat{o}_{flat,2}} = \mathrm{ReLU}\left( \mathbf{W}_{flat,2} \mathbf{\hat{o}_{flat,1}} + \mathbf{b}_{flat,2} \right)
    \]

    \item \textbf{First Transposed Convolutional Layer}: Applies a transposed convolution with 16 filters:
    \[
    \mathbf{\hat{o}_{ConvT,1}} = \mathrm{ReLU}\left( g_{ConvT,1}^{(32,16,3,2,1,0)}(\mathbf{\hat{o}_{flat,2}}) \right)
    \]
    \item \textbf{Second Transposed Convolutional Layer}: Applies a transposed convolution to reconstruct the image:
    \[
    \mathbf{\hat{o}_{ConvT,2}} = \textrm{Sigmoid}\left( g_{ConvT,2}^{(16,1,3,2,1,0)}(\mathbf{\hat{o}_{ConvT,1}}) \right)
    \]
\end{enumerate}

The VAE optimizes a loss function that combines reconstruction accuracy and the Kullback-Leibler (KL) divergence between the approximate posterior and the prior distribution. Letting $\mathbf{\hat{o}_i} = \mathbf{\hat{o}_{ConvT,2}}$ the loss is described as:
\[
\mathcal{L} = \frac{1}{d_{image}} \| \mathbf{o_i} - \hat{\mathbf{o_i}} \|^2 - \beta \cdot \frac{1}{2} \left( 1 + \log \boldsymbol{\sigma_i}^2 - \boldsymbol{\mu_i}^2 - \boldsymbol{\sigma_i}^2 \right)
\]
where $\beta$ is a hyperparameter that balances the two terms, i represents the image patch number, and $d_{image} = H_{patch} \times W_{patch} \times C$

\subsection{ViT}
Input images $\mathbf{O}^{(t)} \in \mathbb{R}^{50 \times 50}$ are sub-divided into four equal patches $\{\mathbf{o_1}^{(t)}, \mathbf{o_2}^{(t)}, \mathbf{o_3}^{(t)}, \mathbf{o_4}^{(t)}\}$, with $\mathbf{o_i}^{(t)}\in\mathbb{R}^{(25\times25)}$. We found that our RL agent learned fastest, was most interpretable, and demonstrated best performance when we used the second flattend encoder layer ($\mathbf{o_{flat,2}}$) as input to the ViT (as oppose to the latent encoding). Hence, for a given patch input $\mathbf{o_i}^{(t)}$ at time $t$, we have the encoding
\begin{equation*}
    \mathbf{\hat{o}_i}^{(t)} = f^*(\mathbf{o_i}^{(t)})
\end{equation*}
where $f^*(\cdot)$ includes encoder components (1)--(4). We also concatenate a (one-hot) positional ($\boldsymbol{\rho_i}$) and temporal  ($\boldsymbol{\tau}$) encoding. Thus the full pre-processed patch input at timestep $t$ is
\begin{equation}
    \mathbf{x_i}^{(t)} = \text{Concat}[\mathbf{\hat{o}_i}^{(t)}, \boldsymbol{\rho_i}, \boldsymbol{\tau}]
\end{equation}

The complete input to the ViT at time step $t$ is:
\begin{equation}
    \mathbf{X}^{(t)}= (\mathbf{x_1}^{(t)}, \mathbf{x_2}^{(t)}, \mathbf{x_3}^{(t)}, \mathbf{x_4}^{(t)})^T \in \mathbb{R}^{4 \times 140}
\end{equation}
The transformer computes queries, keys, and values as:
\begin{align}
    \mathbf{Q} &= (\mathbf{X^{(t)} W_{XQ}}) \odot (\mathbf{H^{(t-1)} W_{HQ}}) \\
    \mathbf{K} &= (\mathbf{X^{(t)} W_{XK}}) \odot (\mathbf{H^{(t-1)} W_{HK}}) \\
    \mathbf{V} &= (\mathbf{X^{(t)} W_{XV}}) \odot (\mathbf{H^{(t-1)} W_{HV}})
\end{align}
where $\mathbf{W_{X\cdot}} \in \mathbb{R}^{140 \times 140}$, $\mathbf{W_{H\cdot}} \in \mathbb{R}^{1024 \times 140}$, $\mathbf{H}^{(t-1)}$ is the activated memory from the previous timestep, $\odot$ denotes Hadamard product, and we have dropped the temporal superscript (implicit). Self-attention is computed as:
\begin{equation}
    \mathbf{V_{{filtered}}} = \text{Softmax}(\mathbf{QK}^T)\mathbf{V}
\end{equation}
The spatially and temporally aware visual percept is constructed as follows:
\begin{equation}
    \mathbf{Z}^{(t)} = \mathbf{X}^{(t)} + \mathbf{V_{filtered}} \in \mathbb{R}^{4 \times 140}
\end{equation}

\subsection{Spatial LSTM}

We adapt the xLSTM architecture \cite{beck2024xlstm} for spatial memory. The LSTM operations are shown in \autoref{eq:recurrent_updates}. The projection matrices have dimensions $\mathbf{W_{x}} \in \mathbb{R}^{140 \times 1024}$, and $\mathbf{R_{x}} \in \mathbb{R}^{140 \times 1024}$. This ensures all output variables have shape $4 \times 1024$. As described above, we call this a patch-based LSTM because there is a hidden state for each patch of the visual scene. Importantly, within the LSTM the hidden states are updated independently. The matrices $\mathbf{Z}^{(t)}$, $\mathbf{C}^{(t)}$, $\mathbf{H}^{(t)}$, $\mathbf{M}^{(t)}$, and $\mathbf{N}^{(t)}$ are of shape $n_{patch}$ by $d$, where $d\in{d_{latent}, d_{mem}}$. Right multiplication by the matrices $\mathbf{W_x}$ or $\mathbf{R_x}$ projects the latent embedding or hidden state of a specific patch to another space, independent of the other patches. By constructions, self-attention is the only mechanism by which information from visual patches (or mnemonic patches) is communicated to other patches. 
\begin{table*}[b] 
\captionsetup{position=bottom} 
\centering
\small
\begin{align*}
    \mathbf{\tilde{I}}^{(t)} &= \mathbf{Z}^{(t)} \mathbf{W_i} + \mathbf{H}^{(t-1)} \mathbf{R_i} & 
    \mathbf{I}^{(t)} &= \exp(\mathbf{\tilde{I}}^{(t)} - \mathbf{M}^{(t)}) & 
    \mathbf{O}^{(t)} &= \sigma(\mathbf{\tilde{O}}^{(t)}) \\
    \mathbf{\tilde{F}}^{(t)} &= \mathbf{Z}^{(t)} \mathbf{W_f} + \mathbf{H}^{(t-1)} \mathbf{R_f} & 
    \mathbf{F}^{(t)} &= \exp(\mathbf{\tilde{F}}^{(t)} + \mathbf{M}^{(t-1)} - \mathbf{M}^{(t)}) & 
    \mathbf{N}^{(t)} &= \mathbf{F}^{(t)} \odot \mathbf{N}^{(t-1)} + \mathbf{I}^{(t)} \\
    \mathbf{\tilde{O}}^{(t)} &= \mathbf{Z}^{(t)} \mathbf{W_o} + \mathbf{H}^{(t-1)} \mathbf{R_o} & 
    \mathbf{M}^{(t)} &= \max(\mathbf{\tilde{F}}^{(t)} + \mathbf{M}^{(t-1)}, \mathbf{\tilde{I}}^{(t)}) & 
    \mathbf{U}^{(t)} &= \tanh(\mathbf{\tilde{U}}^{(t)}) \\
    \mathbf{\tilde{U}}^{(t)} &= \mathbf{Z}^{(t)} \mathbf{W_u} + \mathbf{H}^{(t-1)} \mathbf{R_z} & 
    \mathbf{C}^{(t)} &= \mathbf{C}^{(t-1)} \odot \mathbf{F}^{(t)} + \mathbf{U}^{(t)} \odot \mathbf{I}^{(t)} & 
    \mathbf{H}^{(t)} &= \mathbf{O}^{(t)} \odot (\mathbf{C}^{(t)} / \mathbf{N}^{(t)})
\end{align*}
\caption{Equations defining the recurrent network update process.}
\label{eq:recurrent_updates}
\end{table*}

\subsection{Actor-Critic Reinforcement Learning}

Our model is trained using an actor–critic reinforcement learning (RL) framework~\cite{sutton2018reinforcement} in which the agent learns to select actions that maximize long‐term rewards. At each timestep, the agent observes a mnemonic percept 
\[
\mathbf{H}^{(t)} \in \mathbb{R}^{4\times1024},
\]
which encodes spatial and temporal context, and selects an action from a binary set:
\[
a_t = 
\begin{cases}
0, & \text{(``wait'' action)}\\[1mm]
1, & \text{(``declare change'' action)}
\end{cases}.
\]
The value function
\begin{equation}
    V(H^{(t)}) = \mathbb{E}\!\left[\sum_{\tau=t}^{T} \gamma^{\tau-t} \, r_\tau \,\bigg|\, H^{(t)}\right]
    \label{eq:V_new}
\end{equation}
estimates the expected return from the current memory state, where \(\gamma\in[0,1]\) is the discount factor, \(r_\tau\) is the reward at time \(\tau\), and \(T\) is the final timestep.

Learning is driven by the temporal difference (TD) error,
\begin{equation}
    \delta_t = r_t + \gamma\,V(H^{(t+1)}) - V(H^{(t)}),
    \label{eq:td_error_new}
\end{equation}
which is used to update both the critic (value) and the actor (policy) networks. The policy is adjusted to favor actions with higher estimated returns, enabling the agent to improve its decision-making based on experience.

\subsection{Distributional Framing, Network Architecture, and Loss Functions}

In our approach the critic network estimates a \emph{distributional} Q-function rather than a single scalar value. For a state–action pair \((H_t,a_t)\), the critic outputs a probability distribution over 15 discrete Q-value bins. The critic network is defined as follows:
\begin{align*}
    a' &= a_t W_a + b_a, \\
    q_0 &= \text{Concat}[H',\, a'], \\
    q_1 &= \text{ELU}(q_0 W_1 + b_1), \\
    q_2 &= \text{ELU}(q_1 W_2 + b_2), \\
    q_3 &= \text{ELU}(q_2 W_3 + b_3), \\
    p_\theta(q\mid H_t,a_t) &= \text{Softmax}\Bigl(q_3 W_{\text{out}} + b_{\text{out}}\Bigr),
\end{align*}
where \(H'\in\mathbb{R}^{4096}\) is the flattened mnemonic percept.

The improved (target) policy is defined as
\begin{equation*}
    \pi_{\text{imp}}(a_t\mid s_t) \propto \exp\!\Bigl(\frac{Q_{\theta'}(s_t,a_t)}{\eta}\Bigr)\,\pi_{\theta'}(a_t\mid s_t),
    \label{eq:pi_imp}
\end{equation*}
where \(\theta'\) denotes the parameters of a target network and \(\eta>0\) is a temperature parameter.

The target Q-distribution is computed via the distributional Bellman operator:
\begin{equation*}
\begin{split}
    \Gamma_{\theta'}(q\mid s_t,a_t) =\; &\mathbb{E}_{s_{t+1}}\Biggl[
    \mathbb{E}_{a'\sim\pi_\theta(\cdot\mid s_{t+1})}\Bigl[
    \mathbb{E}_{q'\sim p_\theta(\cdot\mid s_{t+1},a')}\, \\
    &\quad \mathbf{1}_{\left[q-\frac{\epsilon}{2},\,q+\frac{\epsilon}{2}\right]}\Bigl(r_t+\gamma\,q'\Bigr)
    \Bigr]\Biggr].
\end{split}
\label{eq:bellman_new}
\end{equation*}

The actor network maps the mnemonic percept \(H_t\) to an action distribution through a 4-layer feed-forward network:
\begin{align*}
    \mu_1 &= \text{ELU}(H' W_1 + b_1), \\
    \mu_2 &= \text{ELU}(\mu_1 W_2 + b_2), \\
    \mu_3 &= \text{ELU}(\mu_2 W_3 + b_3), \\
    \pi_\theta(a_t\mid H_t) &= \text{Softmax}\Bigl(\mu_3 W_{\text{out}} + b_{\text{out}}\Bigr).
\end{align*}

Training employs a KL-regularized objective that jointly updates the actor and the critic. The actor loss is defined as the KL divergence between the improved policy and the current policy:
\begin{equation*}
\mathcal{L}_{\text{actor}}(\theta) = D_{\text{KL}}\Bigl(\pi_{\text{imp}}(\cdot\mid s_t) \,\|\, \pi_\theta(\cdot\mid s_t)\Bigr),
\label{eq:actor_loss}
\end{equation*}
which, up to an additive constant, is equivalent to
\begin{equation*}
\mathcal{L}_{\text{actor}}(\theta) = -\mathbb{E}_{a\sim \pi_{\text{imp}}}\Bigl[\log \pi_\theta(a\mid s_t)\Bigr].
\label{eq:actor_loss_expanded}
\end{equation*}
Similarly, the critic loss is defined as the KL divergence between the target Q-distribution and the predicted Q-distribution:
\begin{equation*}
\mathcal{L}_{\text{critic}}(\theta) = \beta\,D_{\text{KL}}\Bigl(\Gamma_{\theta'}(q\mid s_t,a_t) \,\|\, p_\theta(q\mid s_t,a_t)\Bigr),
\label{eq:critic_loss}
\end{equation*}
where \(\beta>0\) is a balancing hyperparameter. The overall loss is given by
\begin{equation*}
\mathcal{L}(\theta) = \mathcal{L}_{\text{actor}}(\theta) + \mathcal{L}_{\text{critic}}(\theta).
\label{eq:total_loss}
\end{equation*}

In summary, our model learns to select actions that maximize long-term rewards by jointly training the actor and the distributional critic with a KL-regularized objective \cite{springenberg2024offline}. The network architecture—designed to process spatially structured memory representations—utilizes self-attention and feed-forward layers, with long equations split over multiple lines to ensure clarity in our two-column format.

\subsection{Task Difficulty}

To control task difficulty, Gabor stimuli were corrupted with rotational "noise". Defnining $\theta^*_i$ as the "true" Gabor orientation for $S_i$, the  orientation in the input image shown to the agent is:
\[
\theta_i = \theta^*_i + \delta_{it}
\]
where $\delta_{it} \sim \mathcal{N}(0,\sigma)$ is the rotational noise at time step $t$. If the stimulus is selected for change, then at $t=5$ and $t=6$:
\[
\theta_i = \theta^*_i + \Delta + \delta_{it}
\]
The orientation noise parameter $\sigma$ is set to 5. The orientation change parameter $\Delta$ is a random variable drawn at the beginning of a change trial, with $\Delta \sim \textrm{U}(-k, k)$, where $k$ is adjusted based on the agent's performance, starting at $k=65$ and decreasing as performance improves to increase task difficulty.

\bibliographystyle{unsrt}
\bibliography{ref2}  

\end{document}


\tableofcontents
\title{Supplement  }
\date{\today}
\maketitle

\section{Methods}


\subsection{Model Overview}
\begin{figure}[htbp]
    \centering
    \vspace*{-0mm} 
    \includegraphics[width=0.99\linewidth]{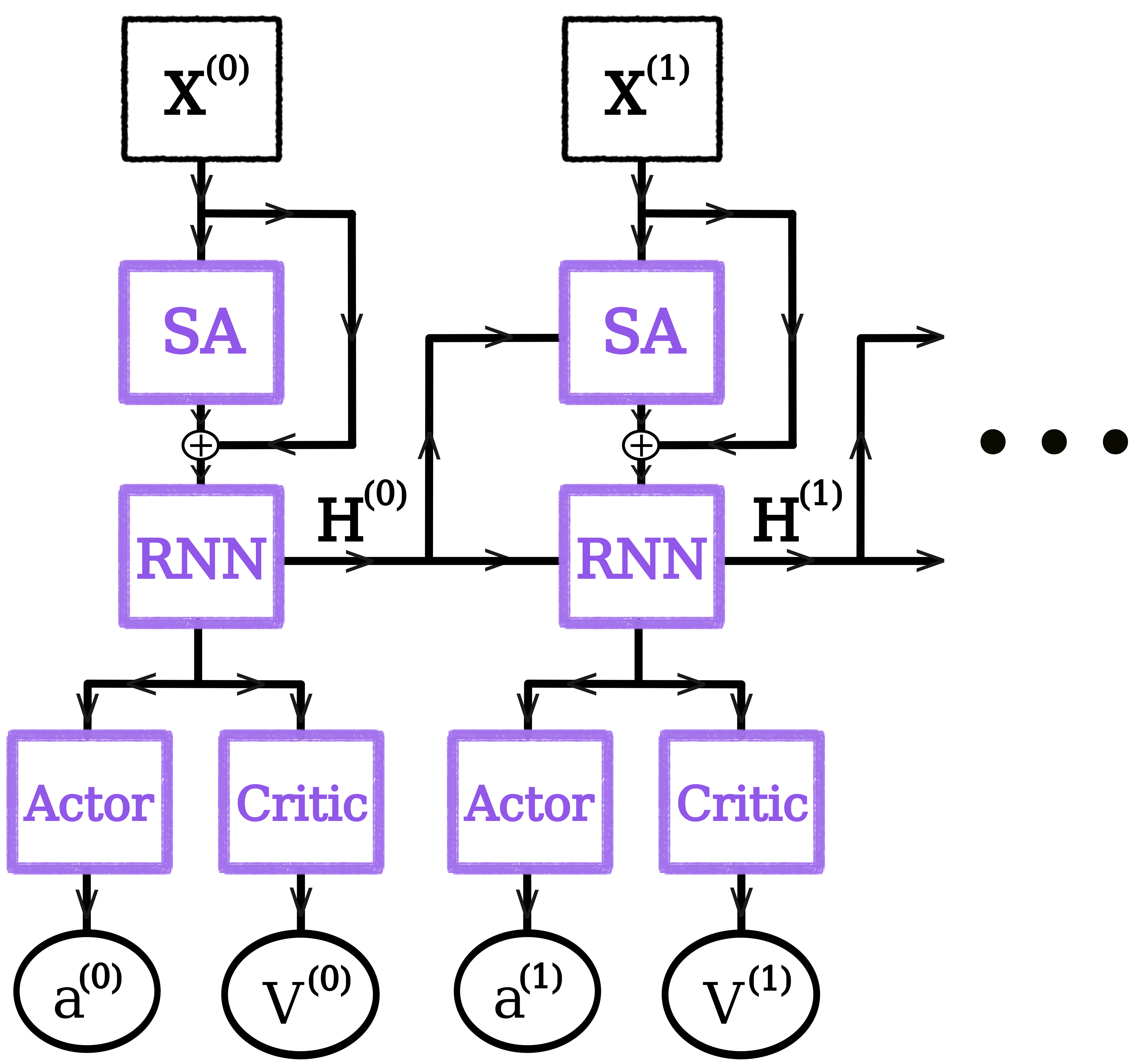}
    \caption{Model Concept. Arrows indicate directional flow of information and processing. Images ($X^{(t)}$) are generated from the environment. These images are then fed through the Recurrent ViT from which the resulting output is passed to an RNN. The internal state of the RNN is updated used in the decision making process, and also passed to the Recurrent ViT and RNN at the next time step. Purple boxes denote artificial neural networks}
    \label{fig:ModHighLevel}
\end{figure}

The objective of the model is to utilize immediate visual inputs in order to update an internal state with sufficient immediate and past visual information such that downstream decoders can estimate value and take action. We utilize a self-attention (SA) mechanism to construct the visual percept used to update the internal state of the RNN. Self-attention is computed based on the immediate visual inputs and feedback from the RNN. The following sections will describe the motivations and details of this process in more depth.

\subsection{Pre-Processing, Content Selection, and Construction for Visual Working Memory}
Given a visual scene (an image) of dimension $H \times W \times C$, denoted by $\mathbf{\mathcal{O}}^{(t)} \in \mathbb{R}^{H \times W \times C}$ at time $t$, the agent views the entire scene through a fixation at center field. During preprocessing, the image is partitioned into a set of patches, $\{\mathbf{{o}_i}^{(t)}\}_{i=1}^{n_{patch}}$, where each patch is of size $H_{patch} \times W_{patch} \times C_{patch}$. The original visual patches are then transformed into a compact set of internal representations, $\{\mathbf{x_i}^{(t)}\}_{i=1}^{n_{patch}}$, each $\mathbf{x_i}^{(t)} \in \mathbb{R}^{H_{patch} \times W_{patch} \times C_{patch}}$ and typically satisfying
\[
    \dim(\mathbf{x_i}^{(t)}) \ll \dim\bigl(\mathbf{o_i}^{(t)}\bigr).
\]
Together, the collection of these feature patches forms the immediate visual information available to the agent at time $t$, denoted by $\mathbf{X}^{(t)} = \{\mathbf{x_i}^{(t)}\}_{i=1}^{n_{patch}}$.

\subsection{Spatially Oriented Visual Working Memory}

\begin{figure}[H]
    \centering
    \vspace*{-1mm} 
    \includegraphics[width=0.89\linewidth]{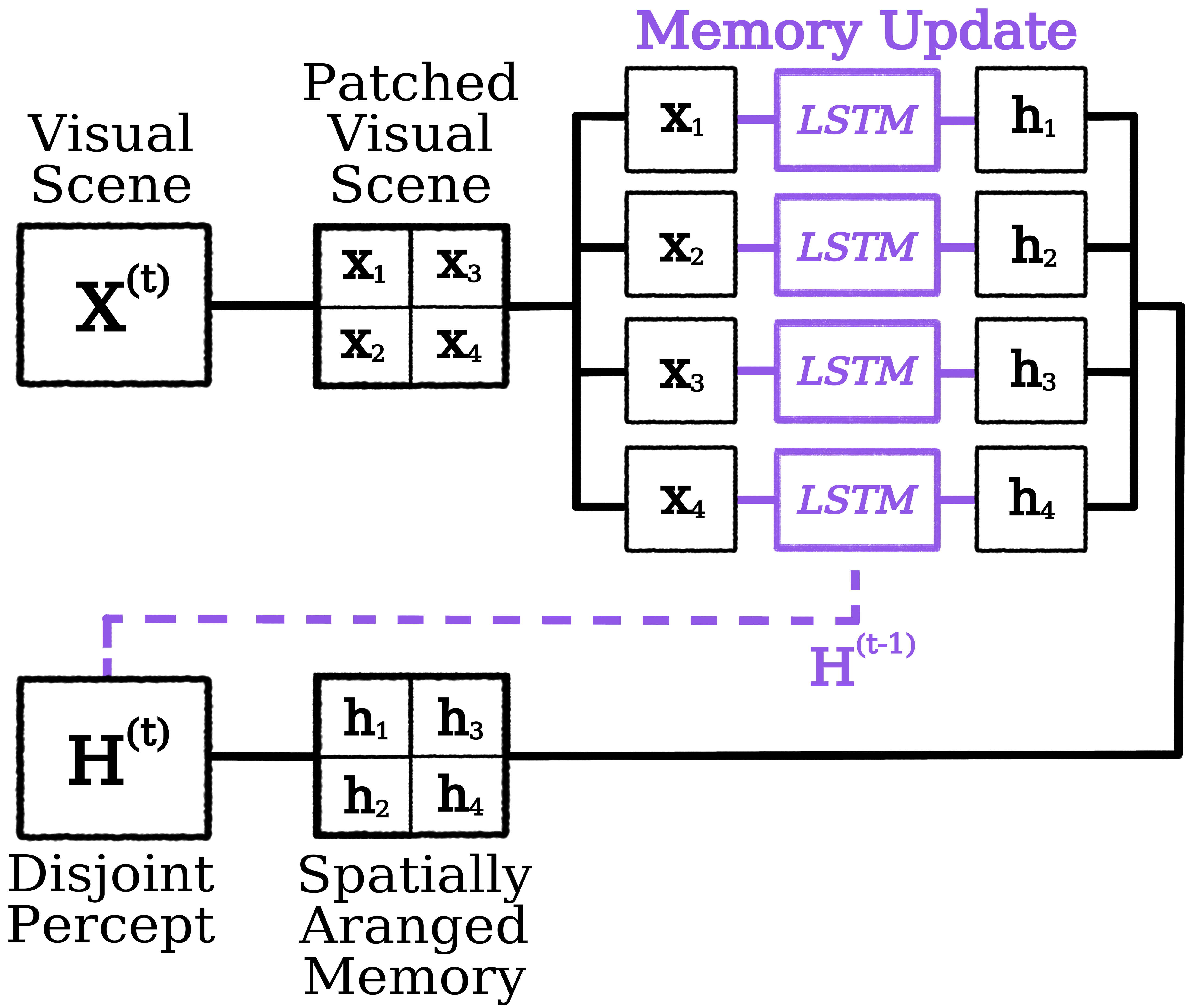}
    \captionsetup{justification=raggedright,singlelinecheck=false}
    \caption{The patch-based LSTM. In essence, it is a standard LSTM architecture \cite{hochreiter1997long, beck2024xlstm}. The only difference is that we apply the LSTM in parallel to all patches in a visual scene. This yields separate recurrent states for each patch. However, the weights local to the LSTM are shared among all patches. I.e., the weights utilized to construct $h_4^{(t)}$ given $x_4^{(t)}$ and $h_4^{(t-1)}$ are the same weights sued to construct $h_3^{(t)}$ given $x_3^{(t)}$ and $h_3^{(t-1)}$ and the two operations occur in parallel. The patch-based LSTM does not mix information among input patches. }
    \label{fig:patch_based_LSTM}
\end{figure}

Following numerous experimental findings, we allow our model to maintain a spatially arranged visual working memory \cite{wheeler2002binding, pertzov2014privileged,schneegans2017neural, van2019human}, in which each patch location $i$ has a corresponding patched memory component $\mathbf{c_i}^{(t)} \in \mathbb{R}^{d_{mem}}$ within the RNN. For updating the internal state of our RNN, we utilize the operations and functions described in the LSTM architecture \cite{hochreiter1997long, beck2024xlstm}. For the remainder of our description, we will refer to the collection $C^{(t)}=\{c_i^{(t)}\}_{i=1}^{n_{patch}}$ the VWM state and $c_i^{(t)}$ a VWM patch, where
\begin{equation}
    \label{eq:C}
    \mathbf{c_i}^{(t)} =  \mathbf{f_i}^{(t)} \odot \mathbf{c_i}^{(t-1)} + \mathbf{u_i}^{(t)} \odot \boldsymbol{\psi_i}^{(t)},
\end{equation}
where 
\[
\mathbf{f_i}^{(t)} = F\!\bigl(\mathbf{x_i}^{(t)},\, \mathbf{h_i}^{(t-1)}\bigr), \quad
\mathbf{u_i}^{(t)} = U\!\bigl(\mathbf{x_i}^{(t)},\, \mathbf{h_i}^{(t-1)}\bigr), \quad
\boldsymbol{\psi_i}^{(t)} = \Psi\!\bigl(\mathbf{x_i}^{(t)},\, \mathbf{h_i}^{(t-1)}\bigr).
\]
The operator $\odot$ denotes elementwise multiplication. Here, $\mathbf{f_i}^{(t)} \in [0,1]^{\,d_{mem}}$ determines which parts of $\mathbf{c_i}^{(t-1)}$ are \textit{forgotten} (i.e., decayed), while $\mathbf{u_i}^{(t)} \in [-1,1]^{\,d_{mem}}$ and $\boldsymbol{\psi_i}^{(t)} \in \mathbb{R}^{d_{mem}}$ selectively modulate and propose new content. Altogether, these operations enable dynamic insertion, maintenance, and forgetting of information in $\mathbf{c_i}^{(t)}$. 

The activated memory patch, $h_i^{(t)}$ is constructed from the VWM patch, $c_i^{(t)}$:
\[
    \mathbf{h_i}^{(t)} = \boldsymbol{\phi_i}^{(t)} \odot \Bigl(\tfrac{\mathbf{c_i}^{(t)}}{\mathbf{n_i}^{(t)}}\Bigr),
\]
where $\boldsymbol{\phi_i}^{(t)} = \Phi\!\bigl(\mathbf{x_i}^{(t)}, \mathbf{h_i}^{(t-1)}\bigr) \in [0,1]^{\,d_{mem}}$ selects elements of $\mathbf{c_i}^{(t)}$ for downstream processing, and $\mathbf{n_i}^{(t)}$ is a normalization term. Each function $F, U, \Psi, \Phi$ is parameterized by a feedforward neural network that receives $\mathbf{x_i}^{(t)}$ and $\mathbf{h_i}^{(t-1)}$ as inputs. Although the parameters of these functions remain fixed once trained, the recurrent operation through $\mathbf{h_i}^{(t)}$ allows the system to track temporal dynamics.

\subsection{A Disjoint Memory}
Following the ideas presented by Knudsen \cite{knudsen2007fundamental}, the activated subset of working memeory is central for decision-making and planning. However, a patched RNN as described above presents a clear shortcoming: each $\mathbf{h_i}^{(t)}$ encodes the content of its own patch independently, without explicit awareness of neighboring patches. If downstream processes (\emph{e.g.}, a decoder $\pi$) require spatial or contextual relationships among patches, they must construct these relationships entirely from the population of VWM patches. Moreover, the problem intensifies over time. If the network must integrate information from patches across multiple timesteps (e.g., $\mathbf{x_i}^{(\tau)}$ and $\mathbf{x_j}^{(\tau)}$ for $\tau < t$), then each VWM patch must \emph{retain} all potentially significant current and past features useful for task-relevant decoding by downstream networks. This approach quickly becomes intractable, as it demands that the architecture store a large number of unique \emph{conjunctions} of spatio-temporal features. This challenge aligns with the combinatorial explosion recognized by Tsotsos \cite{tsotsos1988complexity} as a core difficulty in perceptual organization. 

To circumvent these limitations, we introduce self-attention into the encoding process, encouraging each patch’s representation to reflect the context provided by the other patches within the same timestep. In doing so, we create a spatially integrated or \emph{context-aware} activated memory before the information even updates the VWM.

\subsection{Self-Attention and Spatially Aware activated memory}

\begin{figure}[H]
    \centering
    \vspace*{-0mm} 
    \includegraphics[width=0.99\linewidth]{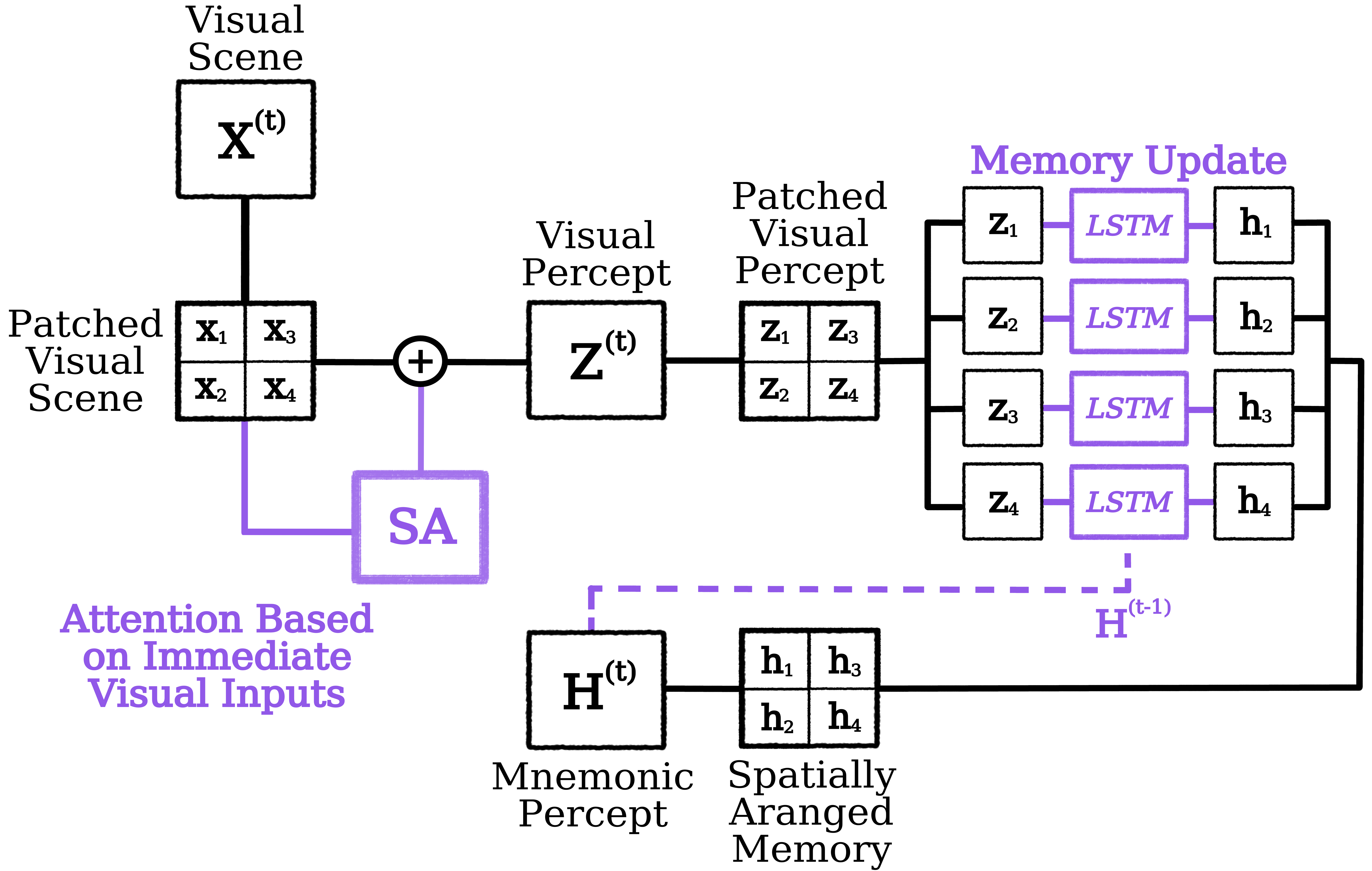}
    \captionsetup{justification=raggedright,singlelinecheck=false}
    \caption{The self-attention mechanism merges immediate visual inputs with spatial context. This allows the patch-based LSTM to store relevant visual information form patches and their potential significance given the content in the visual scene.}
    \label{fig:attention}
\end{figure}

We wish to obtain a scene-level representation $\mathbf{Z}^{(t)} = \{\mathbf{z_i}^{(t)}\}_{i=1}^{n_{patch}}$ such that each $\mathbf{z_i}^{(t)}$ encodes the task-relevant spatial relationships among the visual feature patches $\mathbf{x_1}^{(t)}, \dots, \mathbf{x_{n_{patch}}}^{(t)}$. By doing so, the patch-based LSTM will be able to utilize immediate visual information within a patch and task-relevant spatial context to update internal states. Formally, we want
\[
\mathbf{z_i}^{(t)} = f_z\Bigl(\mathbf{x_i}^{(t)}, \{\mathbf{x_j}^{(t)}\}_{j\neq i}\Bigr),
\]
A straightforward way to implement this is via self-attention:
\begin{equation}
    \label{eq:Z}
    \mathbf{z_i}^{(t)} = \mathbf{x_i}^{(t)} + \sum_{j=1}^{n_{patch}} a_{ij}^{(t)}\, \mathbf{v_j}^{(t)},
\end{equation}
where $\mathbf{v_j}^{(t)} = V\!\bigl(\mathbf{x_j}^{(t)}\bigr)$ is a function that maps the feature patch into a latent space and $a_{ij}^{(t)} = A\bigl(\mathbf{x_i}^{(t)}, \mathbf{x_j}^{(t)}\bigr)$ indicates the \emph{relative importance} of $\mathbf{x_j}^{(t)}$ with respect to $\mathbf{x_i}^{(t)}$. To ensure a proper probability-like weighting, we impose $\sum_{j=1}^{n_{patch}} a_{ij}^{(t)} = 1$ with $a_{ij}^{(t)} \in (0,1)$. A typical choice for $a_{ij}^{(t)}$ is:
\[
    a_{ij}^{(t)} \;=\; \frac{\exp\Bigl(\bigl\langle \mathbf{q_i}^{(t)}, \mathbf{k_j}^{(t)}\bigr\rangle\Bigr)}
    {\sum_{m=1}^{n_{patch}} \exp\Bigl(\bigl\langle \mathbf{q_i}^{(t)}, \mathbf{k_m}^{(t)}\bigr\rangle\Bigr)},
\]
where $\mathbf{q_i}^{(t)} = Q\!\bigl(\mathbf{x_i}^{(t)}\bigr)$ and $\mathbf{k_j}^{(t)} = K\!\bigl(\mathbf{x_j}^{(t)}\bigr)$ are \emph{query} and \emph{key} functions, respectively. Interpreting $a_{ij}^{(t)}$ as a salient feature map has strong parallels to the saliency map hypothesis \cite{koch1984selecting}; however, we adopt a \emph{winner-takes-most} approach rather than a strict winner-takes-all (WTA), common in many self-attention applications. In principle, should $a_{i,j^*}^{(t)} \approx 1$ for some $j^*$ and $a_{i,m}^{(t)} \approx 0$ for $m \neq j^*$, we recover WTA-like mechanism.

After computing $\mathbf{Z}^{(t)}$ for the entire scene, the visual percept patches, $z_i^{(t)}$, are used to update the VWM patches:
\[
    \mathbf{c_i}^{(t)} \;=\;  \mathbf{f_i}^{(t)} \odot \mathbf{c_i}^{(t-1)} \;+\; \mathbf{u_i}^{(t)} \odot \boldsymbol{\psi_i}^{(t)},
\]
where each function is now evaluated using $\mathbf{z_i}^{(t)}$ rather than the immediate visual scene patch in isolation, $\mathbf{x_i}^{(t)}$. This approach solves the \emph{spatial} integration problem in the current timestep. Yet, any feature with contextual importance \emph{across} timesteps remains challenging: we still need a mechanism to capture top-down feedback or \emph{memory-based} salience.

\subsection{Recurrent Feedback From Memory}

\begin{figure}[H]
    \centering
    \vspace*{-1mm} 
    \includegraphics[width=0.99\linewidth]{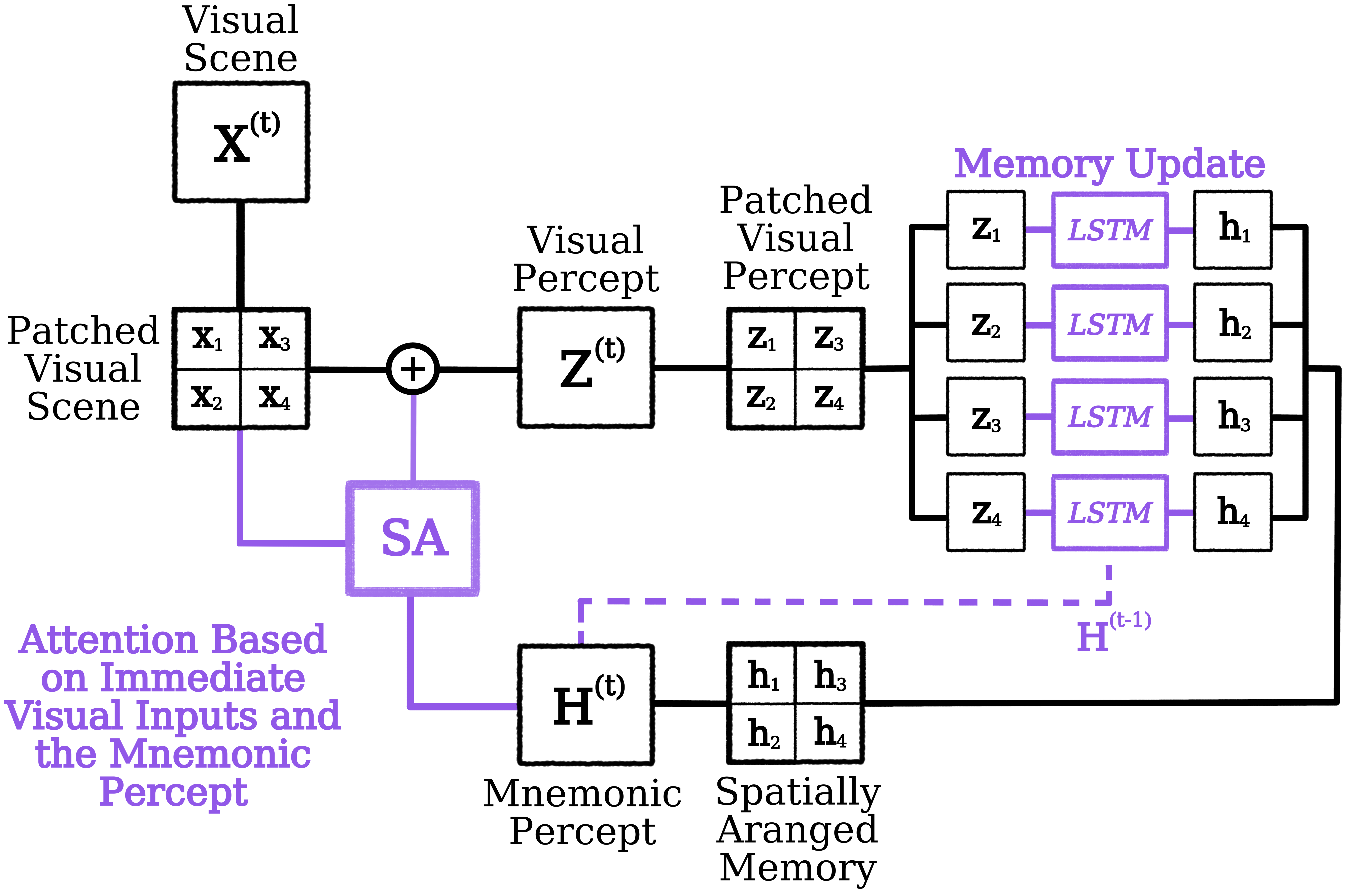}
    \captionsetup{justification=raggedright,singlelinecheck=false}
    \caption{A depiction of recurrent self-attention. In order to construct a meaningful representation of the current visual scene, spatial and temporal context are merged with immediate visual inputs.}
    \label{fig:self_attention_recurrent}
\end{figure}

Knudsen \cite{knudsen2007fundamental} describes a feedback loop in which working memory provides top-down signals that bias neural representations relevant to the organism’s current goals. In the context of the biased competition model \cite{desimone1995neural}, working memory holds an \emph{attentional template} that biases competition in favor of task-relevant representations. However, in practice it is not clear how this mnemonic feedback is/should be implemented. In this we simplify (and constrain) the problem to implementing recurrent feedback from the patch-based LSTM to the self-attention mechanism of the ViT. Hence, we evaluate three different methods in terms of their ability to yield primate-like behavior signatures of attention. We call the vision transformer with mnemonic feedback the recurrent ViT.   

\subsection{Mnemonic Guidance}

\subsubsection{Visual working memory as tokens}
\begin{figure}[H]
    \centering
    \vspace*{-1mm} 
    \includegraphics[width=0.99\linewidth]{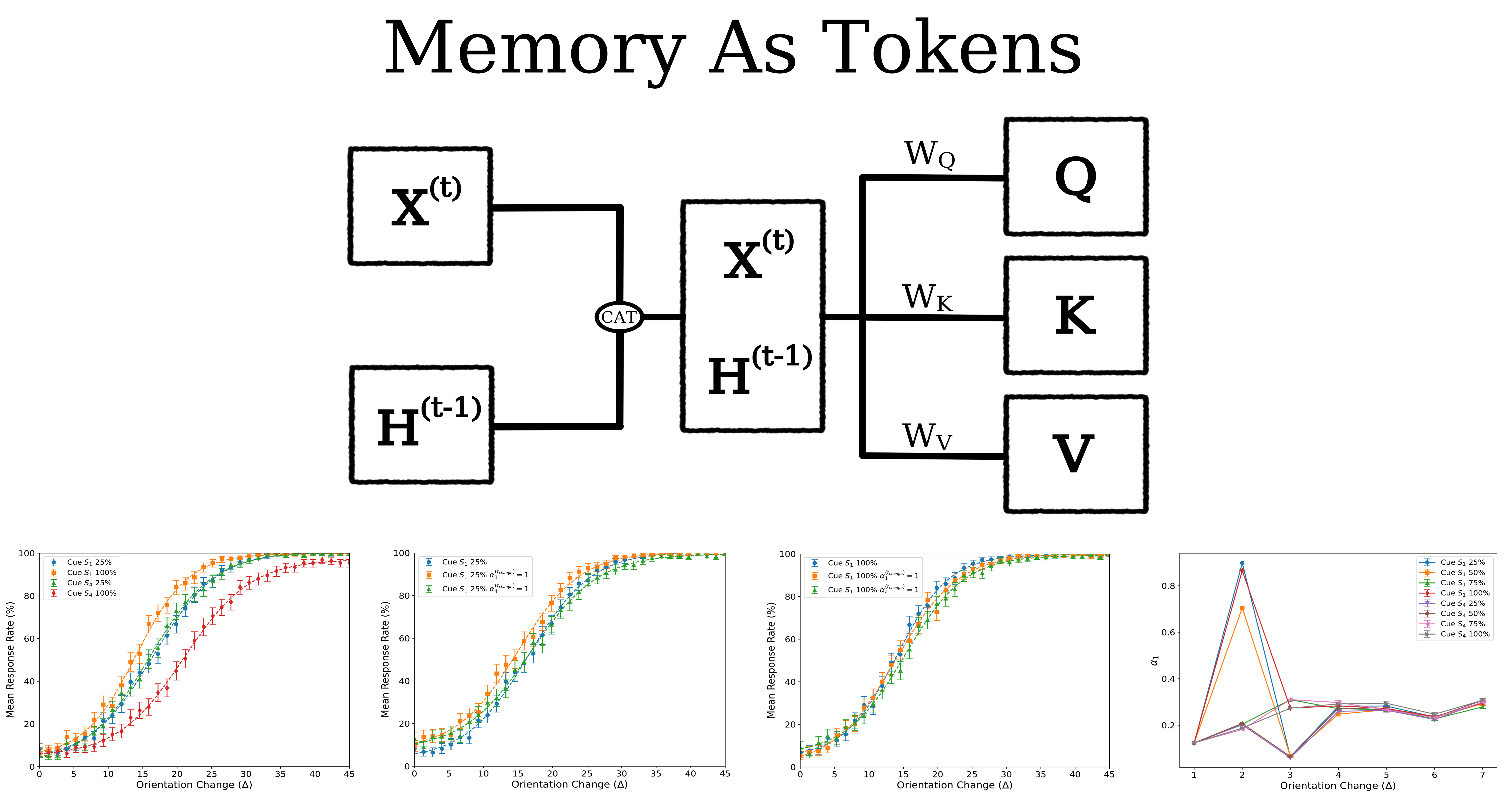}
    \captionsetup{justification=raggedright,singlelinecheck=false}
    \caption{Circuit diagram and some behavioral results for a model in which recurrent feedback is implemented via concatenating recurrent states to immediate visual inputs. The concatenation effectively doubles the number of patches in the visual percept $Z^{(t)}$, but we reduce the number back to $n_{patch}$ before the transmission of the visual percept to the LSTM. While we see cue effects on behavior, this model lacks any significant attention modulation effects. In addition, we also observe that the cue has little effect on downstream deployment of attention at later time points.}
    \label{fig:MemAsToken}
\end{figure}

The first recurrent feedback method we evaluate is one in which we concatenate the mnemonic percept to the visual input. Thus, the input to the self-attention mechanism is
\[
\mathbf{\tilde{X}} = Concatenate[\mathbf{X}^{(t)},\mathbf{H}^{(t-1)}]
\]
where $\mathbf{\tilde{X}}\in\mathbb{R}^{2n_{patch},d_{model}}$. From here we define:
\[
\mathbf{q}_{\tilde{X},i}^{(t)} = Q_{\tilde{X}}\bigl(\mathbf{\tilde{x}}_i^{(t)}\bigr), \quad
\mathbf{k}_{\tilde{X},j}^{(t)} = K_{\tilde{X}}\bigl(\mathbf{\tilde{x}}_j^{(t)}\bigr), \quad
\mathbf{v}_{\tilde{X},j}^{(t)} = V_{\tilde{X}}\bigl(\mathbf{\tilde{x}}_j^{(t)}\bigr),
\]

The attention weights are given by
\begin{equation}
\label{eq:add_alpha}
\alpha_{i,j}^{(t)} \;=\;
\frac{\exp\Bigl(\langle \mathbf{q}_{\tilde{X},i}^{(t)} ,\;
\mathbf{k}_{\tilde{X},j}^{(t)}  \rangle\Bigr)}
{\sum_{m=1}^{n_{\text{patch}}}
\exp\Bigl(\langle \mathbf{q}_{\tilde{X},i}^{(t)} ,\;
\mathbf{k}_{\tilde{X},m}^{(t)} \rangle\Bigr)}.
\end{equation}
We then compute the output representation as
\begin{equation}
\label{eq:add_z}
\mathbf{z}_i^{(t)} \;=\; \mathbf{x}_i^{(t)}
\;+\; \sum_{j=1}^{2n_{\text{patch}}}
\alpha_{i,j}^{(t)} \mathbf{v}_{\tilde{X},j}^{(t)} .
\end{equation}
Here, we only take the first $n_{patch}$ entries $\{\mathbf{z}^{(t)}_{i}\}_{i=1}^{n_{patch}}$. The reason for this is because there are only $n_{patch}$ recurrent states in the patch-based LSTM, and $Z^{(t)}=\{\mathbf{z}^{(t)}\}_{i=1}^{n_{patch}}$ is only used as the input to the LSTM. 

\subsubsection{Additive Feedback from Visual Working Memory}

\begin{figure}[H]
    \centering
    \vspace*{-1mm} 
    \includegraphics[width=0.99\linewidth]{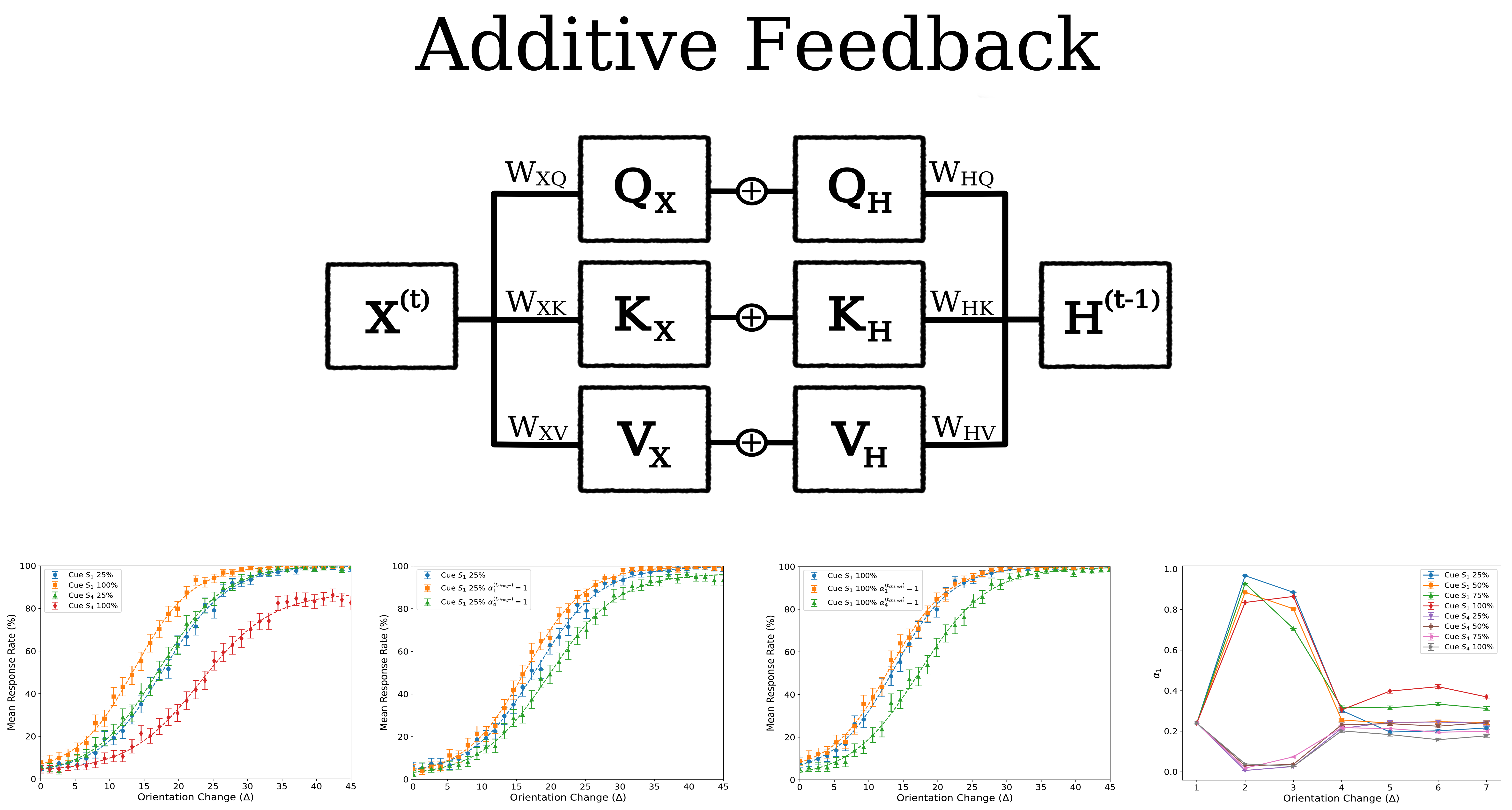}
    \captionsetup{justification=raggedright,singlelinecheck=false}
    \caption{Circuit diagram and some behavioral results for a model in which recurrent feedback is implemented via the addition of parallel projections onto the Q, K, V self-attention components. There are significant cue effects and slight attention modulation effects on behavior. The cue also affects the deployment of attention at the time of change, albeit only slightly.  }
    \label{fig:AdditiveFeedback}
\end{figure}

We split the standard self-attention operation into two parallel pathways: one for the bottom-up immediate visual inputs, \(\mathbf{x}_i^{(t)}\), and one for the top-down mnemonic inputs, \(\mathbf{h}_i^{(t)}\). Define:
\[
\mathbf{q}_{X,i}^{(t)} = Q_X\bigl(\mathbf{x}_i^{(t)}\bigr), \quad
\mathbf{k}_{X,j}^{(t)} = K_X\bigl(\mathbf{x}_j^{(t)}\bigr), \quad
\mathbf{v}_{X,j}^{(t)} = V_X\bigl(\mathbf{x}_j^{(t)}\bigr),
\]
and
\[
\mathbf{q}_{H,i}^{(t)} = Q_H\bigl(\mathbf{h}_i^{(t)}\bigr), \quad
\mathbf{k}_{H,j}^{(t)} = K_H\bigl(\mathbf{h}_j^{(t)}\bigr), \quad
\mathbf{v}_{H,j}^{(t)} = V_H\bigl(\mathbf{h}_j^{(t)}\bigr).
\]
The attention weights are given by
\begin{equation}
\label{eq:add_alpha}
\alpha_{i,j}^{(t)} \;=\;
\frac{\exp\Bigl(\langle \mathbf{q}_{X,i}^{(t)} + \mathbf{q}_{H,i}^{(t)},\;
\mathbf{k}_{X,j}^{(t)} + \mathbf{k}_{H,j}^{(t)} \rangle\Bigr)}
{\sum_{m=1}^{n_{\text{patch}}}
\exp\Bigl(\langle \mathbf{q}_{X,i}^{(t)} + \mathbf{q}_{H,i}^{(t)},\;
\mathbf{k}_{X,m}^{(t)} + \mathbf{k}_{H,m}^{(t)} \rangle\Bigr)}.
\end{equation}
We then compute the output representation as
\begin{equation}
\label{eq:add_z}
\mathbf{z}_i^{(t)} \;=\; \mathbf{x}_i^{(t)}
\;+\; \sum_{j=1}^{n_{\text{patch}}}
\alpha_{i,j}^{(t)} \Bigl(\mathbf{v}_{X,j}^{(t)} \;+\; \mathbf{v}_{H,j}^{(t)}\Bigr).
\end{equation}
In this additive design, features from the visual inputs and the mnemonic percept patches \(\{\mathbf{h}_i^{(t)}\}\) both contribute to the self-attention mechanism by modifying the inner product in the numerator of~\eqref{eq:add_alpha} and by merging the corresponding values in~\eqref{eq:add_z}. 

\subsubsection{Multiplicative Feedback from Visual Working Memory}

To incorporate multiplicative feedback, we instead define:
\begin{equation}
\label{eq:mult_alpha}
\alpha_{i,j}^{(t)} \;=\; 
\frac{\exp\Bigl(\langle \mathbf{q}_{X,i}^{(t)} \odot \mathbf{q}_{H,i}^{(t)},\;
\mathbf{k}_{X,j}^{(t)} \odot \mathbf{k}_{H,j}^{(t)} \rangle\Bigr)}
{\sum_{m=1}^{n_{\text{patch}}}
\exp\Bigl(\langle \mathbf{q}_{X,i}^{(t)} \odot \mathbf{q}_{H,i}^{(t)},\;
\mathbf{k}_{X,m}^{(t)} \odot \mathbf{k}_{H,m}^{(t)} \rangle\Bigr)},
\end{equation}
and
\begin{equation}
\label{eq:mult_z}
\mathbf{z}_i^{(t)} \;=\;
\mathbf{x}_i^{(t)}
\;+\; \sum_{j=1}^{n_{\text{patch}}}
\alpha_{i,j}^{(t)} 
\Bigl(\mathbf{v}_{X,j}^{(t)} \;\odot\; \mathbf{v}_{H,j}^{(t)}\Bigr).
\end{equation}
Here, the top-down feedback pathway multiplicatively gates the bottom-up signals. As a result, larger (smaller) magnitudes in the memory pathway can amplify (suppress) the corresponding magnitudes in the immediate visual pathway. This scheme allows for more direct \emph{control} (through multiplication) of attention weights and context vectors, enabling stronger or weaker gating of specific patches.

\begin{figure}[H]
    \centering
    \includegraphics[width=0.99\linewidth]{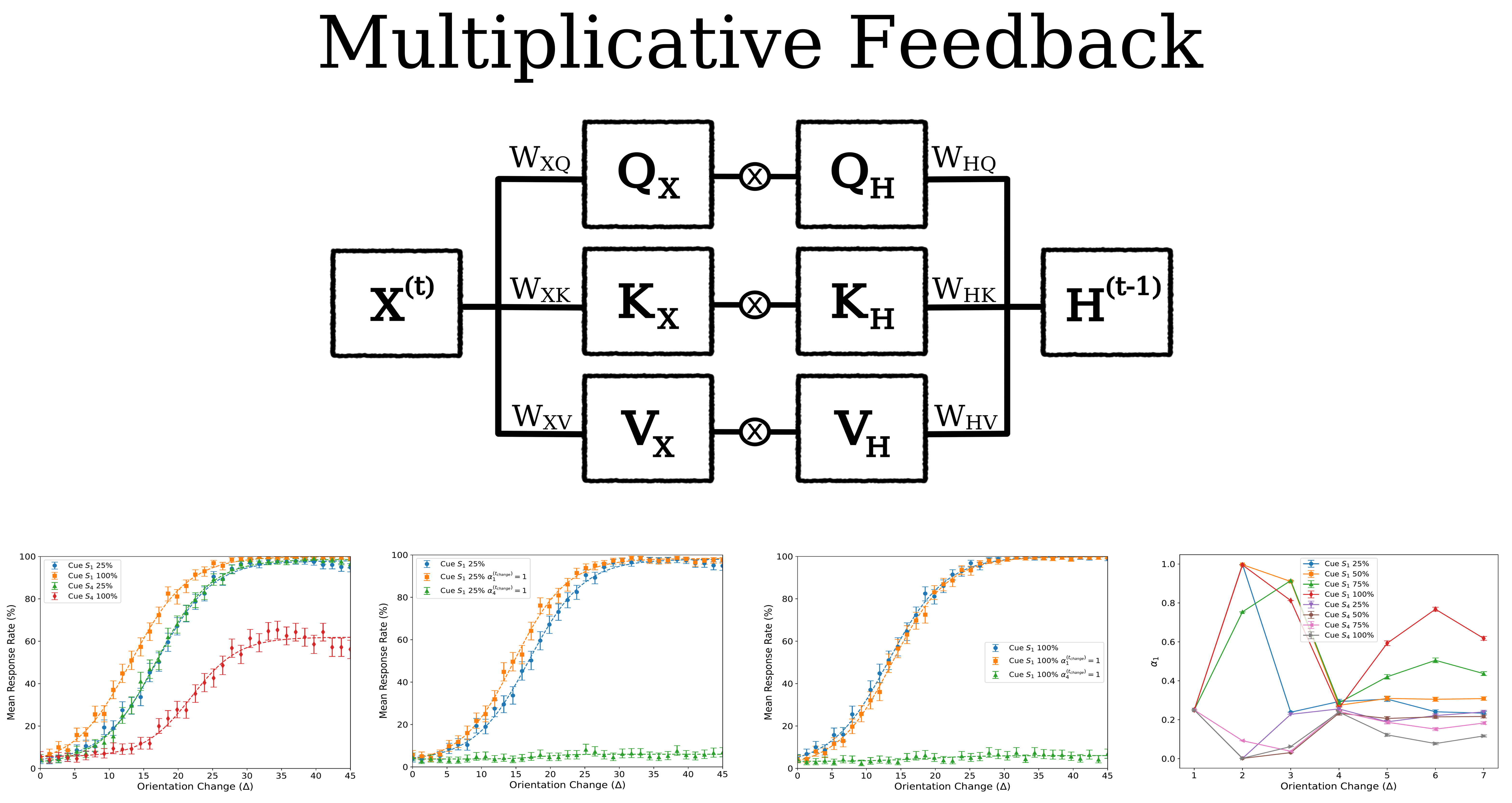}
    \captionsetup{justification=raggedright,singlelinecheck=false}
    \caption{\small Circuit diagram and some behavioral results for a model in which recurrent feedback is implemented via the element-wise multiplication of parallel projections onto the Q, K, V self-attention components. This model shows the strongest cue modulation and the strongest attention modulation effects on behavior. In addition, the cue has a strong influence on the deployment of attention at the time of change. These factors contribute to our choice of this model as the focus of our analysis.}
    \label{fig:QKVs}
\end{figure}

\vspace{2mm}
\noindent

Additive operations can be dominated by whichever pathway has a larger magnitude, potentially diminishing subtler signals. By contrast, multiplicative modulation can act as a direct ``sign-flip'' mechanism or a global rescaling factor, making it inherently well-suited for precise top-down control. For instance, consider a scenario in which \(\mathbf{q}_{X,i}\) or \(\mathbf{k}_{X,j}\) contain elements \(\pm 2\). A feedback mechanism that must flip selected signs via \emph{addition} could require large compensatory values in \(\mathbf{q}_{H,i}\) or \(\mathbf{k}_{H,j}\). In contrast, a multiplicative pathway can achieve such sign flips with a scalar factor of \(-1\), regardless of the original magnitude in \(\mathbf{q}_{X,i}\) or \(\mathbf{k}_{X,j}\).

\section{Model Architecture}

Our model integrates a Vision Transformer (ViT) with a patch-based LSTM. First, a VAE is used to preprocess the raw visual features in a purely feed-forward method. Secondly, we utilize a recurrent ViT in which self-attention has been modified to incorporate immediate and recurrent inputs in order to construct the visual percept transmitted to the patch-based LSTM. Thirdly, the LSTM utilizes the projection from the recurrent ViT to update the patch-based internal states. 

\subsection{VAE Pre-Processing}

A Variational Autoencoder (VAE) is a generative model that learns to encode input data into a latent space and reconstructs the data from this latent representation. It combines principles from deep learning and probabilistic inference, making it suitable for modeling complex data distributions. It consists of two primary components, and encoder ($F$) that encodes visual inputs to a probabilistic latent space ($z_{latent}$), and a decoder ($G$) that decodes a sampled latent vector into a visual input.  

The encoder network $F$ entails multiple operations, $f\in F$ which serve to map an input image patch $\mathbf{o}_i \in \mathbb{R}^{H_{patch} \times W_{patch} \times C}$ to a latent representation characterized by a mean vector $\boldsymbol{z_{\mu}} \in \mathbb{R}^{d_{latent}}$ and a log-variance vector $\boldsymbol{z_{logvar}} \in \mathbb{R}^{d_{latent}}$, where $d_{latent}$ is the dimensionality of the latent space. The encoder consists of convolutional and fully connected layers as follows:

\begin{enumerate}
    \item \textbf{First Convolutional Layer}: Applies a convolution with 16 filters, each of size $3 \times 3$, stride 2, and padding 1. This operation reduces the spatial dimensions while increasing the feature depth. The activation function is ReLU:
    \[
    \mathbf{z_{Conv,1}} = \mathrm{ReLU}\left(f_{Conv,1}^{(1,16,3,2,1)}(\mathbf{o_i}) \right)
    \]
    \item \textbf{Second Convolutional Layer}: Applies a convolution with 32 filters, each of size $3 \times 3$, stride 2, and padding 1:
    \[
    \mathbf{z_{Conv,2}} = \mathrm{ReLU}\left( f_{Conv,2}^{(16,32,3,2,1)}(\mathbf{z_{Conv,1}}) \right)
    \]
    \item \textbf{Flattening}: The output tensor is reshaped into a vector:
    \[
    \mathbf{z_{flat,1}} = \mathrm{Flatten}(\mathbf{z_{Conv,2}})
    \]
    \item \textbf{First Fully Connected Layer}: Maps the flattened vector to a 128-dimensional feature vector:
    \[
    \mathbf{z_{flat,2}} = \mathrm{ReLU}\left( \mathbf{W}_1 \mathbf{z}_\text{flat} + \mathbf{b}_1 \right)
    \]
    \item \textbf{Latent Variable Parameters}: Computes the mean and log-variance vectors using two separate linear transformations:
    \[
    \boldsymbol{z_\mu} = \mathbf{W}_\mu \mathbf{z_{flat,2}} + \mathbf{b}_\mu, \quad \boldsymbol{z_{\log\sigma^2}} = \mathbf{W}_{\log\sigma^2} \mathbf{z_{flat,2}} + \mathbf{b}_{\log\sigma^2}
    \]
\end{enumerate}

To allow gradient-based optimization through stochastic sampling, we employ the reparameterization trick. Letting $\mu = \mathbf{z_\mu}$ and $\sigma = \exp(0.5 \mathbf{z_{logvar}})$ we draw a latent vector $\mathbf{z_{latent}}$ from the approximate posterior:
\[
\mathbf{z_{latent}} = \boldsymbol{\mu} + \boldsymbol{\sigma} \odot \boldsymbol{\epsilon}, \quad \boldsymbol{\epsilon} \sim \mathcal{N}(\mathbf{0}, \mathbf{I})
\]
where $\boldsymbol{\sigma} = \exp\left( \frac{1}{2} \boldsymbol{\log\sigma^2} \right)$, and $\odot$ denotes element-wise multiplication.

The decoder network $G$ maps the latent vector $\mathbf{z}$ back to the reconstructed image $\hat{\mathbf{o}_i}$. The decoder mirrors the encoder but uses transposed convolutions:

\begin{enumerate}
    \item \textbf{First Fully Connected Layer}: Transforms the latent vector to a 128-dimensional vector:
    \[
    \mathbf{\hat{o}_{flat,1}} = \mathrm{ReLU}\left( \mathbf{W}_{flat,1} \mathbf{z_{latent}} + \mathbf{b}_{flat,1} \right)
    \]
    \item \textbf{Second Fully Connected Layer}: Maps the 128-dimensional vector to a shape suitable for convolutional layers:
    \[
    \mathbf{\hat{o}_{flat,2}} = \mathrm{ReLU}\left( \mathbf{W}_{flat,2} \mathbf{\hat{o}_{flat,1}} + \mathbf{b}_{flat,2} \right)
    \]

    \item \textbf{First Transposed Convolutional Layer}: Applies a transposed convolution with 16 filters:
    \[
    \mathbf{\hat{o}_{ConvT,1}} = \mathrm{ReLU}\left( g_{ConvT,1}^{(32,16,3,2,1,0)}(\mathbf{\hat{o}_{flat,2}}) \right)
    \]
    \item \textbf{Second Transposed Convolutional Layer}: Applies a transposed convolution to reconstruct the image:
    \[
    \mathbf{\hat{o}_{ConvT,2}} = \textrm{Sigmoid}\left( g_{ConvT,2}^{(16,1,3,2,1,0)}(\mathbf{\hat{o}_{ConvT,1}}) \right)
    \]
\end{enumerate}

The VAE optimizes a loss function that combines reconstruction accuracy and the Kullback-Leibler (KL) divergence between the approximate posterior and the prior distribution. Letting $\mathbf{\hat{o}_i} = \mathbf{\hat{o}_{ConvT,2}}$ the loss is described as:
\[
\mathcal{L} = \frac{1}{d_{image}} \| \mathbf{o_i} - \hat{\mathbf{o_i}} \|^2 - \beta \cdot \frac{1}{2} \left( 1 + \log \boldsymbol{\sigma_i}^2 - \boldsymbol{\mu_i}^2 - \boldsymbol{\sigma_i}^2 \right)
\]
where $\beta$ is a hyperparameter that balances the two terms, i represents the image patch number, and $d_{image} = H_{patch} \times W_{patch} \times C$

\subsection{ViT}
Input images $\mathbf{O}^{(t)} \in \mathbb{R}^{50 \times 50}$ are sub-divided into four equal patches $\{\mathbf{o_1}^{(t)}, \mathbf{o_2}^{(t)}, \mathbf{o_3}^{(t)}, \mathbf{o_4}^{(t)}\}$, with $\mathbf{o_i}^{(t)}\in\mathbb{R}^{(25\times25)}$. We found that our RL agent learned fastest, was most interpretable, and demonstrated best performance when we used the second flattend encoder layer ($\mathbf{o_{flat,2}}$) as input to the ViT (as oppose to the latent encoding). Hence, for a given patch input $\mathbf{o_i}^{(t)}$ at time $t$, we have the encoding
\begin{equation*}
    \mathbf{\hat{o}_i}^{(t)} = f^*(\mathbf{o_i}^{(t)})
\end{equation*}
where $f^*(\cdot)$ includes encoder components (1)--(4). We also concatenate a (one-hot) positional ($\boldsymbol{\rho_i}$) and temporal  ($\boldsymbol{\tau}$) encoding. Thus the full pre-processed patch input at timestep $t$ is
\begin{equation}
    \mathbf{x_i}^{(t)} = \text{Concat}[\mathbf{\hat{o}_i}^{(t)}, \boldsymbol{\rho_i}, \boldsymbol{\tau}]
\end{equation}

The complete input to the ViT at time step $t$ is:
\begin{equation}
    \mathbf{X}^{(t)}= (\mathbf{x_1}^{(t)}, \mathbf{x_2}^{(t)}, \mathbf{x_3}^{(t)}, \mathbf{x_4}^{(t)})^T \in \mathbb{R}^{4 \times 140}
\end{equation}
The transformer computes queries, keys, and values as:
\begin{align}
    \mathbf{Q} &= (\mathbf{X^{(t)} W_{XQ}}) \odot (\mathbf{H^{(t-1)} W_{HQ}}) \\
    \mathbf{K} &= (\mathbf{X^{(t)} W_{XK}}) \odot (\mathbf{H^{(t-1)} W_{HK}}) \\
    \mathbf{V} &= (\mathbf{X^{(t)} W_{XV}}) \odot (\mathbf{H^{(t-1)} W_{HV}})
\end{align}
where $\mathbf{W_{X\cdot}} \in \mathbb{R}^{140 \times 140}$, $\mathbf{W_{H\cdot}} \in \mathbb{R}^{1024 \times 140}$, $\mathbf{H}^{(t-1)}$ is the activated memory from the previous timestep, $\odot$ denotes Hadamard product, and we have dropped the temporal superscript (implicit). Self-attention is computed as:
\begin{equation}
    \mathbf{V_{{filtered}}} = \text{Softmax}(\mathbf{QK}^T)\mathbf{V}
\end{equation}
The spatially and temporally aware visual percept is constructed as follows:
\begin{equation}
    \mathbf{Z}^{(t)} = \mathbf{X}^{(t)} + \mathbf{V_{filtered}} \in \mathbb{R}^{4 \times 140}
\end{equation}

\subsection{Spatial LSTM}

We adapt the xLSTM architecture \cite{beck2024xlstm} for spatial memory. The LSTM operations are:

{\small
\begin{align*}
    \mathbf{\tilde{I}}^{(t)} &= \mathbf{Z}^{(t)} \mathbf{W_i} + \mathbf{H}^{(t-1)} \mathbf{R_i} & \mathbf{I}^{(t)} &= \exp(\mathbf{\tilde{I}}^{(t)} - \mathbf{M}^{(t)}) & \mathbf{O}^{(t)} &= \sigma(\mathbf{\tilde{O}}^{(t)}) \\
    \mathbf{\tilde{F}}^{(t)} &= \mathbf{Z}^{(t)} \mathbf{W_f} + \mathbf{H}^{(t-1)} \mathbf{R_f} & \mathbf{F}^{(t)} &= \exp(\mathbf{\tilde{F}}^{(t)} + \mathbf{M}^{(t-1)} - \mathbf{M}^{(t)}) & \mathbf{N}^{(t)} &= \mathbf{F}^{(t)} \odot \mathbf{N}^{(t-1)} + \mathbf{I}^{(t)} \\
    \mathbf{\tilde{O}}^{(t)} &= \mathbf{Z}^{(t)} \mathbf{W_o} + \mathbf{H}^{(t-1)} \mathbf{R_o} & \mathbf{M}^{(t)} &= \max(\mathbf{\tilde{F}}^{(t)} + \mathbf{M}^{(t-1)}, \mathbf{\tilde{I}}^{(t)}) & \mathbf{U}^{(t)} &= \tanh(\mathbf{\tilde{U}}^{(t)}) \\
    \mathbf{\tilde{U}}^{(t)} &= \mathbf{Z}^{(t)} \mathbf{W_u} + \mathbf{H}^{(t-1)} \mathbf{R_z} & \mathbf{C}^{(t)} &= \mathbf{C}^{(t-1)} \odot \mathbf{F}^{(t)} + \mathbf{U}^{(t)} \odot \mathbf{I}^{(t)} & \mathbf{H}^{(t)} &= \mathbf{O}^{(t)} \odot (\mathbf{C}^{(t)} / \mathbf{N}^{(t)})
\end{align*}
}
where $\mathbf{W_{x}} \in \mathbb{R}^{140 \times 1024}$, , $\mathbf{R_{x}} \in \mathbb{R}^{140 \times 1024}$, and all other variables $\in \mathbb{R}^{4 \times 1024}$. As described above, we call this a patch-based LSTM because there is a hidden state for each patch of the visual scene. Importantly, within the LSTM the hidden states are updated independently. The matrices $\mathbf{Z}^{(t)}$, $\mathbf{C}^{(t)}$, $\mathbf{H}^{(t)}$, $\mathbf{M}^{(t)}$, and $\mathbf{N}^{(t)}$ are of shape $n_{patch}$ by $d$, where $d\in{d_{latent}, d_{mem}}$. Right multiplication by the matrices $\mathbf{W_x}$ or $\mathbf{R_x}$ projects the latent embedding or hidden state of a specific patch to another space, independent of the other patches. By constructions, self-attention is the only mechanism by which information from visual patches (or mnemonic patches) is communicated to other patches. 

\subsection{Actor-Critic Network}

The mnemonic percept $\mathbf{H}^{(t)} \in \mathbb{R}^{4 \times 1024}$ serves as input to both actor and critic networks. The actor network is a 4-layer feed-forward neural network:
\begin{align}
    \mu_1 &= \text{ELU}(H' W_1 + b_1) \\
    \mu_2 &= \text{ELU}(\mu_1 W_2 + b_2) \\
    \mu_3 &= \text{ELU}(\mu_2 W_3 + b_3) \\
    \pi_\theta(a_t|H_t) &= \text{Softmax}(\mu_3 W_{\text{out}} + b_{\text{out}})
\end{align}
where $H' \in \mathbb{R}^{4096}$ is the flattened $H_t$. The network dimensions decrease from 4096 to 2, with the output representing the action distribution.

The critic network maps $(H_t, a_t)$ to a distributional Q-function:
\begin{align}
    a' &= a_t W_a + b_a \\
    q_0 &= \text{Concat}[H', a'] \\
    q_1 &= \text{ELU}(q_0 W_1 + b_1) \\
    q_2 &= \text{ELU}(q_1 W_2 + b_2) \\
    q_3 &= \text{ELU}(q_2 W_3 + b_3) \\
    p_\theta(q|H_t, a_t) &= \text{Softmax}(q_3 W_{\text{out}} + b_{\text{out}})
\end{align}
where $q_0 \in \mathbb{R}^{8192}$, and the output dimension is 15, representing discretized Q-values. The model is trained using a KL-regularized reinforcement learning objective:
\begin{equation}
    L_Q(\theta) = \mathbb{E}_{D} [D_{\text{KL}}[\pi_{\text{imp}}, \pi_\theta | s_t, \tilde{\pi} = \pi_{\theta'}] + \beta D_{\text{KL}}[\Gamma_{\theta'}(q|s_t, a_t), p_\theta(q|s_t, a_t)]]
\end{equation}
Here, $\pi_{\text{imp}}$ is the improved policy given by:
\begin{equation}
    \pi_{\text{imp}}(a_t|s_t) \propto \exp(Q_{\theta'}(s_t, a_t)/\eta)\pi_{\theta'}(a_t|s_t)
\end{equation}
$\Gamma_{\theta'}(q|s_t, a_t)$ is the target Q-distribution computed using the distributional Bellman operator:
\begin{equation}
    \Gamma_{\theta'}(q|s_t, a_t) = \mathbb{E}_{s_{t+1}} \mathbb{E}_{a' \sim \pi_\theta(\cdot|s_{t+1})} \mathbb{E}_{q' \sim p_\theta(\cdot|s_{t+1}, a')} [\mathbf{1}_{[q-\epsilon/2, q+\epsilon/2]}(r_t + \gamma q')]
\end{equation}
The loss function balances policy improvement (first KL term) with Q-function learning (second KL term). The hyperparameter $\beta$ controls the trade-off between these objectives. This formulation allows for offline reinforcement learning without explicit behavior cloning, relying instead on the KL-regularization to the previous policy $\pi_{\theta'}$ to stabilize learning.

\subsection{Reinforcement Learning}

To emulate learning processes observed in non-human primates (NHPs) and humans, our model is trained using a reinforcement learning (RL) framework~\cite{sutton2018reinforcement}. In this framework, the agent interacts with its environment by observing visual stimuli and taking actions to maximize the expected cumulative rewards over time. The goal of the RL framework is to enable the model to learn optimal policies that maximize future rewards based on the agent's perceptual inputs and previous experiences. The input to the RL module is the mnemonic percept $H^{(t)}$, which encapsulates the relevant features of the visual scene as represented in the working memory module. This activated memory serves as the input to both the action-selection policy $\pi(H^{(t)})$ and the value function $V_\pi(H^{(t)})$, both of which are parameterized by neural networks in our model.

The action-selection policy $\pi(H^{(t)})$ maps the current activated memory to an action that the agent will take at time $t$. In our environment, the agent has two possible actions:
\begin{align*}
    \pi(H^{(t)}) &= 0 \quad \text{(``wait'' action)}, \\
    \pi(H^{(t)}) &= 1 \quad \text{(``declare change'' action)}.
\end{align*}
The ``wait'' action implies that the agent decides not to make a response and continues to process further information from the environment. The ``declare change'' action represents the agent's decision to identify a change in the visual stimulus. The policy network is trained to maximize the expected future rewards by choosing the action that is predicted to have the highest value. The value function $V_\pi(H^{(t)})$ estimates the expected cumulative future reward from the current activated memory $H^{(t)}$:
\begin{equation}
    V(H^{(t)}) = \mathbb{E}\left[ \sum_{\tau = t}^{T} \gamma^{\tau - t} r_\tau \,\bigg|\, H^{(t)} \right],
    \label{eq:V}
\end{equation}
where $\gamma \in [0,1]$ is the discount factor, $r_\tau$ is the reward received at time step $\tau$, and $T$ is the terminal time step for the task. The value function predicts how much reward the agent expects to receive by following its learned policy $\pi$ from the current activated memory.

Learning in the RL module is driven by the temporal difference (TD) error, which measures the difference between the predicted value and the actual reward received at each time step:
\begin{equation}
    \delta_t = r_t + \gamma V(H^{(t+1)}) - V(H^{(t)}),
    \label{eq:td_error}
\end{equation}
where $r_t$ is the reward received at time $t$ and $\delta_t$ is the TD error. This error signal is used to update the value function $V_\pi(H^{(t)})$ and the action-selection policy $\pi(H^{(t)})$ to better predict future rewards and make more optimal decisions. In our model, both the value function and the policy are parameterized by neural networks. The agent's performance improves over time as it receives rewards and updates its predictions based on experience, thereby learning to allocate bias and make decisions that maximize cumulative rewards.

\subsection{Task Difficulty}

To control task difficulty, Gabor stimuli were corrupted with rotational "noise". Defnining $\theta^*_i$ as the "true" Gabor orientation for $S_i$, the  orientation in the input image shown to the agent is:
\[
\theta_i = \theta^*_i + \delta_{it}
\]
where $\delta_{it} \sim \mathcal{N}(0,\sigma)$ is the rotational noise at time step $t$. If the stimulus is selected for change, then at $t=5$ and $t=6$:
\[
\theta_i = \theta^*_i + \Delta + \delta_{it}
\]
The orientation noise parameter $\sigma$ is set to 5. The orientation change parameter $\Delta$ is a random variable drawn at the beginning of a change trial, with $\Delta \sim \textrm{U}(-k, k)$, where $k$ is adjusted based on the agent's performance, starting at $k=65$ and decreasing as performance improves to increase task difficulty.

\section{Logistic Function and Fitting Procedure}

To model the relationship between orientation change and response rates, we employed a logistic function of the form:

\begin{equation}
    f(x) = A + (1 - B) \frac{1}{1 + \exp(-C (x - D))}
\end{equation}

where \( x \) represents the magnitude of the orientation change, and the parameters \( A, B, C, D \) govern the shape and position of the logistic curve.

\begin{itemize}
    \item \( A \) represents the lower asymptote, capturing any baseline response rate unrelated to orientation change.
    \item \( B \) modulates the upper asymptote, accounting for deviations from a perfect detection rate.
    \item \( C \) controls the slope of the curve, determining the rate at which response probability transitions from low to high.
    \item \( D \) corresponds to the inflection point, the orientation change at which the response rate reaches its midpoint.
\end{itemize}

The logistic function was fitted to empirical response rate data by optimizing the parameters to minimize the discrepancy between the observed values and the model predictions. Confidence intervals for the response rates were estimated using a Bayesian credible interval approach based on Jeffreys' prior. The fitted curves provide a smooth characterization of the response behavior across different orientation change levels, allowing for a quantitative comparison across conditions.

\begin{table}[h]
    \centering
    \begin{tabular}{|c|c|c|c|c|c|}
        \hline
        \textbf{Cue} $S_1$ & \textbf{Change} & \textbf{A} & \textbf{B} & \textbf{C} & \textbf{D} \\ 
        \hline
        25\% & $S_1$ & $0.07\pm0.01$ & $0.07\pm0.01$ & $0.30\pm0.01$ & $13.00\pm0.11$ \\ 
        50\% & $S_1$ & $0.07\pm0.01$ & $0.07\pm0.01$ & $0.28\pm0.01$ & $11.89\pm0.14$ \\ 
        75\% & $S_1$ & $0.06\pm0.01$ & $0.05\pm0.01$ & $0.30\pm0.01$ & $10.50\pm0.13$ \\ 
        100\% & $S_1$ & $0.06\pm0.01$ & $0.06\pm0.01$ & $0.30\pm0.01$ & $9.83\pm0.13$ \\ 
        25\% & $S_4$ & $0.08\pm0.01$ & $0.07\pm0.01$ & $0.27\pm0.01$ & $12.66\pm0.10$ \\ 
        50\% & $S_4$ & $0.08\pm0.01$ & $0.08\pm0.01$ & $0.28\pm0.01$ & $13.18\pm0.12$ \\ 
        75\% & $S_4$ & $0.09\pm0.01$ & $0.09\pm0.01$ & $0.27\pm0.01$ & $14.83\pm0.17$ \\ 
        100\% & $S_4$ & $0.10\pm0.02$ & $0.24\pm0.02$ & $0.29\pm0.03$ & $17.49\pm0.34$ \\ 
        \hline
    \end{tabular}
    \caption{Table showing logistic fit parameters for psychometric functions of the trained agent. Cues are always at the $S_1$ location. Standard error in parameter estimates are shown.}
    \label{tab:NoMod}
\end{table}

\begin{table}[h]
    \centering
    \begin{tabular}{|c|c|c|c|c|c|c|}
        \hline
        \textbf{Cue} $S_1$ & \textbf{Change} & $\mathbf{x}$ & \textbf{A} & \textbf{B} & \textbf{C} & \textbf{D} \\ 
        \hline
        25\% & $S_1$ & 1 & $0.06\pm0.01$ & $0.05\pm0.01$ & $0.29\pm0.01$ & $10.69\pm0.15$ \\ 
        25\% & $S_1$ & 4 & $0.12\pm0.02$ & $0.91\pm0.02$ & $0.20\pm0.09$ & $15.64\pm2.63$ \\ 
        100\% & $S_1$ & 1 & $0.07\pm0.01$ & $0.07\pm0.01$ & $0.28\pm0.01$ & $9.15\pm0.13$ \\ 
        100\% & $S_1$ & 4 & $0.06\pm0.01$ & $0.92\pm0.01$ & $0.28\pm0.10$ & $14.89\pm1.41$ \\ 
        100\% & $S_4$ & 1 & $0.12\pm0.04$ & $0.07\pm0.08$ & $0.27\pm0.07$ & $12.66\pm5.41$ \\ 
        100\% & $S_4$ & 4 & $0.05\pm0.01$ & $0.06\pm0.02$ & $0.27\pm0.01$ & $13.42\pm0.18$ \\ 
        \hline
    \end{tabular}
    \caption{Table showing logistic fit parameters for psychometric functions where we have artificially increased self-attention scores $\alpha_x^{(t_{change})}$ on either $S_1$ ($x=1$) or $S_4$ ($x=4$) at the time of change ($t_{change}$). Cues are at the $S_1$ location. Standard error in parameter estimates are shown. }
    \label{tab:MOD}
\end{table}





\section{Decoding Analysis}

\subsection{Decoder Architecture}

To extract and interpret the information encoded within our patch-based LSTM model, we implemented a decoder architecture. This decoder is designed to process the output from various layers of the patch-based LSTM and produce task-relevant predictions. The architecture consists of a feed-forward neural network with the following structure:
\begin{equation}
    \text{Decoder}(x) = f_3(f_2(f_1(x)))
\end{equation}
where $x \in \mathbb{R}^{d_\text{in}}$ is the input vector (typically a flattened output from the patch-based LSTM), and $f_1$, $f_2$, and $f_3$ are layer functions defined as:

\begin{align}
    f_1(x) &= \text{ELU}(\text{LN}_1(W_1x + b_1)) \\
    f_2(x) &= \text{ELU}(\text{LN}_2(W_2x + b_2)) \\
    f_3(x) &= W_3x + b_3
\end{align}
Here, $W_1 \in \mathbb{R}^{512 \times d_\text{in}}$, $W_2 \in \mathbb{R}^{256 \times 512}$, and $W_3 \in \mathbb{R}^{d_\text{out} \times 256}$ are weight matrices, $b_1$, $b_2$, and $b_3$ are bias vectors, $\text{LN}_1$ and $\text{LN}_2$ are layer normalization operations, and $\text{ELU}$ is the Exponential Linear Unit activation function. The final output of the decoder is a vector in $\mathbb{R}^{d_\text{out}}$, with $d_\text{out}$ depending on the specific decoding task.

\subsection{Decoding Analysis of patch-based LSTM Components}

\subsubsection{Decoding from the Complete mnemonic percept}

The mnemonic percept in our patch-based LSTM, denoted as $H^{(t)} \in \mathbb{R}^{4 \times 1024}$, comprises four slots corresponding to the four image patches in our visual field. To analyze the information content of this hidden state, we flatten $H^{(t)}$ to $H_\text{flat} \in \mathbb{R}^{4096}$ and train a decoder to predict the location of the orientation change ($S_1$, $S_2$, $S_3$, or $S_4$). 
\begin{figure}[htbp]
    \centering
    \includegraphics[width=0.99\linewidth]{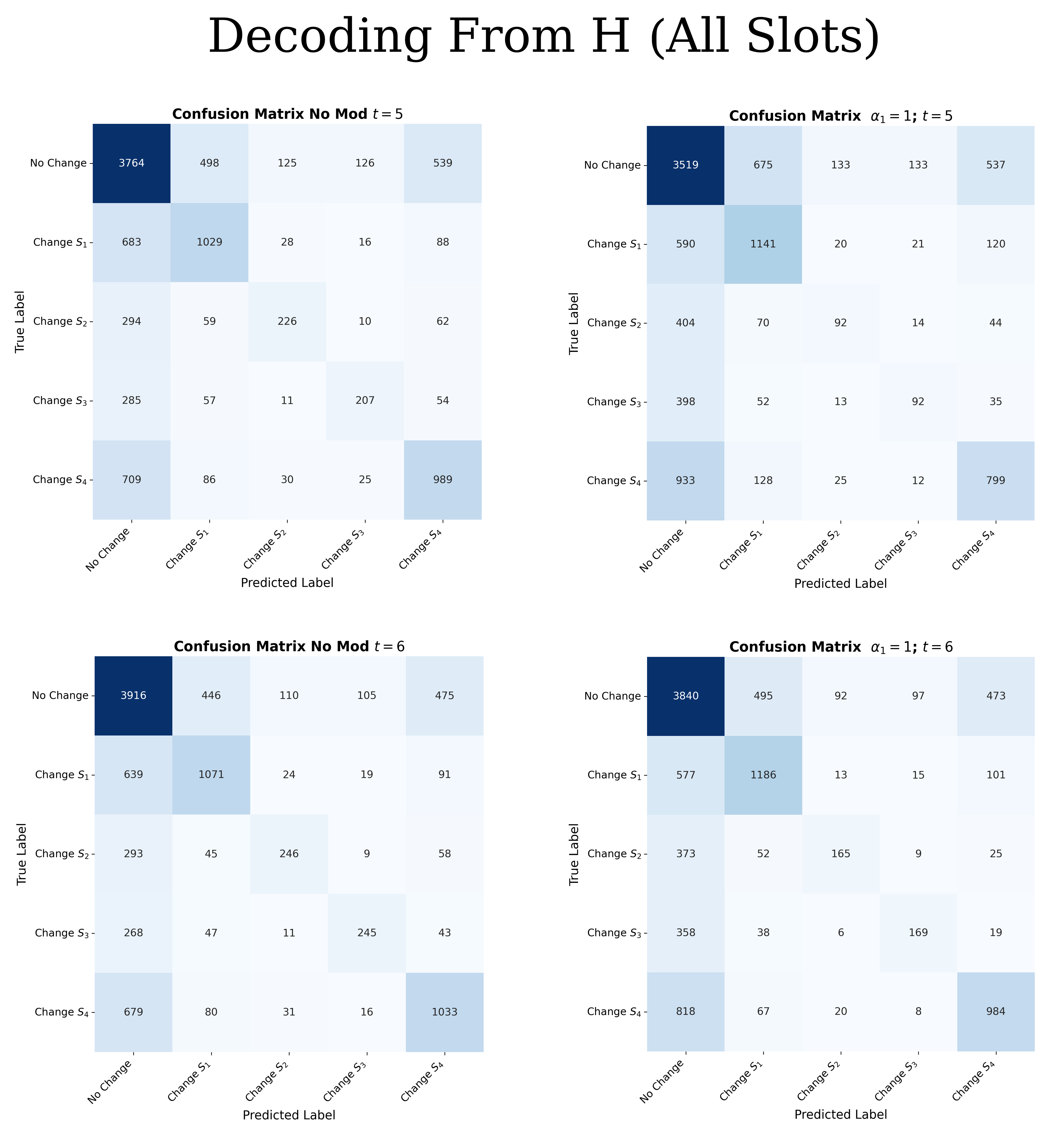}
    \caption{Confusion matrices for decoding change location from the mnemonic percept under various conditions.}
    \label{fig:ConfMatHAllSlots}
\end{figure}
Figure \ref{fig:ConfMatHAllSlots} presents confusion matrices for our decoder evaluated on a test set. The decoder was trained on data (the mnemonic percept $H^{(t)}$ at time $t$) generated from an environment identical to that of the RL agent's training, without artificial modulations. The results in Figure \ref{fig:ConfMatHAllSlots} demonstrate that the decoder can detect change locations, albeit with suboptimal accuracy. We observe several key phenomena. First, there is a temporal enhancement effect: increased time ($t=5$ to $t=6$) enhances change decodability from $H$ collected at later time points, suggesting a temporal integration of change information. Secondly, inhibiting attention at a stimulus location had the effect of reducing change decodability of said stimulus, but slightly increased decodability of other change locations ($S_1$ in Figure \ref{fig:ConfMatHAllSlots}, third column). Conversely, we found that directing attention towards a specific stimulus patch ($S_1$ in Figure \ref{fig:ConfMatHAllSlots}, right column) slightly improves decodability for changes in $S_1$ while marginally decreasing decodability for other patches, indicating a trade-off in representational capacity. 

\begin{figure}[htbp]
    \centering
    \includegraphics[width=0.99\linewidth]{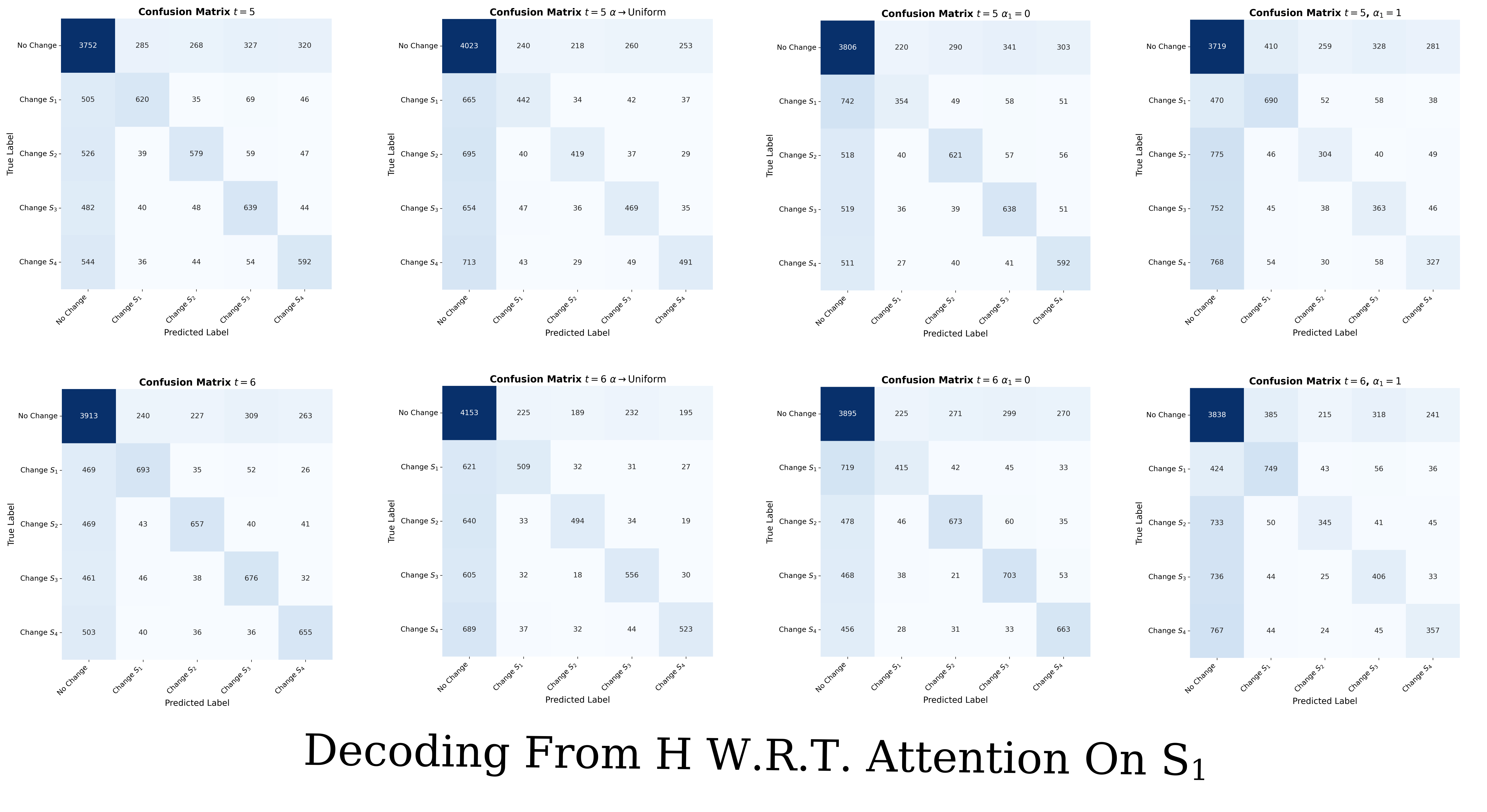}
    \caption{Confusion matrices for decoding change location with varying attention modulations on $S_1$.}
    \label{fig:ConfMatHAllSlotsCue025}
\end{figure}

To further investigate the impact of attentional modulation, we conducted an experiment with a forced 25\% cue at the $S_1$ location, indicating a 0.25 probability of a stimulus becoming the change stimulus, given a change trial. Figure \ref{fig:ConfMatHAllSlotsCue025} presents the results of this analysis. In Figure \ref{fig:ConfMatHAllSlotsCue025}, we observe several important effects. The first column shows normal decoding performance under the 25\% cue condition. In the second column, we enforced a uniform Self-Attention map $A$ ($\alpha_i = 0.25$ for all $i$), which leads to a slight bias in classifying $H$ as being derived from a no-change trial. The third column shows the effect of inhibiting attention on $S_1$ ($\alpha_1=0$), which results in an increased number of no-change classifications when $S_1$ is the change target. This demonstrates the importance of attention for change detection in the attended location. Finally, in the fourth column, we enforced maximum attention on $S_1$ ($\alpha_1=1$). This increased change classifications when $S_1$ was the change target but decreased change classification for other targets.

\subsubsection{Decoding from a Single Mnemonic Percept Patch}

\begin{figure}[htbp]
    \centering
    \includegraphics[width=0.99\linewidth]{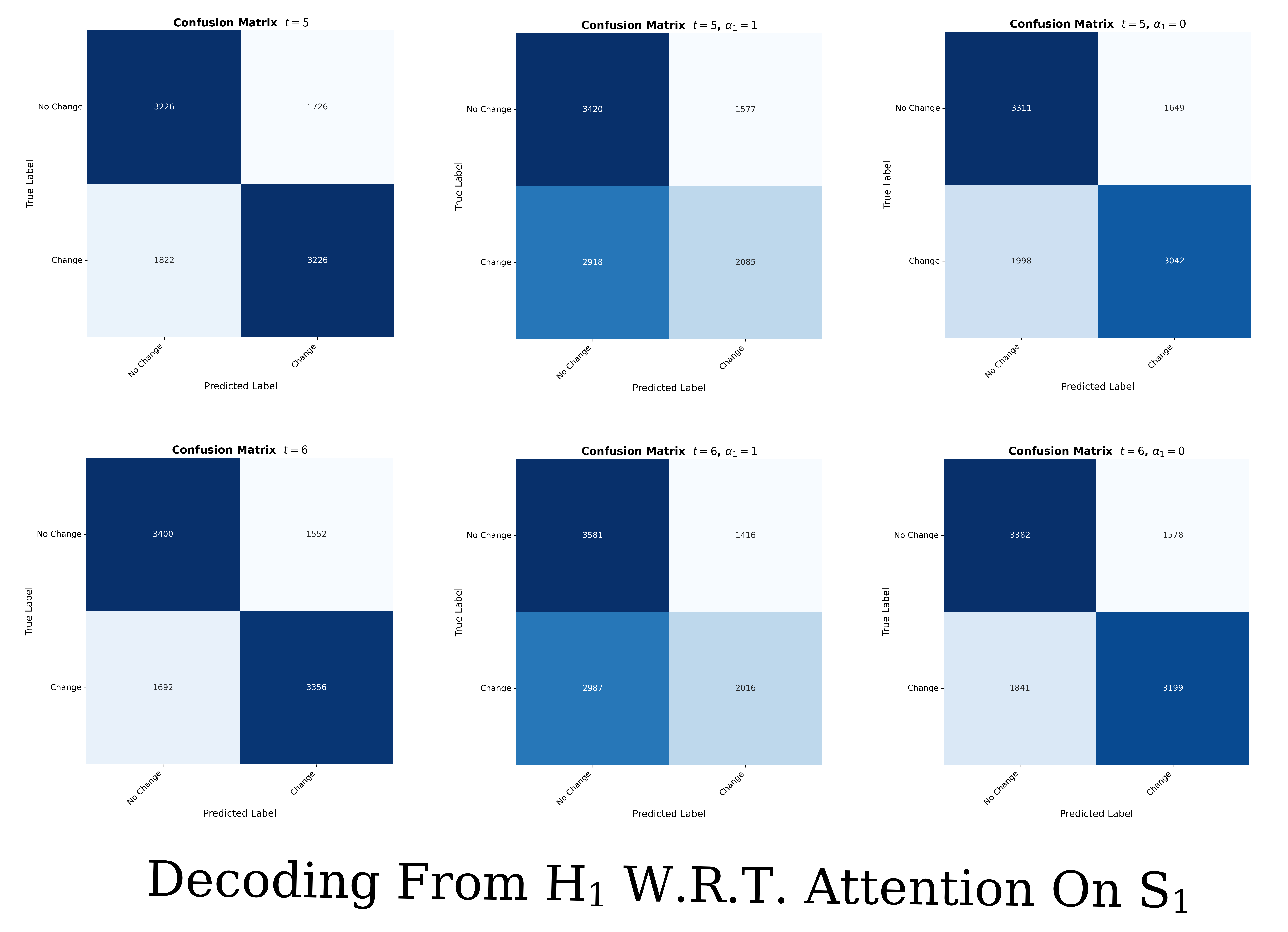}
    \caption{Confusion matrices for decoding change occurrence from the first mnemonic percept patch under varying attention modulations.}
    \label{fig:ConfMatFirstSlot}
\end{figure}

To investigate the information transfer between patches facilitated by self-attention, we decoded exclusively from the first mnemonic percept patch of $H^{(t)}$, $h^{(t)}_1 \in \mathbb{R}^{1024}$, corresponding to the $S_1$ location. It is important to note that the change location was randomly selected in all trials. Figure \ref{fig:ConfMatFirstSlot} presents the results of this analysis. Our results reveal several key findings. In the case where attention is not modulated (first column), the decoder can correctly classify the occurrence of change in the majority of trials just from the $h_1^{(t)}$ patch, indicating that a single mnemonic patch contains information about changes across all patches (the probability of a change occurring at $S_1$ is 0.2). When we force maximum attention on $S_1$ ($\alpha_1^{(t)}=1$, second column), we observe a loss of 'change' classifications on change trials (i.e., can only reliably decode the changes from the $S_1$ location). This suggests that maximal attention on $S_1$ suppresses the propagation of change signals from other patches to the $S_1$ slot. When we force no attention on $S_1$ ($\alpha_1=0$, not shown), the decodability of change from slot $S_1$ is only slightly affected. This is because $S_1$ changes can be decoded without attention due to direct access to $h_1^{(t)}$ patch information. These results demonstrate the crucial role of self-attention in propagating change information across memory slots.

\subsubsection{Decoding from the Actor Network's First Activation Layer}

\begin{figure}[htbp]
    \centering
    \includegraphics[width=0.99\linewidth]{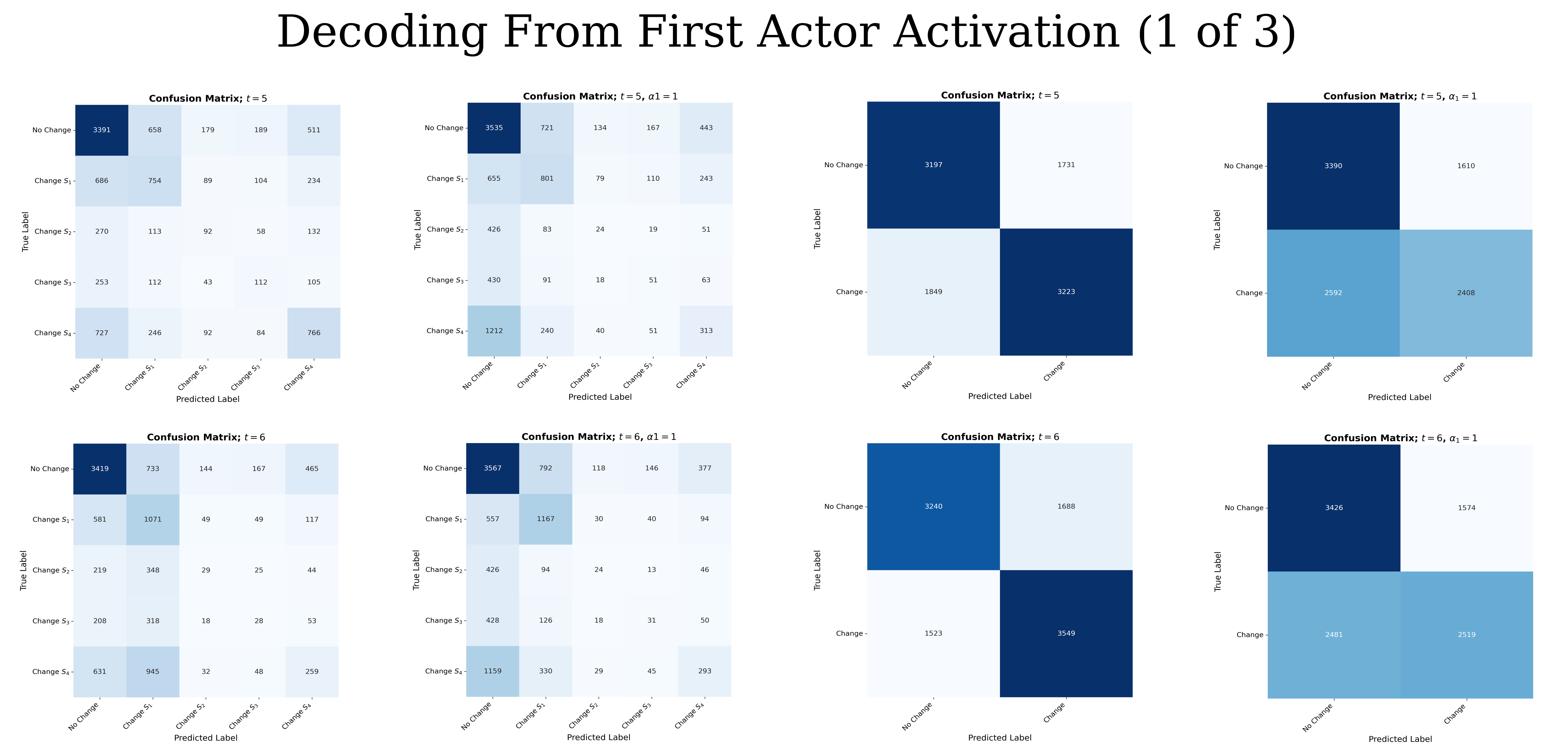}
    \caption{Confusion matrices for decoding change location and occurrence from the actor network's first activation layer.}
    \label{fig:ActorLayer1Activation}
\end{figure}

To understand how the actor network processes the information from the patch-based LSTM, we trained a decoder on the first activation layer of the actor network. Figure \ref{fig:ActorLayer1Activation} presents the results of this analysis. The results in Figure \ref{fig:ActorLayer1Activation} reveal several important aspects of information processing in the actor network. The left two columns show that much of the spatial information is lost in this first activation layer, as evidenced by the poor classification of orientation change locations compared to decoding from the hidden state (Figure \ref{fig:ConfMatHAllSlots}). This suggests a loss of spatial specificity in the actor network. However, the right two columns demonstrate that changes can still be successfully decoded from this layer, indicating that change information is preserved and potentially emphasized in the actor network.

When we force $\alpha_1=1$ (rightmost column), we observe a loss of change classification. Importantly, this is not due to the absence of the change signal (as it is still present in the memory slot associated with the change location) but rather due to a weaker presence of the change signal. This suggests that self-attention serves to amplify the change signal by propagating it across all memory slots.

\begin{figure}[htbp]
    \centering
    \includegraphics[width=0.99\linewidth]{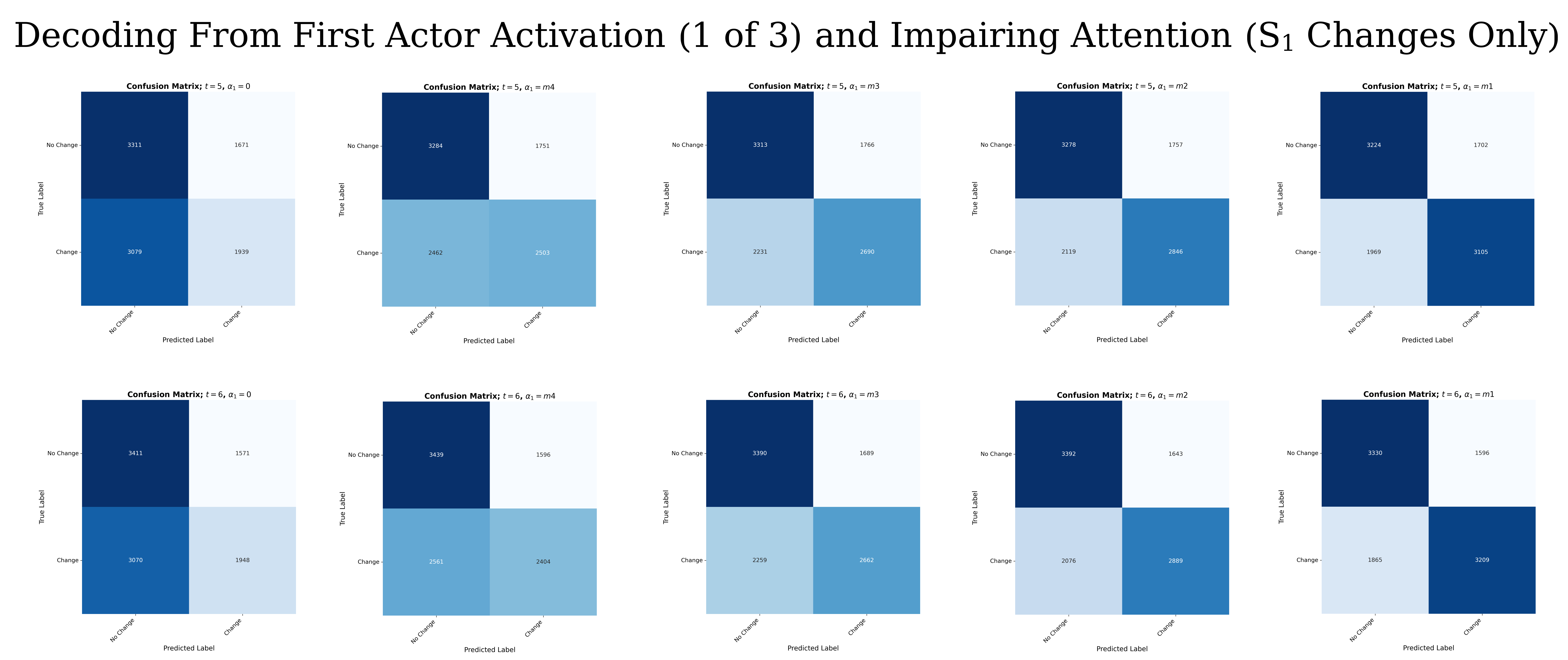}
    \caption{Confusion matrices for decoding $S_1$ changes from the actor network's first activation layer with varying levels of attention inhibition.}
    \label{fig:ActorLayer1ActivationS1Changes}
\end{figure}

To further investigate the impact of attentional modulation on change detection in the actor network, we conducted an additional experiment focusing only on $S_1$ changes with varying levels of attention inhibition. Figure \ref{fig:ActorLayer1ActivationS1Changes} presents these results.
In Figure \ref{fig:ActorLayer1ActivationS1Changes}, we observe a clear progression as the inhibition of attention is relaxed. When self-attention on $S_1$ is completely inhibited (left column), the model fails to classify the majority of changes. However, the change signal is not entirely eradicated, as evidenced by the non-zero number of change classifications. As inhibition decreases (right columns), the confusion matrices show increasingly accurate classification structures. This demonstrates the graded nature of attentional modulation in the detection of changes in the actor network.

\subsubsection{Analysis of Actor Network Logits}

\begin{figure}[htbp]
    \centering
    \includegraphics[width=0.99\linewidth]{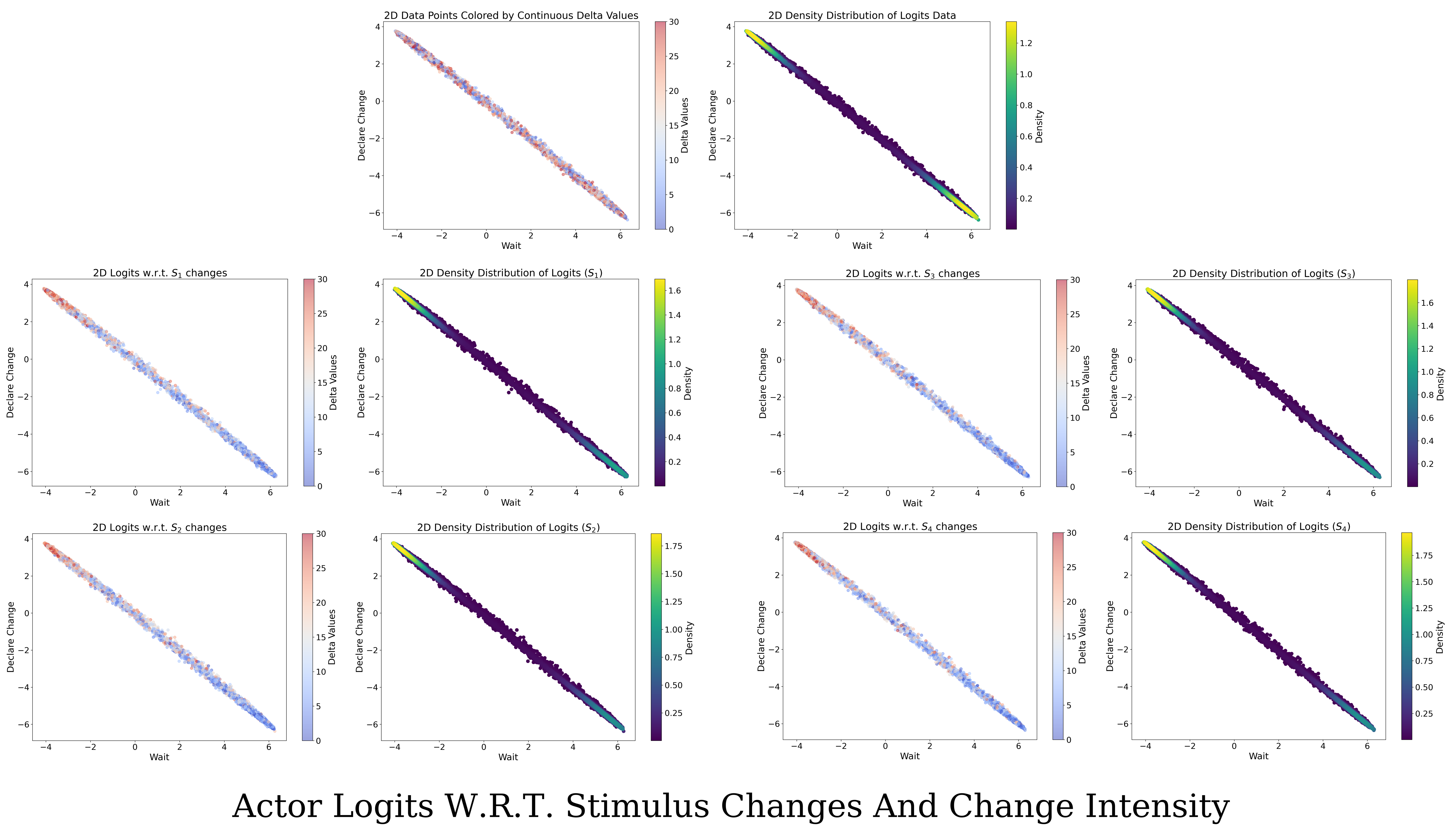}
    \caption{Relationship between 'Declare Change' and 'Wait' logits in the actor network's final layer.}
    \label{fig:LogitsChangeLocation}
\end{figure}

To gain insight into the decision-making process of the actor network, we analyzed the logits of its final layer (pre-softmax activation). Figure \ref{fig:LogitsChangeLocation} presents this analysis. Figure \ref{fig:LogitsChangeLocation} reveals several key aspects of the decision-making process of the actor network. First, we observe a clear inverse linear relationship between 'Declare Change' and 'Wait' logits, indicating a competitive decision-making process. Second, we note a strong dependence on change intensity: 'Wait' logits tend to be high when change intensity ($\Delta$) is low, while 'Declare Change' logits are high for large $\Delta$ values. This suggests that the actor network has learned to respond in a graded manner with respect to the change signal. Interestingly, the spatial location of the change does not show any obvious influence on this relationship, indicating that the actor network has learned to make decisions based primarily on change signal independent of change location.

\begin{figure}[htbp]
    \centering
    \includegraphics[width=0.99\linewidth]{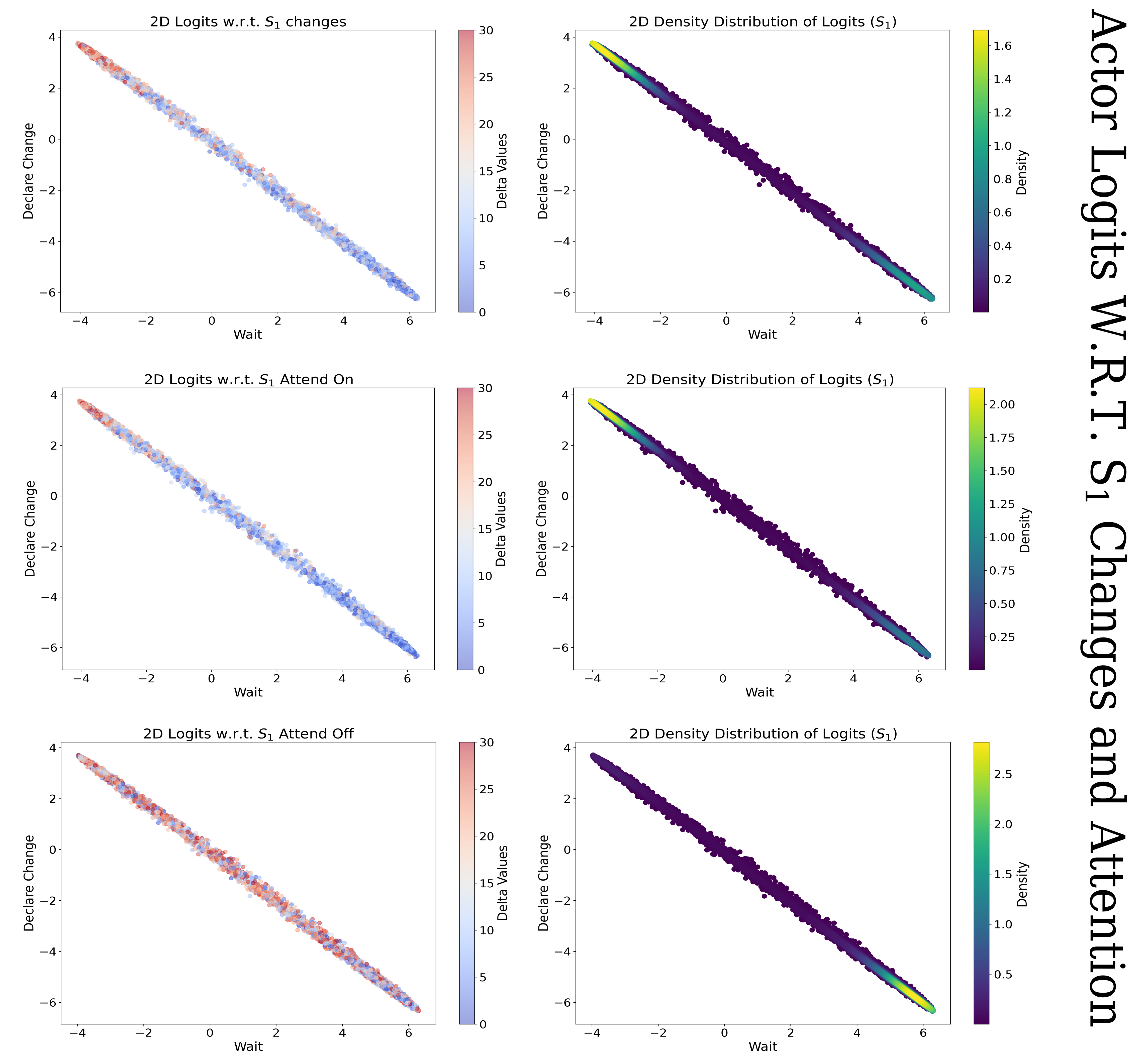}
    \caption{Effect of attentional modulation on actor network logits for $S_1$ changes.}
    \label{fig:LogitsS1Attention}
\end{figure}

To investigate the impact of attentional modulation on the actor network's decision-making process, we analyzed the logit structure under varying attention modulations for $S_1$ changes. Figure \ref{fig:LogitsS1Attention} presents these results. The results in Figure \ref{fig:LogitsS1Attention} demonstrate the significant impact of attentional modulation on the decision-making process of the actor network. Maximizing $\alpha_1$ slightly increases the density of points in the high 'Declare Change' logit space, suggesting that increased attention facilitates change detection. Conversely, minimizing $\alpha_1$ significantly decreases the density of logits in the high-valued \lq Declare Change \rq logit space and increases the density in the high-valued \lq Wait \rq logit space. This demonstrates that attentional withdrawal substantially impairs the network's ability to detect changes.

\newpage

\subsection{Self-Attention Amplifies Change Signals}

Let $C^{(t)} \in \mathbb{R}^{4 \times 1024}$ represent the mnemonic state of the Recurrent ViT at time $t$, where each row corresponds to a memory slot associated with one of the four visual patches. Additionally, let $k^*$ be the indexed location of the stimuli that experienced an orientation change, $S_{k^*}$, and $\psi(\cdot)$ be some measure of the strength of a change signal. While the many parameters and complexity make a closed form description of how a change signal is communicated among memory slots challenging, based on the above results and structure of the Recurrent ViT, we can characterize some key properties of the process. From the structure of the Recurrent ViT, we know that self-attention is the only mechanism that allows information to flow from one mnemonic patch to another. Thus, if  $i \neq k^*$ and $a_{i,k^*}^{(t)} = 0: \forall t \geq t_{change}$, then $\psi(c_i^{(t)}) = 0$, where $c_i^{(t)}$ is the memory patch associated with patch $i$. Additionally, we know that if $i = k^*$, then attention is not needed to propagate change information to the memory patch $c_i^{(t)}$, i.e., $\psi(c_i^{(t)};a_{i,k^*}^{(t)}) \approx \psi(c_i^{(t)}; 0): \forall t \geq t_{change}$. Finally, we know that if we apply this measure to all memory patches, $\psi(C^{(t)})$, we know that $\psi(C^{(t)}; \alpha_{k^*})$ increases as $\alpha_{k^*}$ increases, where 
\begin{align*}
    \alpha_{k^*} = \sum_{i=1}^4 a_{i,k^*}
\end{align*}
Since the first layer of the actor network seems to corrupt spatial information but preserve binary change information \autoref{fig:ActorLayer1Activation}, we can extrapolate that the primary function of self-attention in this task is to amplify change signals.

\section{Influence of Induced Bias on Value Estimates and Temporal Difference Errors}

Reinforcement learning (RL) provides a computational framework for modeling how agents learn to make decisions through interactions with their environment~\cite{sutton2018reinforcement}. In neuroscience, RL has been applied to explain how animals and humans adapt their behavior based on reward feedback~\cite{dayan2008decision, niv2009reinforcement}. A key component of RL is the \textit{value function} $V(H^{(t)})$ (\autoref{eq:V}), which estimates the expected cumulative future reward from a given the mnemonic percept $H^{(t)}$ at time $t$. The value function is updated using the \textit{temporal difference (TD) error} $\delta_t$ (\autoref{eq:td_error}), which quantifies the discrepancy between expected and received rewards. In the brain, dopaminergic neurons have been associated with encoding TD errors~\cite{schultz2000neuronal, holroyd2002neural}. Fluctuations in dopamine release have been found to correspond to reward prediction errors, influencing synaptic plasticity and learning processes~\cite{glimcher2011understanding}. Attention mechanisms, which prioritize certain sensory inputs over others, could be motivated by value~\cite{maunsell2004neuronal, hikosaka2006basal, anderson2016role, anderson2016attention, failing2018selection} while also potentially influencing value estimates~\cite{lim2011decision, maunsell2015neuronal, leong2017dynamic}. Biasing attention toward specific stimuli may thus modulate the computation of TD errors. 

\subsection{Effects of Bias Manipulation on Temporal Difference Errors}

\begin{figure}[htbp]
    \centering
    \includegraphics[width=0.99\linewidth]{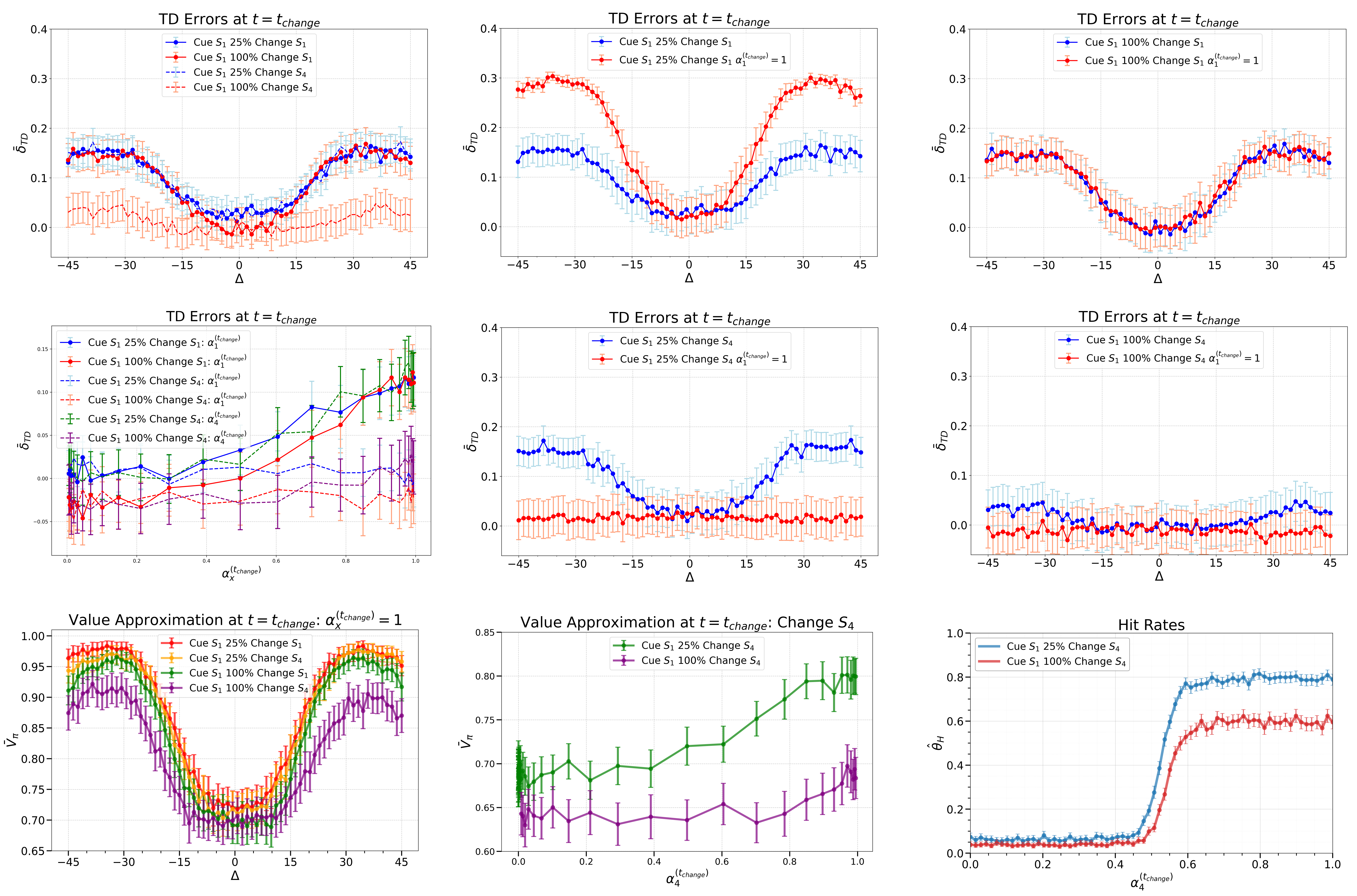}
    \caption{ Plots showing the TD errors ($\delta_{TD}$ from \autoref{eq:td_error} in \textbf{A--F}) and value estimates ($V_\pi$ from \autoref{eq:V} in \textbf{G} and \textbf{H}). Data points are averaged over 500 trials. Cues always occur at the $S_1$ location and changes occur at either the $S_1$ or $S_4$ location. In \textbf{D}, bias is modulated with respect to the bias indicated in the legend, i.e., $\alpha_x^{(t_{change})}=\alpha_4^{(t_{change})}$ for the dashed purple line. In \textbf{G}, bias is maximized with respect to the change location for all $\Delta$ shown, i.e., for the purple curve $\alpha_4^{(t_{change})}=1$. In \textbf{I} the hit-rate is shown for modulation of bias $\alpha_4^{(t_{change}}$ and changes on $S_4$ }
    \label{fig:3} 
\end{figure}

In Figure~\ref{fig:3}, we present the TD errors and value estimates from our model under various conditions. Figure~\ref{fig:3}A compares the TD errors when a cue is presented at location $S_1$, and a change occurs either at $S_1$ or $S_4$. When a high-validity cue is shown at a location different from where the change occurs (dashed red line in Figure~\ref{fig:3}A), we observe significant suppression of the TD errors. Figures~\ref{fig:3}B and \ref{fig:3}C demonstrate the influence of bias manipulations on the TD errors. For low-validity cues, maximizing the bias on the change location (setting $\alpha_{1}^{(t_{change})} = 1$: red curves in \autoref{fig:3}B and C) results in significant increases in the TD error (red line in Figure~\ref{fig:3}B). However, for high-validity cues, even a maximized bias has negligible effects (overlapping curves in Figure~\ref{fig:3}C). When we force a high bias at locations other than the change location (Figures~\ref{fig:3}E and \ref{fig:3}F), we observe a nearly complete suppression of the TD errors. In summary, the TD errors significantly increase with respect to the bias when the biased location and the change location coincide (red, blue, and green dashed lined in Figure~\ref{fig:3}D). However, when the change location and the assigned bias do not coincide, or a high validity cue was shown at a location different than the change location, we see a significant suppression of TD errors (dashed blue, dashed red, and dashed purple lines in Figure~\ref{fig:3}D).

\subsection{Effects of Bias Manipulation on Value Estimates}

The cue location, probability assignment by the cue, and the induced bias all affect the TD errors at the time of change. Similar to the method used to generate Figures~\ref{fig:2}H and \ref{fig:2}K, we sought to isolate the effects of the cue from the secondary bias manipulation the cue induces at the time of change. Therefore, in Figure~\ref{fig:3}G, we varied the cue and change conditions but maximized the bias on the change location at the time of change (setting $\alpha_{1}^{(t_{\text{change}})} = 1$ for a change occurring at location $S_{1}$). A high-validity cue on a location different from the change location causes a decrease in value estimates, even for large orientation changes $\Delta$ (purple lines in Figure~\ref{fig:3}G). We also show that the value estimate increases approximately linearly with the bias, $\alpha_4^{(t_{change})}$, when there is a low-validity cue at an opposing location and the change location ($S_4$) differs from the cued location (green line in Figure~\ref{fig:3}H). The value estimation is significantly suppressed when there is a high-validity cue on a location different from the change (purple line in Figure~\ref{fig:3}H).

Considering the results shown in Figures~\ref{fig:3}G and \ref{fig:3}H together with those in Figure~\ref{fig:2}H, we might be tempted to attribute the increased decision criterion resulting from high validity cues to a decrease in perceived value, since our model's action selection is based solely on value estimates. However, in Figure~\ref{fig:3}I, we demonstrate that there is a sharp increase in response rates before any such increase in value estimates (red curve in Figure~\ref{fig:3}I compared to the purple curve in Figure~\ref{fig:3}H). This suggests that there is some level of generalization beyond that of value estimates in our model's action selection policy.

\section{Manipulating Bias Influences Criterion and Sensitivity}

\subsection{Criterion and Sensitivity in Signal Detection Theory}

Signal detection theory provides a framework for analyzing how observers detect signals in the presence of noise. Two key measures in this theory are criterion and sensitivity. The criterion represents the observer's decision threshold for reporting a signal as present. Mathematically, it is defined as:
\[c = -\frac{1}{2}(z(\theta_H) + z(\theta_{FA}))\]
Where $z(\theta_H)$ is the z-score of the hit rate ($\theta_H$) and $z(\theta_{FA})$ is the z-score of the false alarm rate ($\theta_{FA}$). Intuitively, the criterion reflects how liberal or conservative the observer is in reporting signals. A lower (more negative) criterion indicates a liberal bias - the observer is more willing to report signals even with weak evidence. A higher (more positive) criterion indicates a conservative bias - the observer requires stronger evidence to report a signal.

Sensitivity, often denoted as $d^\prime$ (d-prime), measures the observer's ability to discriminate between signal and noise. It is calculated as:
\[d' = z(\theta_H) - z(\theta_{FA})\]
Sensitivity represents the standardized difference between the means of the signal and noise distributions. A higher $d^\prime$ indicates better ability to distinguish signal from noise. These measures allow researchers to separate an observer's inherent sensitivity to signals from their decision-making strategy (criterion). By analyzing both, we can understand not just how well an observer detects signals, but also their underlying decision-making processes.

\subsection{Random Variable Interpretation of the Hit-Rate and False-Alarm Rate}
Let $X_H$ and $X_{FA}$ be random variables associated with the agent's outcome on a single trial. On change trials, when $X_H=1$, the agent successfully detected a change and the outcome is a hit. If $X_H=0$, the agent did not detect a change and the outcome of the trial is a miss. On no-change trials, $X_{FA}=1$ if the outcome of the trial was a false alarm and $X_{FA}=0$ when the outcome is a correct reject. If we condition on the trial type, then the two random variables are independent, i.e.,
\begin{align*}
    X_{H}|\{\text{\textquotedblleft Change Trial\textquotedblright}\} \perp X_{FA}|\{\text{\textquotedblleft No-Change Trial\textquotedblright}\}
\end{align*}
If we let $\theta_{H}$ and $\theta_{FA}$ be the hit-rate and false-alarm-rate, then 
\begin{equation}
\begin{aligned}
\hat{\theta}_{H} &= \frac{n_H}{n_H + n_M} & \hat{\theta}_{FA} &= \frac{n_{FA}}{n_{FA} + n_{CR}}
\end{aligned}
\end{equation}
where $\hat{\theta}_H$ and $\hat{\theta}_{FA}$ is an estimator for the true hit-rate and false-alarm-rate, $\theta_H$ and $\theta_{FA}$.
For brevity, let $X_p$ and $\theta_p$ be such that $p \in \{\text{H,FA}\}$, and assume the trial conditioning described above. Given a sequence of $n$ random variables, $\{X_p^{(i)}\}_{i=1}^{n}$, we can model $\theta_p$ as 
\begin{align*}
    \hat{\theta}_p &= \frac{1}{n}\sum_{i=1}^{n}X_p^{(i)} \\
    X_p^{(i)} &\sim \text{Bern}(\theta_p)
\end{align*}
where $\text{Bern}(\theta_p)$ is a Bernoulli random variable with parameter $\theta_p$. By the law of large numbers, we know that this estimator converges to the true parameter in the large sample limit
\begin{align*}
    \lim_{n \rightarrow \infty} \frac{1}{n}\sum_{i=1}^{n}X_p^{(i)} = \theta_p
\end{align*}

Using the Central Limit Theorem (CLT), we can also compute the variance of this estimate. By CLT, 
\begin{align*}
    \sqrt{n}\left[ \hat{\theta}_p - \theta_p \right] \rightarrow \mathcal{N}(0,\hat{\sigma}_p^2)
\end{align*}
where $\hat{\sigma}_p^{2} = \hat{\theta}_p(1-\hat{\theta}_p)$ and $\hat{\sigma}^2_p\rightarrow \sigma^2_p$ in the large sample limit. Thus, the variance of our parameter estimate is 
\begin{align*}
    Var(\hat{\theta}_p) \approx  \frac{\hat{\sigma}_p^2}{n}
\end{align*}

\subsection{z-Score as a Random Variable Transformation}
If we treat $\hat{\theta}_p$ as a random variable, then computing the z-score is a transformation of this random variable. Hence, we can use the Delta Method to derive the approximate properties of this transformation. By the Delta Method:
\begin{align*}
    \sqrt{n}\left[ z(\hat{\theta}_p) - z(\theta_p) \right] \rightarrow \mathcal{N}\left( 0, \hat{\sigma}_p^2 [z^\prime(\hat{\theta}_p)]^2 \right)
\end{align*}
where $z^\prime(\hat{\theta}_p)=\frac{dz}{dx}|_{\hat{\theta}_p}$. This gives
\begin{align*}
    Var(\hat{\theta}_p) \approx \frac{\hat{\sigma}_p^2}{n} z^\prime(\hat{\theta}_p)
\end{align*}
We can evaluate $z^\prime(\hat{\theta}_p)$ by using the Inverse Function Theorem (IFT). Let $\psi(x)$ be the standard normal probability density function, i.e.,
\begin{align*}
    \psi(x) = \frac{1}{\sqrt{2\pi}}X^{\frac{-x^2}{2}}
\end{align*}
Then the cumulative distribution function is
\begin{align*}
    \phi(x) = \int_{-\infty}^x \psi(t)dt
\end{align*}
This gives $z^\prime(\hat{\theta}_p) = \frac{d\phi^{-1}}{dx}|_{(\hat{\theta}_p)}$. By IFT,
\begin{align*}
    z^\prime(\hat{\theta}_p) &=\left. \frac{d\phi^{-1}}{dx}\right|_{\hat{\theta}_p} \\
    &= \frac{1}{\left( \left. \frac{d\phi}{dx} \right|_{\phi^{-1}(\hat{\theta}_p)} \right)} \\
    &= \frac{1}{\psi(\phi^{-1}(\hat{\theta}_p))}
\end{align*}
Together with the above, this yields
\begin{align*}
    Var(\hat{\theta}_p) \approx \left(\frac{\hat{\sigma}_p^2}{n}\right) \left(\frac{1}{\psi(\phi^{-1}(\hat{\theta}_p))}\right)
\end{align*}

\subsection{Criterion and Sensitivity are Normally Distributed in the Large Sample Limit}
Since we are conditioning on the trial type, our estimates of the hit-rate and false-alarm-rate are independent. Therefore, the criterion and sensitivity consists of the summation of two independent normal (approximate) random variables. For the estimate ($\hat{c}$) of a true criterion ($c$), we have
\begin{align*}
    \hat{c} \sim \mathcal{N}\left( c,  \frac{1}{4}\left(\frac{\hat{\sigma}_{H}^2}{n_{CT}} \left(\frac{1}{\psi(\phi^{-1}(\hat{\theta}_{H}))}\right) + \frac{\hat{\sigma}_{FA}^2}{n_{NT}} \left(\frac{1}{\psi(\phi^{-1}(\hat{\theta}_{FA}))}\right)\right)\right)
\end{align*}
For the estimate ($\hat{d}^\prime$) of a true sensitivity ($d^\prime$), we have
\begin{align*}
    \hat{d}^\prime \sim \mathcal{N}\left( d^\prime,  \left(\frac{\hat{\sigma}_{H}^2}{n_{CT}} \left(\frac{1}{\psi(\phi^{-1}(\hat{\theta}_{H}))}\right) + \frac{\hat{\sigma}_{FA}^2}{n_{NT}}  \left(\frac{1}{\psi(\phi^{-1}(\hat{\theta}_{FA}))}\right)\right)\right)
\end{align*}

\begin{figure}[htbp]
    \centering
    \vspace*{-0mm} 
    \includegraphics[width=0.99\linewidth]{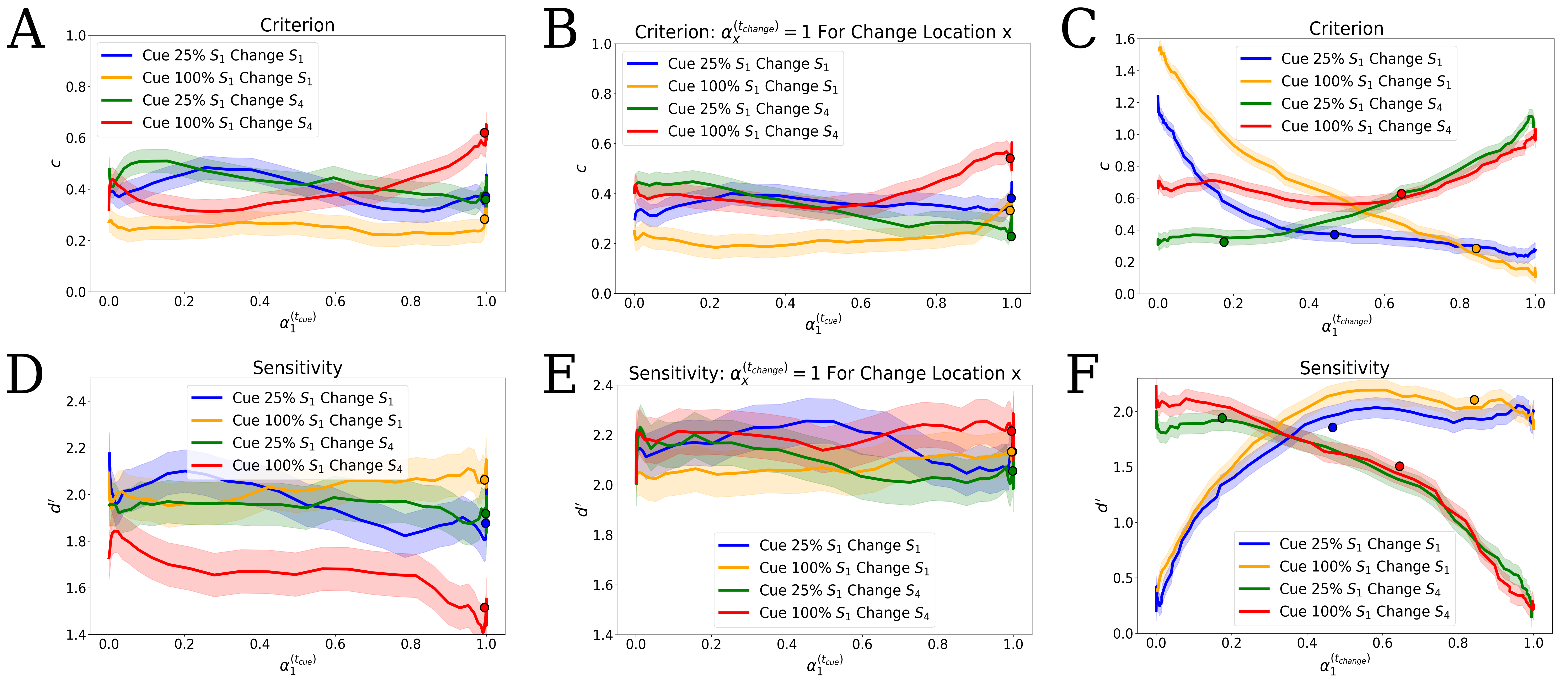}
    \caption{Plots showing the effect of artificially modulating the bias. All data points are the result of an average over 500 trials. Artificial modulation involves inducing a high bias in one of the columns of the self-attention map. In all cases, the bias is induced on the competitors with respect to the $S_1$ ($\alpha_1^{(t)}$ and $\xi_1^{(t)}$) location or the $S_4$ ($\alpha_4^{(t)}$ and $\xi_4^{(t)}$) location. In \textbf{A--C}, we plot the hit rates versus the orientation change $\Delta$. If $\alpha_i^{(t_{\text{change}})}=1$, this indicates that the visual percept $Z^{(t_{\text{change}})}$ has been completely biased toward $\xi_i^{(t_{\text{change}})}$, i.e., there are no traces of other competitors in the visual percept. We manipulate the bias to this extreme for either the $\xi_1^{(t_{\text{change}})}$ or $\xi_4^{(t_{\text{change}})}$ competitors in \textbf{A--C} and then again for the reaction time plots in \textbf{D--F}. In \textbf{G--I}, we show the criterion measured (Equation~\ref{arxiveq:crit}) after manipulating the bias with respect to the $S_1$ location at the time of cue ($\alpha_1^{(t_{\text{cue}})}$ in \textbf{G} and \textbf{H}) or at the time of orientation change ($\alpha_1^{(t_{\text{change}})}$ in \textbf{I}). The colored dots indicate the average criterion measured without bias manipulations. The criterion was evaluated over 500 trials where there was a 50\% chance of the change occurring at the indicated locations; otherwise, no change occurred. This procedure is repeated in \textbf{J--L} with the exception that we are measuring the sensitivity (Equation~\ref{eq:sens}). In \textbf{H} and \textbf{K}, we modulate the bias with respect to the $S_1$ location at the time of cue ($\alpha_1^{(t_{\text{cue}})}$), but then immediately following the cue, we maximize the bias for the location where the change will occur ($\alpha_x^{(t_{\text{change}})}$ for a change on $S_x$, $x \in \{1,4\}$). This is to prevent the secondary biasing effects that the cue induces at change time from affecting our comparisons.}
    \label{fig:2}
\end{figure}

To quantify the effects of bias modulation further, we analyzed the decision criterion and sensitivity measures under different bias manipulations. In signal detection theory, an observer's ability to distinguish between signal and noise is measured using two key metrics: sensitivity ($d'$) and criterion ($c$). Sensitivity measures how well the observer can distinguish between signal and noise, calculated using the hit rate ($\hat{\theta}_{H}$) and the false alarm rate ($\hat{\theta}_{FA}$). The formula for $d'$ is:
\begin{equation}
    d' = Z(\hat{\theta}_{H}) - Z(\hat{\theta}_{FA}),
    \label{eq:sens}
\end{equation}
where $Z$ represents the inverse cumulative distribution function (CDF) of the standard normal distribution, also known as the z-score (not to be confused with the visual percept $Z^{(t)}$ in our model). Criterion measures the observer's decision bias, indicating the tendency to favor one type of response over another:
\begin{equation}
    c = -\frac{1}{2} \left[ Z(\hat{\theta}_{H}) + Z(\hat{\theta}_{FA}) \right].
    \label{eq:crit}
\end{equation}
We assessed these metrics for individual stimulus locations by modifying the task environment so that only the stimulus location in question could experience an orientation change, with change trials occurring with probability $0.5$. For example, if we are interested in the criterion of the agent's response with respect to stimulus location $S_4$ given a $100\%$ validity cue at $S_1$, then we run multiple trials with the cue at $S_1$ and only allow orientation changes to occur at $S_4$. We then compute Equations~\eqref{eq:sens} and \eqref{eq:crit} over the obtained behavioral data.

\subsection{Bias effects on criterion and sensitivity}
In the literature, there is ongoing debate regarding how attention modulates criterion and sensitivity in visual tasks. Some studies have found that attention can modulate both the decision criterion and sensitivity, indicating that attention enhances perceptual processing and decision-making~\cite{luo2018attentional}. Specifically, Luo and Maunsell~\cite{luo2018attentional} showed that attentional cues not only improve the ability to discriminate between stimuli (increasing sensitivity) but also affect the observer's response bias (criterion). Contrastingly, recent experimental work suggests that presaccadic attention may predominantly influence the decision criterion rather than sensitivity. For instance, Gupta et al.~\cite{gupta2024presaccadic} found that shifts of attention just before eye movements adjust the criterion without significantly affecting sensitivity, implying that attention reallocations linked to saccades might alter decisional strategies more than perceptual encoding. Other studies propose that strategically modulating the decision criterion can indirectly enhance sensitivity. Wang et al.~\cite{wang2022neuronal} argue that criterion shifts, when optimized, can lead to improvements in sensitivity by better aligning the decision boundary with the signal distribution, particularly in tasks with asymmetric signal distributions or unequal prior probabilities. This suggests a complex interplay between criterion and sensitivity, where attentional mechanisms can influence perceptual performance both directly and indirectly.

Our findings contribute to this debate by showing that artificially manipulating the bias in our model can differentially affect criterion and sensitivity. In Figure~\ref{fig:2}G and J, we manipulated the bias at the time of cue $t_{\text{cue}}$ with respect to the cue location (e.g., adjusting $\alpha_1^{(t_{\text{cue}})}$ for a cue at $S_1$). We observed that maximizing the bias at the cue location, for a high validity assigning cue, increases the criterion and decreases sensitivity when the change occurs at a different location (red curves in \autoref{fig:2}G and J). This suggests that the model's attentional focus at the cue time can affect its ability to detect changes elsewhere. Notably, under natural conditions without artificial bias manipulations, attention is already maximized at the cue location (dots in Figures~\ref{fig:2}G, H, J, K). 

We know that spatial bias at the time of change is affected by the attributes of the cue. For example, a 100\% valid cue at the $S_1$ location at the time of cue leads to higher bias on the cued location at the time of change. To isolate the effect of bias at the time of cue from secondary effects at the time of change, we conducted additional experiments where we maximized the bias for the change location at $t_{\text{change}}$ (Figures~\ref{fig:2}H and K). This negated the secondary bias effect resulting from the cue and offers more equality in comparisons. We found that the cue's effect on the criterion remained largely unchanged (\autoref{fig:2}H), but the impact on sensitivity was abolished (\autoref{fig:2}K), indicating that the decrease in sensitivity observed in \autoref{fig:2}J is a result of spatial bias being directed away from the change location at the time of change.

In Figures~\ref{fig:2}I and L, we demonstrate that manipulating the bias at the time of change ($\alpha_1^{(t_{\text{change}})}$) significantly influences critetion and sensitivity. Increasing bias toward the change location decreases the criterion and increases sensitivity (blue and gold curves in \autoref{fig:2}I and L), while increasing bias toward non-change locations has the opposite effect (green and red curves in \autoref{fig:2}I and L). This occurs because increasing $\alpha_i^{(t_{\text{change}})}$ amplifies the expression level of the spatially corresponding internal representation $\xi_i^{(t)}$. If the change occurred at stimulus location $S_i$ at time $t = t_{\text{change}}$, then the change information would initially be localized in $\xi_i^{(t_{\text{change}})}$, since $xi_i^{(t)}$ is the only competing internal representation that consists of information from the immediate visual patch $x_i^{(t)}$. Increasing $\alpha_i^{(t_{\text{change}})}$ increases the signal strength, or expression, of this change information in the visual percept $Z^{(t)}$ to working memory. Since biasing the internal representation associated with the change location is directly amplifying the most task relevant information in the visual percept, it seems more fitting to describe attention at the time of change as mostly modulating sensitivity, and that criterion decreases is a result of conversions from misses to hits. This effect mirrors the enhanced perceptual sensitivity observed in microstimulation experiments~\cite{moore2003selective, cavanaugh2006enhanced}.

It is important to note that these results are not capable of resolving the debate about the attentional effects on criterion and sensitivity. Rather, they show that the effects of attention on criterion and sensitivity can be quite complex, and possibly task dependent. For example, the criterion modulation shown in \autoref{fig:2} G and H is a result of cue information being transmitted to working memory, where this information has changed the structure of the internal representations in VWM. The cue information has changed the decodability of activated memory transmitted to the reinforcement learning module, affecting behavior in a way that results in a criterion shift. Bias at the time of change mostly results in sensitivity changes. However, bias assignment at the time of change involves multiple interactions. When activated memory patches are drawn from the VWM patches at the time of change, they interact with immediate visual inputs to bias attention on the spatial internal representations associated with the cue location. Moreover, to observe an orientation change there must be stored information of the original stimulus orientations in working memory (since these are drawn randomly). This information also interacts with visual inputs to assign bias to potential change locations. These results and the dynamics displayed by our model demonstrate that criterion and sensitivity effects can involve multiple overlapping pathways and interactions between working memory and attention. 

\section{Comparison of supervised and RL training signals}

\begin{figure}[htbp]
    \centering
    \vspace*{-0mm} 
    \includegraphics[width=0.95\linewidth]{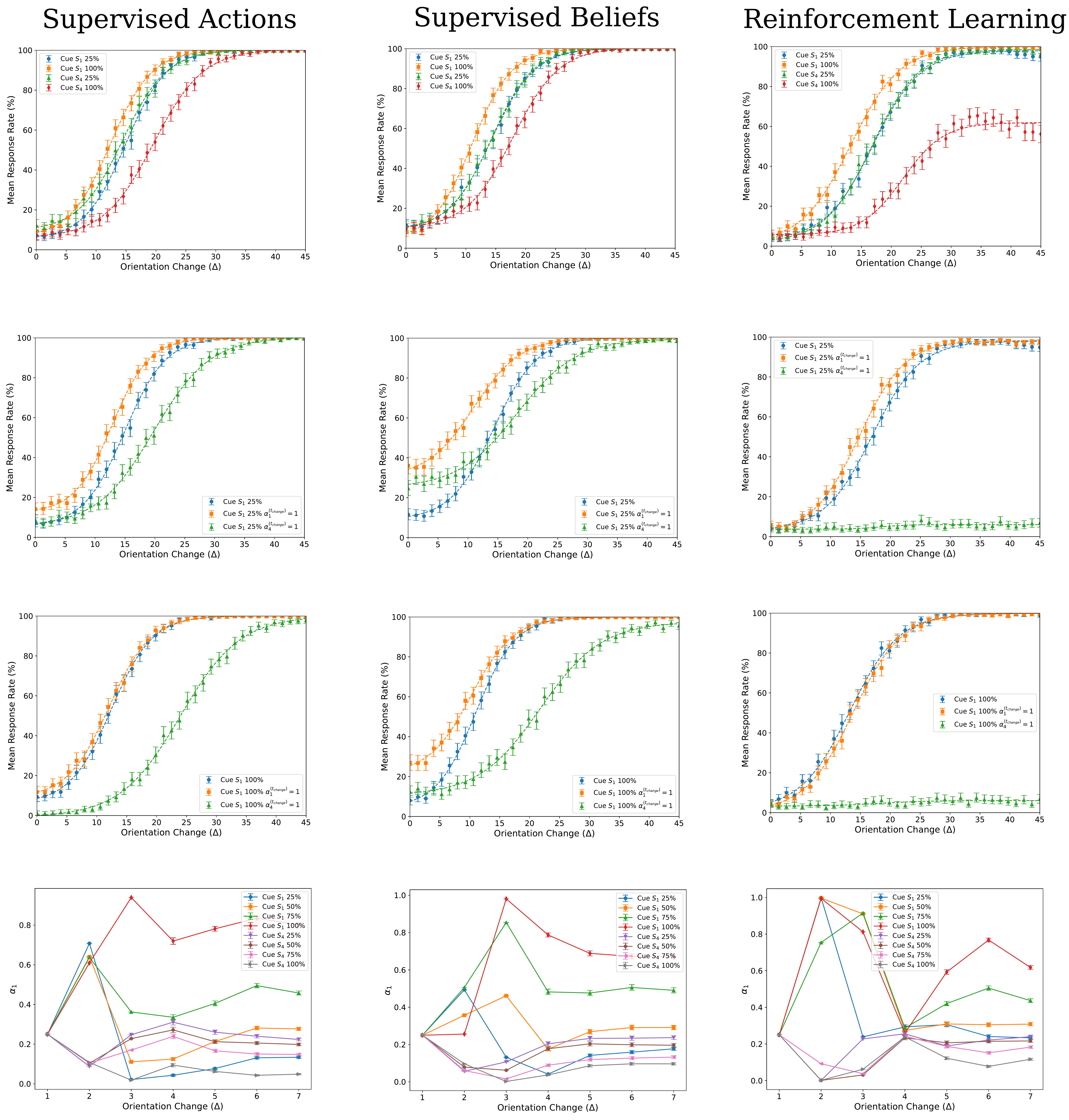}
    \caption{First three rows are behavioral results for the models being compared. The last row is the attention modulations at differing points in time of a trial. The first column is a model trained to predict the correct action given a sequence of visual inputs. The middle column is one in which the model is trained to predict whether the trial is a change trial or not. The right column is the model trained using reinforcement learning.}
    \label{fig:supervised}
\end{figure}

For the majority of our analysis, we study a model trained using a reinforcement learning signal, i.e., reward feedback. Here we show behavioral results from two models trained with a supervised signal \autoref{fig:supervised}. In both supervised training models, the ViT and LSTM architectures are identical to that of the model trained in reinforcement learning. Instead of an actor and critic neural network that decodes the mnemonic percept, we use a single decoder that takes $H^{(t)}$ as input and outputs a class $c^{(t)}$. This decoder is structured as follows: 

We first reshape the mnemonic percept 
\[
H^{(t)} \in \mathbb{R}^{n_{\text{patch}} \times d_{\text{ff}}}
\]
into a vector 
\[
H' \in \mathbb{R}^{n_{\text{patch}} \cdot d_{\text{ff}}}
\]
via
\[
H' = \mathrm{Flatten}\bigl(H^{(t)}\bigr).
\]
The decoder then computes an intermediate representation through a series of three fully-connected layers with layer normalization and ELU nonlinearity:
\[
\begin{aligned}
s_1 &= \mathrm{ELU}\Bigl(\mathrm{LN}_1\bigl(W_1 H' + b_1\bigr)\Bigr),\\[1mm]
s_2 &= \mathrm{ELU}\Bigl(\mathrm{LN}_2\bigl(W_2 s_1 + b_2\bigr)\Bigr),\\[1mm]
s_3 &= \mathrm{ELU}\Bigl(\mathrm{LN}_3\bigl(W_3 s_2 + b_3\bigr)\Bigr),
\end{aligned}
\]
where for \(i=1,2,3\), the matrices \(W_i\) and vectors \(b_i\) denote the weights and biases of the \(i\)th fully-connected layer, and \(\mathrm{LN}_i\) is the corresponding layer normalization operation. Finally, a fourth fully-connected layer followed by a sigmoid activation produces the scalar output:
\[
c = \sigma\Bigl(W_4 s_3 + b_4\Bigr),
\]
with \(W_4\) and \(b_4\) being the weight matrix and bias of the final layer, and \(\sigma(\cdot)\) denoting the sigmoid function. This scalar \(c\) is then used to determine either the action $a^{(t)}$ (left column) or the belief $b^{(t)}$ (middle column).

In the first column of \autoref{fig:supervised} we supervise the actions. Each training sample consists of a sequence of visual inputs $\{X^{(t)}\}_{t=1}^T$ and a sequence of ``correct'' actions, $\{\bar{a}^{(t)}\}_{t=1}^T$. The correct action to take until the time of change is always fixed to $\bar{a}^{(t)}=0$ for $t < t_{change}$. This is equivalent to the ``wait'' action in the RL model. If the trial is a change trial, the correct actions following change onset are $\bar{a}^{(t)}=1$ for $t\geq t_{change}$. On the other hand, the correct actions for ``no change'' trials is $\bar{a}^{(t)}=0$ for $t\geq t_{change}$.

In the middle column of \autoref{fig:supervised}, we supervise the beliefs over whether the trial is a change trial or not. The inputs are the same as the actions, but the labels are now $\{\bar{b}^{(t)}\}_{t=1}^{(T)}$, with $\bar{b}^{(t)}=1$ for $\forall t$ if the trial is a change trial and $\bar{b}^{(t)}=0$ $\forall t$ otherwise. To determine whether the agent ``declared'' a change or not, at $t\geq t_{change}$ we round the agent output $b^{(t)}$ to either 1 or 0, with 1 being a response and 0 indicating no response. We call this supervised training a belief signal, because the correct response from the agent for time points before the time of change is $b^{(t)}=0.5$ for $t<t_{change}$

\newpage

\bibliographystyle{plain}
\bibliography{ref2}  